\documentclass{article}

\usepackage{times}
\usepackage{epsfig} 
\usepackage{graphicx} 
\usepackage{ dsfont }

\usepackage{url}
\usepackage{mathtools}
\usepackage{subcaption}

\usepackage[export]{adjustbox}
\usepackage{makecell}
\usepackage{siunitx}

\usepackage[utf8]{inputenc} 
\usepackage[T1]{fontenc}    
\usepackage{url}            
\usepackage{booktabs}       
\usepackage{amsfonts}       
\usepackage{nicefrac}       
\usepackage{microtype}      
\usepackage{caption}
\usepackage{amsmath,amssymb} 
\usepackage{color}
\usepackage[ruled]{algorithm}
\usepackage{tabularx}
\usepackage{multirow}
\usepackage{xcolor}

\usepackage[font={small}]{caption}
\usepackage[compact]{titlesec}

\usepackage{natbib}

\usepackage[pagebackref=true,breaklinks=true,letterpaper=true,bookmarks=false]{hyperref}

\usepackage[accepted]{icml2017}

\newcommand{\x}{{\mathbf{x}}}

\newcommand{\p}{{\mathbf{p}}}
\newcommand{\h}{{\mathbf{h}}}
\newcommand{\w}{{\mathbf{w}}}
\newcommand{\cc}{{\mathbf{c}}}
\newcommand{\ignore}[1]{}

\iftrue 
        \newcommand{\cutsectionup}{\vspace*{-0.1in}}
        \newcommand{\cutsectiondown}{\vspace*{-0.05in}}

        \newcommand{\cutsubsectionup}{\vspace*{-0.09in}}
        \newcommand{\cutsubsectiondown}{\vspace*{-0.04in}}

\else 
        \newcommand{\cutsectionup}{}
        \newcommand{\cutsectiondown}{}

        \newcommand{\cutsubsectionup}{}
        \newcommand{\cutsubsectiondown}{}

\fi

\icmltitlerunning{Learning to Generate Long-term Future via Hierarchical Prediction}

\begin{document} 

\twocolumn[
\icmltitle{Learning to Generate Long-term Future via Hierarchical Prediction}



\icmlsetsymbol{equal}{*}

\renewcommand*{\thefootnote}{\fnsymbol{footnote}}

\begin{icmlauthorlist}
\icmlauthor{Ruben Villegas}{umich,equal}
\icmlauthor{Jimei Yang}{ado}
\icmlauthor{Yuliang Zou}{umich}
\icmlauthor{Sungryull Sohn}{umich}
\icmlauthor{Xunyu Lin}{chi}
\icmlauthor{Honglak Lee}{umich,goo} 
\end{icmlauthorlist}

\renewcommand*{\thefootnote}{\arabic{footnote}}
\setcounter{footnote}{0}

\icmlaffiliation{umich}{Department of Electrical Engineering and Computer Science, University of Michigan, Ann Arbor, MI, USA.}
\icmlaffiliation{ado}{Adobe Research, San Jose, CA.}
\icmlaffiliation{goo}{Google Brain, Mountain View, CA}
\icmlaffiliation{chi}{Beihang University, Beijing, China.}

\icmlcorrespondingauthor{Ruben Villegas}{rubville@umich.edu}

\icmlkeywords{deep learning, machine learning, ICML, video prediction, future prediction}

\vskip 0.3in
]



\printAffiliationsAndNotice{\icmlEqualContribution} 

\begin{abstract}
    We propose a hierarchical approach for making long-term predictions of future frames.
    To avoid inherent compounding errors in recursive pixel-level prediction, we propose to first estimate high-level structure in the input frames, then predict how that structure evolves in the future, and finally by observing a single frame from the past and the predicted high-level structure, we construct the future frames without having to observe any of the pixel-level predictions.
    Long-term video prediction is difficult to perform by recurrently observing the predicted frames because the small errors in pixel space exponentially amplify as predictions are made deeper into the future.
    Our approach prevents pixel-level error propagation from happening by removing the need to observe the predicted frames.
    Our model is built with a combination of LSTM and analogy-based encoder-decoder convolutional neural networks, which independently predict the video structure and generate the future frames, respectively.
    In experiments, our model is evaluated on the Human3.6M and Penn Action datasets on the task of long-term pixel-level video prediction of humans performing actions and demonstrate significantly better results than the state-of-the-art.
\end{abstract}
\cutsectionup
\cutsectionup
\cutsectionup
\section{Introduction} \label{sec:introduction}
\cutsectiondown
Learning to predict the future has emerged as an important research problem in machine learning and artificial intelligence. 
Given the great progress in recognition (e.g., \cite{alexnet,szegedy2015going}), prediction becomes an essential module for intelligent agents to plan actions or to make decisions in real-world application scenarios~\cite{jayaraman2015learning,jayaraman2016look,FinnGL16}.
For example, robots can quickly learn manipulation skills when predicting the consequences of physical interactions.
Also, an autonomous car can brake or slow down when predicting a person walking across the driving lane.
In this paper, we investigate long-term future frame prediction that provides full descriptions of the visual world.

Recent recursive approaches to pixel-level video prediction highly depend on observing the generated frames in the past to make predictions further into the future~\cite{Oh15,Mathieu16,Goroshin15,Srivastava15,Ranzato14,FinnGL16,Villegas17,Lotter17}.
In order to make reasonable long-term frame predictions in natural videos, these approaches need to be highly robust to pixel-level noise. However, the noise amplifies quickly through time until it overwhelms the signal.
It is common that the first few prediction steps are of decent quality, but then the prediction degrades dramatically until all the video context is lost.
Other existing works focus on predicting high-level semantics, such as trajectories or action labels~\cite{Walker14,chao:cvpr2017,yuen2010data,lee2015modeling}, driven by immediate applications (e.g., video surveillance).
We note that such high-level representations are the major factors for explaining the pixel variations into the future.
%
%
%
In this work, we assume that the high-dimensional video data is generated from low-dimensional high-level structures, which we hypothesize will be critical for making long-term visual predictions.
Our main contribution is the hierarchical approach for video prediction that involves generative modeling of video using high-level structures.
Concretely, our algorithm first estimates high-level structures of observed frames, and then predicts their future states, and finally generates future frames conditioned on predicted high-level structures.

The prediction of future structure is performed by an LSTM that observes a sequence of structures estimated by a CNN, encodes the observed dynamics, and predicts the future sequence of such structures.
%
We note that \citet{Fragkiadaki} developed an LSTM architecture that can straightforwardly be adapted to our method.
However, our main contribution is the hierarchical approach for video prediction, so we choose a simpler LSTM architecture to convey our idea.
Our approach then observes a single frame from the past and predicts the entire future described by the predicted structure sequence using an analogy-making network~\cite{reed2015deep}.
In particular, we propose an image generator that learns a shared embedding between image and high-level structure information which allows us convert an input image into a future image guided by the structure difference between the input image and the future image.
We evaluate the proposed model on challenging real-world human action video datasets.
We use 2D human poses as our high-level structures similar to ~\citet{reed2016learning}.
Thus, our LSTM network models the dynamics of human poses while our analogy-based image generator network learns a joint image-pose embedding that allows the pose difference between an observed frame and a predicted frame to be transferred to image domain for future frame generation.
As a result, this pose-conditioned generation strategy prevents our network from propagating prediction errors through time, which in turn leads to very high quality future frame generation for long periods of time. 
Overall, the promising results of our approach suggest that it can be greatly beneficial to incorporate proper high-level structures into the generative process.

The rest of the paper is organized as follows:
A review of the related work is presented in Section~\ref{sec:relatedwork}.
The overview of the proposed algorithm is presented in Section~\ref{sec:overview}. 
The network configurations and their training algorithms are described in Section~\ref{sec:architecture} and Section~\ref{sec:training}, respectively.
We present the experimental details and results in Section~\ref{sec:experiments}, and conclude the paper with discussions of future work in Section~\ref{sec:conclusion}.

\cutsectionup
\section{Related Work} \label{sec:relatedwork}
\cutsectiondown
%
Early work on future frame prediction focused on small patches containing simple predictable motions \cite{NIPS2008_3567, michalski_grammar_cells,icml2014c2_mittelman14} and motions in real videos \cite{Ranzato14, Srivastava15}.
High resolution videos contain far more complicated motion which cannot be modeled in a patch-wise manner due to the well known aperture problem.
The aperture problem causes blockiness in predictions as we move forward in time.
\citet{Ranzato14} tried to solve blockiness by averaging over spatial displacements after predicting patches; however, this approach does not work for long-term predictions.

Recent approaches in video prediction have moved from predicting patches to full frame prediction.
\citet{Oh15} proposed a network architecture for action conditioned video prediction in Atari games.
\citet{Mathieu16} proposed an adversarial loss for video prediction and a multi-scale network architecture that results in high quality prediction for a few timesteps in natural video; however, the frame prediction quality degrades quickly.
\citet{FinnGL16} proposed a network architecture to directly transform pixels from a current frame into the next frame by  predicting a distribution over pixel motion from previous frames.
\citet{visualdynamics16} proposed a probabilistic model for predicting possible motions of a single input frame by training a motion encoder in a variational autoencoder approach.
\citet{Vondrick16} built a model that generates realistic looking video by separating background and foreground motion.
\citet{Villegas17} improved the convolutional encoder/decoder architecture by separating motion and content features. 
\citet{Lotter17} built an architecture inspired by the predictive coding concept in neuroscience literature that predicts realistic looking frames.

All the previously mentioned approaches attempt to perform video generation in a pixel-to-pixel process.
We aim to perform the prediction of future frames in video by taking a hierarchical approach of first predicting the high-level structure and then using the predicted structure to predict the future in the video from a single frame input.

To the best of our knowledge, this is the first hierarchical approach to pixel-level video prediction.
Our hierarchical architecture makes it possible to generate good quality long-term predictions that outperform current approaches.
The main success from our algorithm comes from the novel idea of first making high-level structure predictions which allows us to observe a single image and generate the future video by visual-structure analogy.
Our image generator learns a shared embedding between image and structure inputs that allows us to transform high-level image features into a future image driven by the predicted structure sequence.

\cutsectionup
\section{Overview} \label{sec:overview}
\cutsectiondown
\begin{figure*}[t]
    \begin{subfigure}{1\linewidth}
        \centering
	    \includegraphics[width=.9\linewidth]{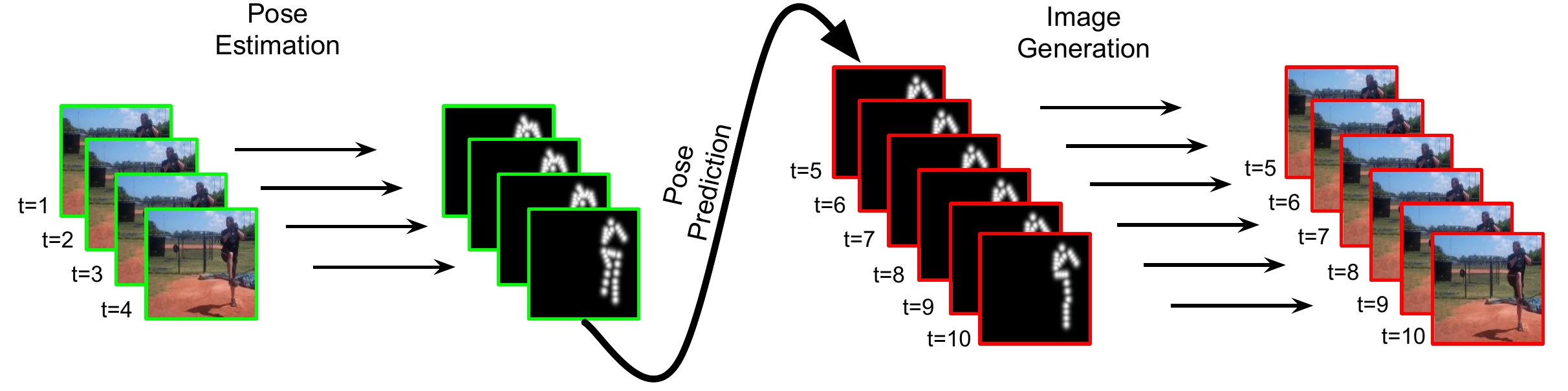}
	\end{subfigure} 
    \vspace{-0.15in}
    \caption{Overall hierarchical approach to pixel-level video prediction. Our algorithm first observes frames from the past and estimate the high-level structure, in this case human pose xy-coordinates, in each frame. The estimated structure is then used to predict the future structures in a sequence to sequence manner. Finally, our algorithm takes the last observed frame, its estimated structure, and the predicted structure sequence, in this case represented as heatmaps, and generates the future frames. Green denotes input to our network and red denotes output from our network.}
\label{fig:arch}
\vspace{-0.1in}
\end{figure*}

This paper tackles the task of long-term video prediction in a hierarchical perspective.
Given the input high-level structure sequence $\p_{1:t}$ and frame $\x_t$, our algorithm is asked to predict the future structure sequence $\p_{t+1:t+T}$ and subsequently generate frames $\x_{t+1:t+T}$.
The problem with video frame prediction originates from modeling pixels directly in a sequence-to-sequence manner and attempting to generate frames in a recurrent fashion.
Current state-of-the-art approaches recurrently observe the predicted frames, which causes rapidly increasing error accumulation through time.
Our objective is to avoid having to observe generated future frames at all during the full video prediction procedure.

Figure~\ref{fig:arch} illustrates our hierarchical approach.
Our full pipeline consists of 1) performing high-level structure estimation from the input sequence, 2) predicting a sequence of future high-level structures, and 3) generating future images from the predicted structures by visual-structure analogy-making given an observed image and the predicted structures.
We explore our idea by performing pixel-level video prediction of human actions while treating human pose as the high-level structure.
Hourglass network~\cite{hourglass} is used for pose estimation on input images.
Subsequently, a sequence-to-sequence LSTM-recurrent network is trained to read the outputs of hourglass network and to predict the future pose sequence.
Finally, we generate the future frames by analogy making using the pose relationship in feature space to transform the last observed frame.

The proposed algorithm makes it possible to decompose the task of video frame prediction to sub-tasks of future high-level structure prediction and structure-conditioned frame generation.
Therefore, we remove the recursive dependency of generated frames that causes the compound errors of pixel-level prediction in previous methods, and so our method performs very long-term video prediction.
\cutsectionup
\section{Architecture} \label{sec:architecture}
\cutsectiondown
This section describes the architecture of the proposed algorithm using human pose as a high-level structure.
Our full network is composed of two modules: an encoder-decoder LSTM that observes and outputs xy-coordinates, and an image generator that performs visual analogy based on high-level structure heatmaps constructed from the xy-coordinates output from LSTM.

\cutsubsectionup
\subsection{Future Prediction of High-Level Structures}
\cutsubsectiondown

\begin{figure}[t]
    \begin{subfigure}{1\linewidth}
	    \includegraphics[width=1\linewidth]{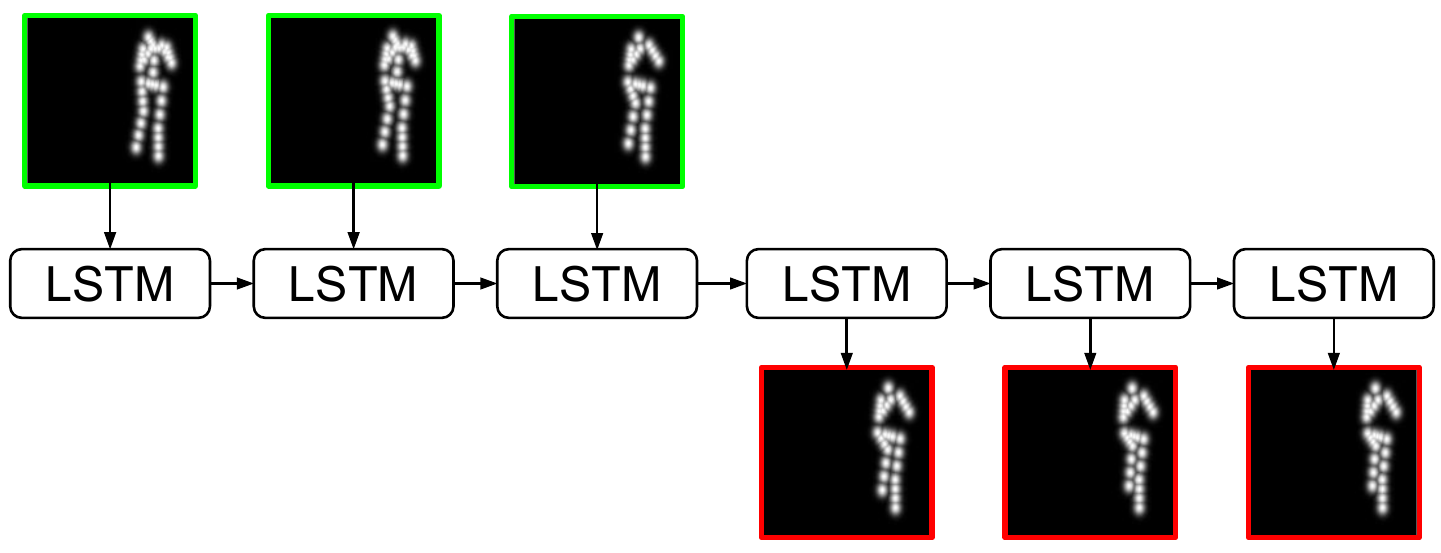}
	\end{subfigure} 
\caption{Illustration of our pose predictor. LSTM observes $k$ consecutive human pose inputs and predicts the pose for the next $T$ timesteps. Note that the human heatmaps are used for illustration purposes, but our network observes and outputs xy-coordinates.}
\label{fig:pose_predictor}
\vspace{-0.3in}
\end{figure}

Figure~\ref{fig:pose_predictor} illustrates our pose predictor.
Our network first encodes the observed structure dynamics by
%
\begin{equation}
\left[\h_{t},\cc_{t}\right]=\text{LSTM}\left(\p_t,\h_{t-1},\cc_{t-1}\right),
\label{eq:lstm}
\end{equation}
where $\h_{t} \in \mathbb{R}^H$ represents the observed dynamics up to time $t$, $\cc_{t} \in \mathbb{R}^H$ is the \textit{memory cell} that retains information from the history of pose inputs, $\p_t \in \mathbb{R}^{2L}$ is the pose at time $t$ (i.e., 2D coordinate positions of $L$ joints). 
In order to make a reasonable prediction of the future pose, LSTM has to first observe a few pose inputs to identify the type of motion occurring in the pose sequence and how it is changing over time.
LSTM also has to be able to remove noise present in the input pose, which can come from annotation error if using the dataset-provided pose annotation or pose estimation error if using a pose estimation algorithm.
After a few pose inputs have been observed, LSTM generates the future pose by
\begin{equation}
\hat{\p}_{t} = f\left(\w^{\top}\h_{t}\right),
\label{eq:corr1}
\end{equation}
where $\w$ is a projection matrix, $f$ is a function on the projection (i.e. tanh or identity), and $\hat{\p}_{t} \in \mathbb{R}^{2L}$ is the predicted pose.
In the subsequent predictions, our LSTM does not observe the previously generated pose.
Not observing generated pose in LSTM prevents errors in the pose prediction from being propagated into the future, and it also encourages the LSTM internal representation to contain robust high-level features that allow it to generate the future sequence from only the original observation.
As a result, the representation obtained in the pose input encoding phase must obtain all the necessary information for generating the correct action sequence in the decoding phase.
After we have set the human pose sequence for the future frames, we proceed to generate the pixel-level visual future.

\cutsubsectionup
\subsection{Image Generation by Visual-Structure Analogy}
\cutsubsectiondown

\begin{figure}[t]
    \vspace{-0.05in}
    \hspace*{-.1cm}
    \centering
	\includegraphics[width=0.95\linewidth]{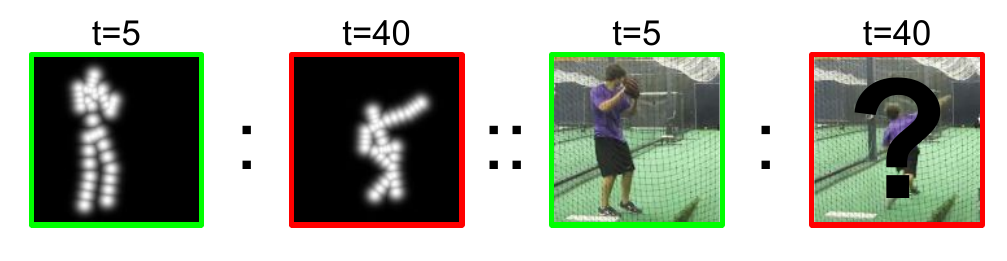}
	\vspace{-0.15in}
    \caption{Generating image frames by making analogies between high-level structures and image pixels.}
\label{fig:analogy}
\vspace{-0.05in}
\end{figure}

\begin{figure}[t]
    \includegraphics[width=1\linewidth]{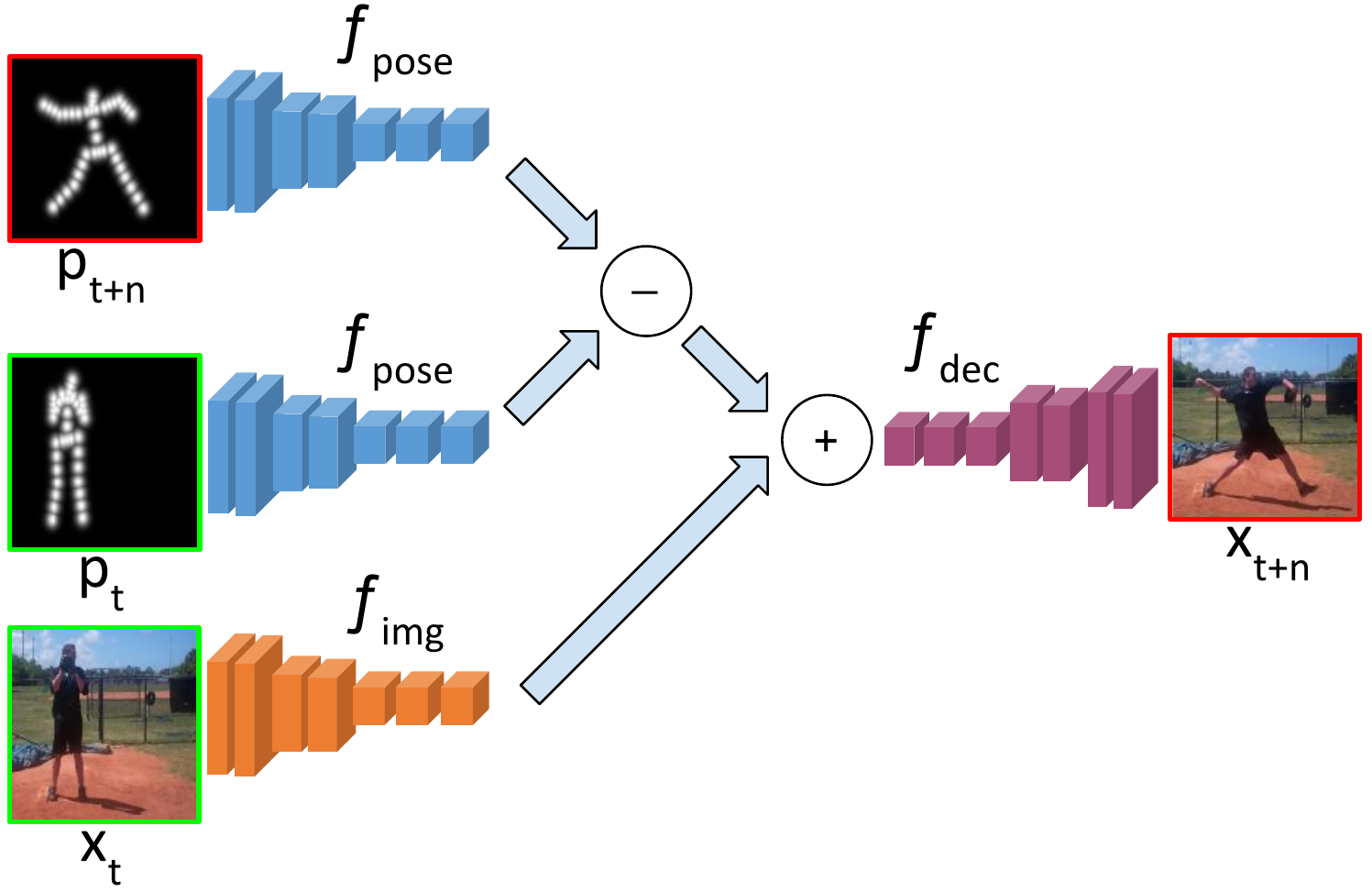}
    \vspace{-0.3in}
    \caption{Illustration of our image generator. Our image generator observes an input image, its corresponding human pose, and the human pose of the future image. Through analogy making, our network generates the next frame.}
\label{fig:image_generator}
\vspace{-0.1in}
\end{figure}

To synthesize a future frame given its pose structure, we make a visual-structure analogy inspired by \citet{reed2015deep} following $\p_t$ : $\p_{t+n}$ :: $\x_t$ : $\x_{t+n}$, read as "$\p_t$ is to $\p_{t+n}$ as $\x_t$ is to $\x_{t+n}$" as illustrated in Figure~\ref{fig:analogy}. Intuitively, the future frame $\x_{t+n}$ can be generated by transferring the structure transformation from $\p_t$ to $\p_{t+n}$ to the observed frame $\x_t$.
Our image generator instantiates this idea using a pose encoder $f_{\text{pose}}$, an image encoder $f_{\text{img}}$ and an image decoder $f_{\text{dec}}$.
Specifically, $f_{\text{pose}}$ is a convolutional encoder that specializes on identifying key pose features from the pose input that reflects high-level human structure.\footnote{Each input pose to our image generator is converted to concatenated heatmaps of each landmark before computing features.} 
$f_{\text{img}}$ is also a convolutional encoder that acts on an image input by mapping the observed appearance into a feature space where the pose feature transformations can be easily imposed to synthesize the future frame using the convolutional decoder $f_{\text{dec}}$. 
The visual-structure analogy is then performed by
%
\begin{equation}
\hat{\x}_{t+n} = f_{\text{dec}}\left(f_{\text{pose}}\left(g\left(\hat{\p}_{t+n}\right)\right)-f_{\text{pose}}\left(g\left(\p_t\right)\right)+f_{\text{img}}\left(\x_t\right)\right), 
\label{eq:corr2}
\end{equation}
%
where $\hat{\x}_{t+n}$ and $\hat{\p}_{t+n}$ are the generated image and corresponding predicted pose at time $t+n$, $\x_t$ and $\p_t$ are the input image and corresponding estimated pose at time $t$, and $g\left(.\right)$ is a function that maps the output xy-coordinates from LSTM into $L$ depth-concatenated heatmaps.\footnote{We independently construct the heatmap with a Gaussian function around the xy-coordinates of each landmark.} 
Intuitively, $f_{\text{pose}}$ infers features whose ``substractive" relationship is the same subtractive relationship between $\x_{t+n}$ and $\x_t$ in the feature space computed by $f_{\text{img}}$, i.e.,  $f_{\text{pose}}(g(\hat{\p}_{t+n}))-f_{\text{pose}}(g(\hat{\p}_{t})) \approx f_{\text{img}}(\x_{t+n}) - f_{\text{img}}(\x_t) $.
The network diagram is illustrated in  in Figure~\ref{fig:image_generator}.
The relationship discovered by our network allows for highly non-linear transformations between images to be inferred by a simple addition/subtraction in feature space.

\cutsectionup
\section{Training} \label{sec:training}
\cutsectionup

In this section, we first summarize the multi-step video prediction algorithm using our networks and then describe the training strategies of the high-level structure LSTM and of the visual-structure analogy network.
We train our high-level structure LSTM independent from the visual-structure analogy network, but both are combined during test time to perform video prediction.

\cutsubsectionup
\subsection{Multi-Step Prediction}
\cutsubsectiondown
Our algorithm multi-step video prediction procedure is described in Algorithm~\ref{fig:algo}.
Given input video frames, we use the Hourglass network~\cite{hourglass} to estimate the human poses $\p_{1:k}$.
High-level structure LSTM then observes $\p_{1:k}$, and proceeds to generate a pose sequence $\hat{\p}_{k+1:k+T}$ where $T$ is the desired number of time steps to predict.
Next, our visual-structure analogy network takes $\x_k$, $\p_k$, and $\hat{\p}_{k+1:k+T}$ and proceeds to generate future frames $\hat{\x}_{k+1:k+T}$ one by one.
Note that the future frame prediction is performed by observing pixel information from only $\x_k$, that is, we never observe any of the predicted frames.

\begin{algorithm}[t]
\caption{Video Prediction Procedure \label{fig:algo}}
\begin{algorithmic}
\STATE input: $\mathbf{x}_{1:k}$
\STATE output: $\hat{\mathbf{x}}_{k+1:k+T}$
\FOR{$t$=$1$ to $k$}
    \STATE{$\p_t \gets \text{Hourglass}(\x_{t})$}
    \STATE{$\left[\h_t,\cc_t\right] \gets \text{LSTM}(\p_{t},\h_{t-1},\cc_{t-1})$}
\ENDFOR

\FOR{$t$=$k+1$ to $k+T$}
    \STATE{$\left[\h_t,\cc_t\right] \gets \text{LSTM}(\h_{t-1},\cc_{t-1})$}
    \STATE{$\hat{\p}_t \gets f\left(\w^{\top}\h_{t}\right)$}
	\STATE{$\hat{\x}_{t} \gets f_{\text{dec}}\left(f_{\text{pose}}\left(g\left(\hat{\p}_{t}\right)\right)-f_{\text{pose}}\left(g\left(\p_k\right)\right)+f_{\text{img}}\left(\x_k\right)\right)$}
\ENDFOR
\end{algorithmic}
\label{alg:multi_pred}
\end{algorithm}

\cutsubsectionup
\subsection{High-Level Structure LSTM Training} \label{sec:lstm_training}
\cutsubsectiondown
We employ a sequence-to-sequence approach to predict the future structures (i.e. future human pose).
Our LSTM is \textit{unrolled} for $k$ timesteps to allow it to observe $k$ pose inputs before making any prediction.
Then we minimize the prediction loss defined by
\vspace{-5pt}
\begin{equation}
\mathcal{L}_{\text{pose}} = \frac{1}{TL}\sum_{t=1}^T\sum_{l=1}^L\mathds{1}_{\{m_{k+t}^l=1\}}\|\hat{\p}_{k+t}^l-\p_{k+t}^l\|^2_2,
\label{eq:full_loss_lstm}
\vspace{-5pt}
\end{equation}
where $\hat{\p}_{k+t}^l$ and $\p_{k+t}^l$ are the predicted and ground-truth pose $l$-th landmark, respectively, $\mathds{1}_{\{.\}}$ is the indicator function, and $m_{k+t}^l$ tells us whether a landmark is visible or not (i.e. not present in the ground-truth).
Intuitively, the indicator function allows our LSTM to make a guess of the non-visible landmarks even when not present at training.
Even in the absence of a few landmarks during training, LSTM is able to internally understand the human structure and observed motion.
Our training strategy allows LSTM to make a reasonable guess of the landmarks not present in the training data by using the landmarks available as context.



\cutsubsectionup
\subsection{Visual-Structure Analogy Training} \label{sec:G_training}
\cutsubsectiondown
Training our network to transform an input image into a target image that is too close in image space can lead to suboptimal parameters being learned due to the simplicity of such task that requires only changing a few pixels.
Because of this, we train our network to perform random \textit{jumps in time} within a video clip.
Specifically, we let our network observe a frame $\x_t$ and its corresponding human pose $\p_t$, and force it to generate frame $\x_{t+n}$ given pose $\p_{t+n}$, where $n$ is defined randomly for every iteration at training time.
Training to jump to random frames in time gives our network a clear signal the task at hand due to the large pixel difference between frames far apart in time.

To train our network, we use the compound loss from \citet{psim}. Our network is optimized to minimize the objective given by
\vspace{-5pt}
\begin{equation}
\mathcal{L} = \mathcal{L}_{\text{img}}+\mathcal{L}_{\text{feat}}+\mathcal{L}_{\text{Gen}},
\label{eq:full_loss}
\vspace{-5pt}
\end{equation}
where $\mathcal{L}_{\text{img}}$ is the loss in image space defined by
\vspace{-3pt}
\begin{equation}
\mathcal{L}_{\text{img}} = \|\x_{t+n}-\hat{\x}_{t+n}\|_2^2,
\label{eq:loss_img}
\vspace{-7pt}
\end{equation}

where $\x_{t+n}$ and $\hat{\x}_{t+n}$ are the target and predicted frames, respectively. The image loss intuitively guides our network towards a rough blurry pixel-leven frame prediction that reflects most details of the target image. $\mathcal{L}_{\text{feat}}$ is the loss in feature space define by
%
\begin{equation}
\begin{aligned}
\mathcal{L}_{\text{feat}}&= \|C_1\left(\x_{t+n}\right)-C_1\left(\hat{\x}_{t+n}\right)\|_{2}^{2}\\
&+\|C_2\left(\x_{t+n}\right)-C_2\left(\hat{\x}_{t+n}\right)\|_{2}^{2},
\end{aligned}
\label{eq:feat_loss}
\end{equation}
where $C_1\left(.\right)$ extracts features representing mostly image appearance, and $C_2\left(.\right)$ extracts features representing mostly image structure.
Combining appearance sensitive features with structure sensitive features gives our network a learning signal that allows it to make frame predictions with accurate appearance while also enforcing correct structure.
$\mathcal{L}_{\text{Gen}}$ is the term in adversarial loss that allows our model to generate realistic looking images and is defined by
%
\begin{equation}
\mathcal{L}_{\text{Gen}} = -\log D\left(\left[\p_{t+n},\hat{\x}_{t+n}\right]\right) ,
\label{eq:loss_adv}
\end{equation}
%
where $\hat{\x}_{t+n}$ is the predicted image, $\p_{t+n}$ is the human pose corresponding to the target image, and $D\left(.\right)$ is the discriminator network in adversarial loss.
This sub-loss allows our network to generate images that reflect a similar level of detail as the images observed in the training data.

During the optimization of $D$, we use the mismatch term proposed by \citet{reed2016generative}, which allows the discriminator $D$ to become sensitive to mismatch between the generation and the condition.
The discriminator loss is defined by
\begin{equation}\label{eq:loss_disc}
\begin{aligned}
\mathcal{L}_{\text{Disc}} &= -\log D\left(\left[\p_{t+n},\x_{t+n}\right]\right) \\
&-0.5\log\left(1-D\left(\left[\p_{t+n},\hat{\x}_{t+n}\right]\right)\right) \\
&-0.5\log\left(1-D\left(\left[\p_{t+n},\x_{t}\right]\right)\right),
\end{aligned}
\end{equation}
while optimizing our generator with respect to the adversarial loss, the mismatch-aware term sends a stronger signal to our generator resulting in higher quality image generation, and network optimization.
Essentially, having a discriminator that knows the correct structure-image relationship, reduces the parameter search space of our generator while optimizing to fool the discriminator into believing the generated image is real.
The latter in combination with the other loss terms allows our network to produce high quality image generation given the structure condition.

\cutsectionup
\section{Experiments} \label{sec:experiments}
\cutsectiondown

\begin{table*}[!hbt] 
\centering
\small
\setlength{\tabcolsep}{3pt}
\begin{tabular}{c|c|c|c}
\Xhline{4\arrayrulewidth}
Method & Temporal Stream & Spatial Stream & Combined \\
\Xhline{4\arrayrulewidth}
Real Test Data * & 66.6\% & 63.3\% & 72.1\% \\
\hline
Ours & \textbf{35.7}\% & \textbf{52.7}\% & \textbf{59.0}\% \\
Convolutional LSTM & 13.9\% & 45.1\% & 46.4\% \\
Optical Flow & 13.9\% & 39.2\% & 34.9\% \\
\Xhline{4\arrayrulewidth}
\end{tabular}
\vspace{-2pt}
\caption{Activity recognition evaluation.}
\label{table:activityrec}
\end{table*}

\begin{table*}[!hbt]
\centering
\small
\setlength{\tabcolsep}{3pt}
\begin{tabular}{c|cccccc|c}
\Xhline{4\arrayrulewidth}
"Which video is more realistic?" & Baseball & Clean \& jerk & Golf & Jumping jacks & Jump rope & Tennis & Mean\\
\Xhline{4\arrayrulewidth}
Prefers ours over Convolutional LSTM & 89.5\% & 87.2\%  & 84.7\% & 83.0\% & 66.7\% & 88.2\% & 82.4\%\\
Prefers ours over Optical Flow& 87.8\% & 86.5\% & 80.3\%  & 88.9\% & 86.2\% & 85.6\% & 86.1\%\\
\end{tabular}
\vspace{-2pt}
\caption{\textbf{Penn Action Video Generation Preference:} We show videos from two methods to Amazon Mechanical Turk workers and ask them to indicate which is more realistic. The table shows the percentage of times workers preferred our model against baselines. A majority of the time workers prefer predictions from our model. We merged baseball pitch and baseball swing into baseball, and tennis forehand and tennis serve into tennis.}
 \label{table:penn}
 \vspace{-10pt}
\end{table*}

In this section, we present experiments on pixel-level video prediction of human actions on the Penn Action \cite{pennaction} and Human 3.6M datasets \cite{human36m}.
Pose landmarks and video frames are normalized to be between -1 and 1, and frames are cropped based on temporal tubes to remove as much background as possible while making sure the human of interest is in all frames.
For the feature similarity loss term (Equation~\ref{eq:feat_loss}), we use we use the last convolutional layer in AlexNet \cite{alexnet} as $C_1$, and the last layer of the Hourglass Network in \citet{hourglass} as $C_2$.
We augmented the available video data by performing horizontal flips randomly at training time for Penn Action.
Motion-based pixel-level quantitative evaluation using Peak Signal-to-Noise Ratio (PSNR), analysis, and control experiments can be found in the supplementary material.
For video illustration of our method, please refer to the project website: \url{https://sites.google.com/a/umich.edu/rubenevillegas/hierch_vid}.

We compare our method against two baselines based on convolutional LSTM and optical flow.
The convolutional LSTM baseline \cite{convlstm} was trained with adversarial loss \cite{Mathieu16} and the feature similarity loss (Equation~\ref{eq:feat_loss}).
An optical flow based baseline used the last observed optical flow~\cite{farneback2003two} to move the pixels of the last observed frame into the future.

We follow a human psycho-physical quantitative evaluation metric similar to \citet{Vondrick16}.
Amazon Mechanical Turk (AMT) workers are given a two-alternative choice to indicate which of two videos looks more realistic.
Specifically, the workers are shown a pair of videos (generated by two different methods) consisting of the same input frames indicated by a green box and predicted frames indicated by a red box, in addition to the action label of the video.
The workers are instructed to make their decision based on the frames in the red box.
Additionally, we train a Two-stream action recognition network \cite{Simonyan14} on the Penn Action dataset and test on the generated videos to evaluate if our network is able to generate videos predicting the activities observed in the original dataset.
We do not perform action classification experiments on the Human3.6M dataset due to high uncertainty in the human movements and high motion similarity amongst actions.

\vspace{-8pt}
\paragraph{Architectures.}
The sequence prediction LSTM is made of a single layer encoder-decoder LSTM with tied parameters, 1024 hidden units, and tanh output activation.
Note that the decoder LSTM does not observe any inputs other than the hidden units from the encoder LSTM as initial hidden units.
The image and pose encoders are built with the same architecture as VGG16 \cite{Vgg16} up to the third pooling layer, except that the pose encoder takes in the pose heat-maps as an image made of $L$ channels, and the image encoder takes a regular $3$-channel image.
The decoder is the mirrored architecture of the image encoder where we perform unpooling followed by deconvolution, and a final tanh activation.
The convolutional LSTM baseline is built with the same architecture as the image encoder and decoder, but there is a convolutional LSTM layer with the same kernel size and number of channels as the last layer in the image encoder connecting them.

\cutsubsectionup
\subsection{Penn Action Dataset} \label{exp:penn}
\cutsubsectiondown
\paragraph{Experimental setting.}
The Penn Action dataset is composed of 2326 video sequences of 15 different actions and 13 human joint annotations for each sequence.
To train our image generator, we use the standard train split provided in the dataset.
To train our pose predictor, we sub-sample the actions in the standard train-test split due to very noisy joint ground-truth.
We used videos from the actions of baseball pitch, baseball swing, clean and jerk, golf swing, jumping jacks, jump rope, tennis forehand, and tennis serve.
Our pose predictor is trained to observe 10 inputs and predict 32 steps, and tested on predicting up to 64 steps (some videos' groundtruth end before 64 steps).
Our image generator is trained to make single random jumps within 30 steps into the future.
Our evaluations are performed on a single clip that starts at the first frame of each video.

\vspace{-8pt}
\paragraph{AMT results.}
These experiments were performed by 66 unique workers, where a total of 1848 comparisons were made (934 against convolutional LSTM and 914 against optical flow baseline).
As shown in Table~\ref{table:penn} and Figure~\ref{fig:qualitative}, our method is capable of generating more realistic sequences compared to the baselines.
Quantitatively, the action sequences generated by our network are perceptually higher quality than the baselines and also predict the correct action sequence. 
A relatively small (although still substantial) margin is observed when comparing to convolutional LSTM for the jump rope action (i.e., 66.7\% for ours vs 33.3\% for Convolutional LSTM).
We hypothesize that convolutional LSTM is able to do a reasonable job for this action class due the highly cyclic motion nature of jumping up and down in place.
The remainder of the human actions contain more complicated non-linear motion, which is much more complicated to predict.
Overall, our method outperforms the baselines by a large margin (i.e. 82.4\% for ours vs 17.6\% for Convolutional LSTM, and 86.1\% for ours vs 13.9\% for Optical Flow).
Side by side video comparison for all actions can be found in our \href{https://goo.gl/U7UOfy}{project website}.

\begin{table*}[!hbt]
\centering
\small
\setlength{\tabcolsep}{3pt}
\begin{tabular}{c|ccccccc}
\Xhline{4\arrayrulewidth}
"Which video is more realistic?" & Directions & Discussion & Eating & Greeting & Phoning & Photo & Posing \\
\Xhline{4\arrayrulewidth}
Prefers ours over Convolutional LSTM & 67.6\% & 75.9\% & 74.7\% & 79.5\% & 69.7\% & 66.2\% & 69.7\%\\
Prefers ours over Optical Flow & 61.4\% & 89.3\% & 43.8\% & 80.3\% & 84.5\% & 52.0\% & 75.3\%\\
\end{tabular}
\setlength{\tabcolsep}{3pt}
\begin{tabular}{c|cccccc|c}
\Xhline{4\arrayrulewidth}
"Which video is more realistic?" & Purchases & Sitting & Sittingdown & Smoking & Waiting & Walking & Mean \\
\Xhline{4\arrayrulewidth}
Prefers ours over Convolutional LSTM & 79.0\% & 38.0\% & 54.7\% & 70.4\% & 50.0\% & 86.0\% & 70.3\% \\
Prefers ours over Optical Flow & 85.7\% & 35.1\% & 46.7\% & 73.3\% & 84.3\% & 90.8\% & 72.3\%\\
\end{tabular}
\caption{\textbf{Human3.6M Video Generation Preference:} We show videos from two methods to Amazon Mechanical Turk workers and ask them to indicate which of the the two looks more realistic. The table shows the percentage of times workers preferred our model against baselines. Most of the time workers prefer predictions from our model. We merge the activity categories of walking, walking dog, and walking together into walking.}
 \label{table:human}
 \vspace{-10pt}
\end{table*}

\vspace{-8pt}
\paragraph{Action recognition results.} 
To see whether the generated videos contain actions that can fool a CNN trained for action recognition, we train a Two-Stream CNN on the PennAction dataset.
In Table~\ref{table:activityrec}, ``Temporal Stream'' denotes the network that observes motion as concatenated optical flow (Farneback's optical flow) images as input, and ``Spatial Stream'' denotes the network that observes single image as input.
``Combined'' denotes the averaging of the output probability vectors from the Temporal and Spatial stream.
``Real test data'' denotes evaluation on the ground-truth videos (i.e. perfect prediction).

From Table~\ref{table:activityrec}, it is shown that our network is able to generate videos that are far more representative of the correct action compared to all baselines, in both Temporal and Spatial stream, regardless of using a neural network as the judge.
When combining both Temporal and Spatial streams, our network achieves the best quality videos in terms of making a pixel-level prediction of the correct action.

\vspace{-8pt}
\paragraph{Pixel-level evaluation and control experiments.} \label{exp:penn_pixel}
We evaluate the frames generated by our method using PSNR as measure, and separated the test data based on amount of motion, as suggested by \citet{Villegas17}.
From these experiments, we conclude that pixel-level evaluation highly depends on predicting the exact future observed in the ground-truth.
Highest PSNR scores are achieved when trajectories of the exact future is used to generate the future frames.
Due to space constraints, we ask the reader to please refer to the supplementary material for more detailed quantitative and qualitative analysis.

\cutsubsectionup
\subsection{Human3.6M Dataset}\label{sec:experiments_h36m}
\cutsubsectiondown
\paragraph{Experimental settings.}
The Human3.6M dataset \cite{human36m} is composed of 3.6 million 3D human poses (we use the provided 2D pose projections) composed of 32 joints and corresponding images taken from 11 professional actors in 17 scenarios.
For training, we use subjects number 1, 5, 6, 7, and 8, and test on subjects number 9 and 11.
Our pose predictor is trained to observe 10 inputs and predict 64 steps, and tested on predicting 128 steps.
Our image generator is trained to make single random jumps anywhere in the training videos.
We evaluate on a single clip from each test video that starts at the exact middle of the video to make sure there is motion occurring.

\begin{figure*}[thbp]
    \vspace{-6pt}
    \centering
	\begin{subfigure}{0.04\linewidth}
        \raggedleft
        \rotatebox{90}{
        \hspace{-.3cm}
        \parbox{2cm}{\centering Input frames} \parbox{2cm}{\centering Groundtruth} \parbox{2cm}{\centering Conv LSTM} \parbox{2cm}{\centering Predicted frames (ours)} \parbox{2cm}{\centering Predicted pose (ours)}
        }
    \end{subfigure}
    \begin{subfigure}{0.12\linewidth}
        \caption*{t=11}
        \vspace{-7pt}
	    \includegraphics[width=1\linewidth]{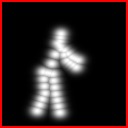} 
  		\includegraphics[width=1\linewidth]{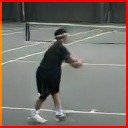}
  		\includegraphics[width=1\linewidth]{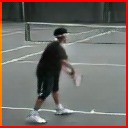}
  		\includegraphics[width=1\linewidth]{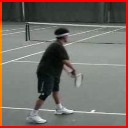}
  		\includegraphics[width=1\linewidth]{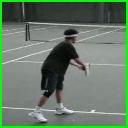}
	\end{subfigure} 
    \begin{subfigure}{0.12\linewidth}
        \caption*{t=20}
        \vspace{-7pt}
	    \includegraphics[width=1\linewidth]{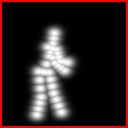} 
  		\includegraphics[width=1\linewidth]{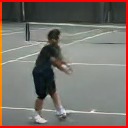}
  		\includegraphics[width=1\linewidth]{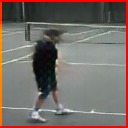}
  		\includegraphics[width=1\linewidth]{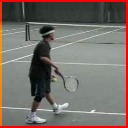}
  		\includegraphics[width=1\linewidth]{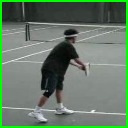}
	\end{subfigure} 
    \begin{subfigure}{0.12\linewidth}
        \caption*{t=29}
        \vspace{-7pt}
	    \includegraphics[width=1\linewidth]{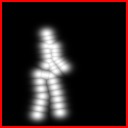} 
  		\includegraphics[width=1\linewidth]{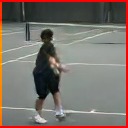}
  		\includegraphics[width=1\linewidth]{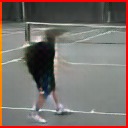}
  		\includegraphics[width=1\linewidth]{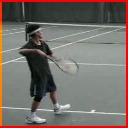}
  		\includegraphics[width=1\linewidth]{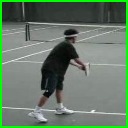}
	\end{subfigure} 
    \begin{subfigure}{0.12\linewidth}
        \caption*{t=38}
        \vspace{-7pt}
	    \includegraphics[width=1\linewidth]{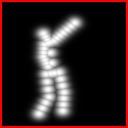} 
  		\includegraphics[width=1\linewidth]{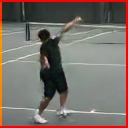}
  		\includegraphics[width=1\linewidth]{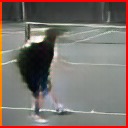}
  		\includegraphics[width=1\linewidth]{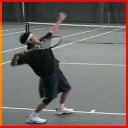}
  		\includegraphics[width=1\linewidth]{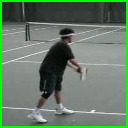}
	\end{subfigure}
	\begin{subfigure}{0.12\linewidth}
        \caption*{t=47}
        \vspace{-7pt}
	    \includegraphics[width=1\linewidth]{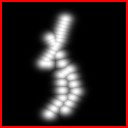} 
  		\includegraphics[width=1\linewidth]{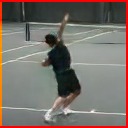}
  		\includegraphics[width=1\linewidth]{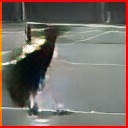}
  		\includegraphics[width=1\linewidth]{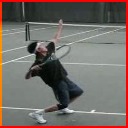}
  		\includegraphics[width=1\linewidth]{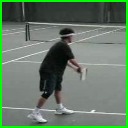}
	\end{subfigure}
	\begin{subfigure}{0.12\linewidth}
        \caption*{t=56}
        \vspace{-7pt}
	    \includegraphics[width=1\linewidth]{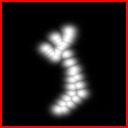} 
  		\includegraphics[width=1\linewidth]{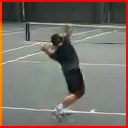}
  		\includegraphics[width=1\linewidth]{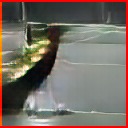}
  		\includegraphics[width=1\linewidth]{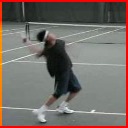}
  		\includegraphics[width=1\linewidth]{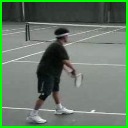}
	\end{subfigure}
	\begin{subfigure}{0.12\linewidth}
        \caption*{t=65}
        \vspace{-7pt}
	    \includegraphics[width=1\linewidth]{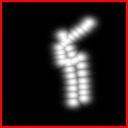} 
  		\includegraphics[width=1\linewidth]{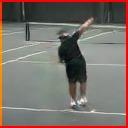}
  		\includegraphics[width=1\linewidth]{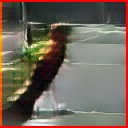}
  		\includegraphics[width=1\linewidth]{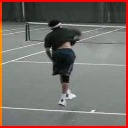}
  		\includegraphics[width=1\linewidth]{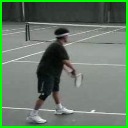}
	\end{subfigure}
    \vspace{.1cm}\\
    \centering

	\begin{subfigure}{0.04\linewidth}
        \raggedleft
        \rotatebox{90}{
        \hspace{-.3cm}
        \parbox{2cm}{\centering Input frames} \parbox{2cm}{\centering Groundtruth} \parbox{2cm}{\centering Conv LSTM} \parbox{2cm}{\centering Predicted frames (ours)} \parbox{2cm}{\centering Predicted pose (ours)}
        }
    \end{subfigure}
    \begin{subfigure}{0.12\linewidth}
        \caption*{t=11}
        \vspace{-7pt}
	    \includegraphics[width=1\linewidth]{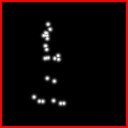} 
  		\includegraphics[width=1\linewidth]{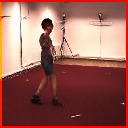}
  		\includegraphics[width=1\linewidth]{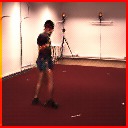}
  		\includegraphics[width=1\linewidth]{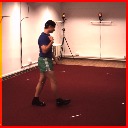}
  		\includegraphics[width=1\linewidth]{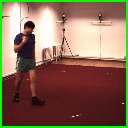}
	\end{subfigure} 
    \begin{subfigure}{0.12\linewidth}
        \caption*{t=29}
        \vspace{-7pt}
	    \includegraphics[width=1\linewidth]{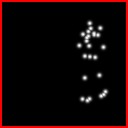} 
  		\includegraphics[width=1\linewidth]{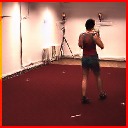}
  		\includegraphics[width=1\linewidth]{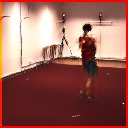}
  		\includegraphics[width=1\linewidth]{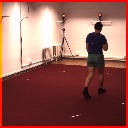}
  		\includegraphics[width=1\linewidth]{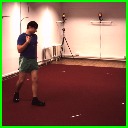}
	\end{subfigure} 
    \begin{subfigure}{0.12\linewidth}
        \caption*{t=47}
        \vspace{-7pt}
	    \includegraphics[width=1\linewidth]{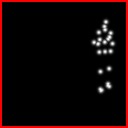} 
  		\includegraphics[width=1\linewidth]{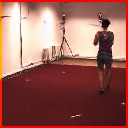}
  		\includegraphics[width=1\linewidth]{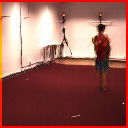}
  		\includegraphics[width=1\linewidth]{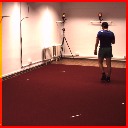}
  		\includegraphics[width=1\linewidth]{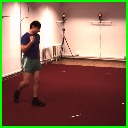}
	\end{subfigure} 
    \begin{subfigure}{0.12\linewidth}
        \caption*{t=65}
        \vspace{-7pt}
	    \includegraphics[width=1\linewidth]{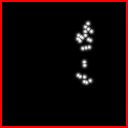} 
  		\includegraphics[width=1\linewidth]{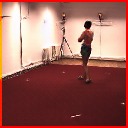}
  		\includegraphics[width=1\linewidth]{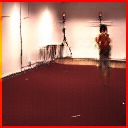}
  		\includegraphics[width=1\linewidth]{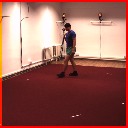}
  		\includegraphics[width=1\linewidth]{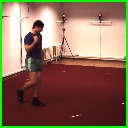}
	\end{subfigure}
	\begin{subfigure}{0.12\linewidth}
        \caption*{t=83}
        \vspace{-7pt}
	    \includegraphics[width=1\linewidth]{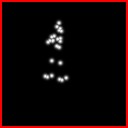} 
  		\includegraphics[width=1\linewidth]{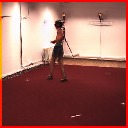}
  		\includegraphics[width=1\linewidth]{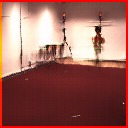}
  		\includegraphics[width=1\linewidth]{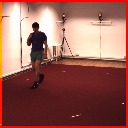}
  		\includegraphics[width=1\linewidth]{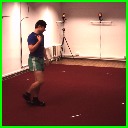}
	\end{subfigure}
	\begin{subfigure}{0.12\linewidth}
        \caption*{t=101}
        \vspace{-7pt}
	    \includegraphics[width=1\linewidth]{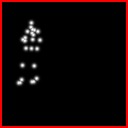} 
  		\includegraphics[width=1\linewidth]{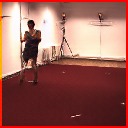}
  		\includegraphics[width=1\linewidth]{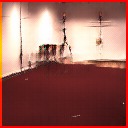}
  		\includegraphics[width=1\linewidth]{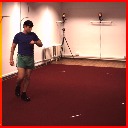}
  		\includegraphics[width=1\linewidth]{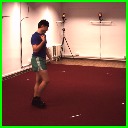}
	\end{subfigure}
	\begin{subfigure}{0.12\linewidth}
        \caption*{t=119}
        \vspace{-7pt}
	    \includegraphics[width=1\linewidth]{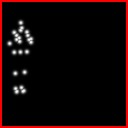} 
  		\includegraphics[width=1\linewidth]{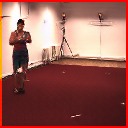}
  		\includegraphics[width=1\linewidth]{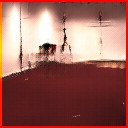}
  		\includegraphics[width=1\linewidth]{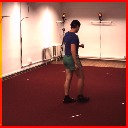}
  		\includegraphics[width=1\linewidth]{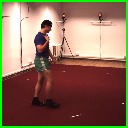}
	\end{subfigure}
    \vspace{-5pt}

    \caption{Qualitative evaluation of our network for 55 step prediction on Penn Action (top rows), and 109 step prediction on Human3.6M (bottom rows). Our algorithm observes 10 previous input frames, estimates the human pose, predicts the pose sequence of the future, and it finally generates the future frames. Green box denotes input and red box denotes prediction. We show the last 7 input frames. Side by side video comparisons can be found in our \href{https://goo.gl/U7UOfy}{project website}.}
\label{fig:qualitative}
\end{figure*}

\vspace{-8pt}
\paragraph{AMT results.}
We collected a total of 2203 comparisons (1086 against convolutional LSTM and 1117 against optical flow baseline) from 71 unique workers.
As shown in Table~\ref{table:human}, the videos generated by our network are perceptually higher quality and reflect a reasonable future compared to the baselines on average.
Unexpectedly, our network does not perform well on videos where the action involves minimal motion,  such as sitting,  sitting down, eating, taking a photo, and waiting.
These actions usually involve the person staying still or making very unnoticeable motion which can result in a static prediction (by convolutional LSTM and/or optical flow) making frames look far more realistic than the prediction from our network.
Overall, our method outperforms the baselines by a large margin (i.e. 70.3\% for ours vs 29.7\% for Convolutional LSTM, and 72.3\% for ours vs 27.7\% for Optical Flow).
Figure~\ref{fig:qualitative} shows that our network generates far higher quality future frames compared to the convolutional LSTM baseline.
Side by side video comparison for all actions can be found in our \href{https://goo.gl/U7UOfy}{project website}.

\vspace{-8pt}
\paragraph{Pixel-level evaluation and control experiments.}  \label{exp:h36m_pixel}
Following the same procedure as Section~\ref{exp:penn_pixel}, we evaluated the predicted videos using PSNR and separated the test data by motion.
Due to the high uncertainty and number of prediction steps in these videos, the predicted future can largely deviate from the exact future observed in the ground-truth.
The highest PSNR scores are again achieved when the exact future pose is used to generate the video frames; however, there is an even larger gap compared to the results in Section~\ref{exp:penn}.
Due to space constraints, we ask the reader to please refer to the supplementary material for more detailed quantitative and qualitative analysis.

\cutsectionup
\section{Conclusion and Future Work}\label{sec:conclusion}
\cutsectiondown
We propose a hierarchical approach of pixel-level video prediction.
Using human action videos as benchmark, we have demonstrated that our hierarchical prediction approach is able to predict up to 128 future frames, which is an order of magnitude improvement in terms of effective temporal scale of the prediction. 

\vspace{-3pt}
The success of our approach demonstrates that it can be greatly beneficial to incorporate the proper high-level structure into the generative process.
At the same time, an important open research question would be how to automatically learn such structures without domain knowledge. We leave this as future work.

\vspace{-3pt}
Another limitation of this work is that it generates a single future trajectory.
For an agent to make a better estimation of what the future looks like, we would need more than one generated future.
Future work will involve the generation of many futures given using a probabilistic sequence model.

\vspace{-3pt}
Finally, our model does not handle background motion.
This is a highly challenging task since background comes in and out of sight.
Predicting background motion will require a generative model that hallucinates the unseen background.
We also leave this as future work.


\cutsectionup
\section*{Acknowledgments}
\cutsectiondown
This work was supported in part by ONR N00014-13-1-0762, NSF CAREER
IIS-1453651, Gift from Bosch Research, and Sloan Research Fellowship.
We thank NVIDIA for donating K40c and TITAN X GPUs.


\bibliography{submission}
\bibliographystyle{icml2017}

\clearpage
\newpage

\onecolumn
\section*{\fontsize{15}{18}\selectfont Appendix}
\begin{appendix}
\section{Motion-Based Pixel-Level Evaluation, Analysis, and Control Experiments}
\cutsubsectiondown
In this section, we evaluate the predictions by deciles of motion similar to \citet{Villegas17} using Peak Signal-to-Noise Ratio (PSNR) measure, where the 10$^{\text{th}}$ decile contains videos with the most overall motion.
We add a modification to our hierarchical method based on a simple heuristic by which we copy the background pixels from the last observed frame using the predicted pose heat-maps as foreground/background masks (\texttt{Ours BG}).
Additionally, we perform experiments based on an \emph{oracle} that provides our image generator the exact future pose trajectories (\texttt{Ours GT-pose$^*$}) and we also apply the previously mentioned heuristics (\texttt{Ours GT-pose BG$^*$}). We put * marks to clarify that these are \emph{hypothetical} methods as they require ground-truth future pose trajectories. 

In our method, the future frames are strictly dictated by the future structure.
Therefore, the prediction based on the future pose oracle sheds light on how much predicting a different future structure affects PSNR scores. 
(Note: many future trajectories are possible given a single past trajectory.)
Further, we show that our conditional image generator given the perfect knowledge of the future pose trajectory (e.g., \texttt{Ours GT-pose$^*$}) produces high-quality video prediction that both matches the ground-truth video closely and achieves much higher PNSRs. 
These results suggest that our hierarchical approach is a step in the right direction towards solving the problem of long-term pixel-level video prediction.

\cutsubsectionup
\subsection{Penn Action}\label{supp:penn}
\begin{figure}[htb!]
\centering
\includegraphics[width=0.40\linewidth] {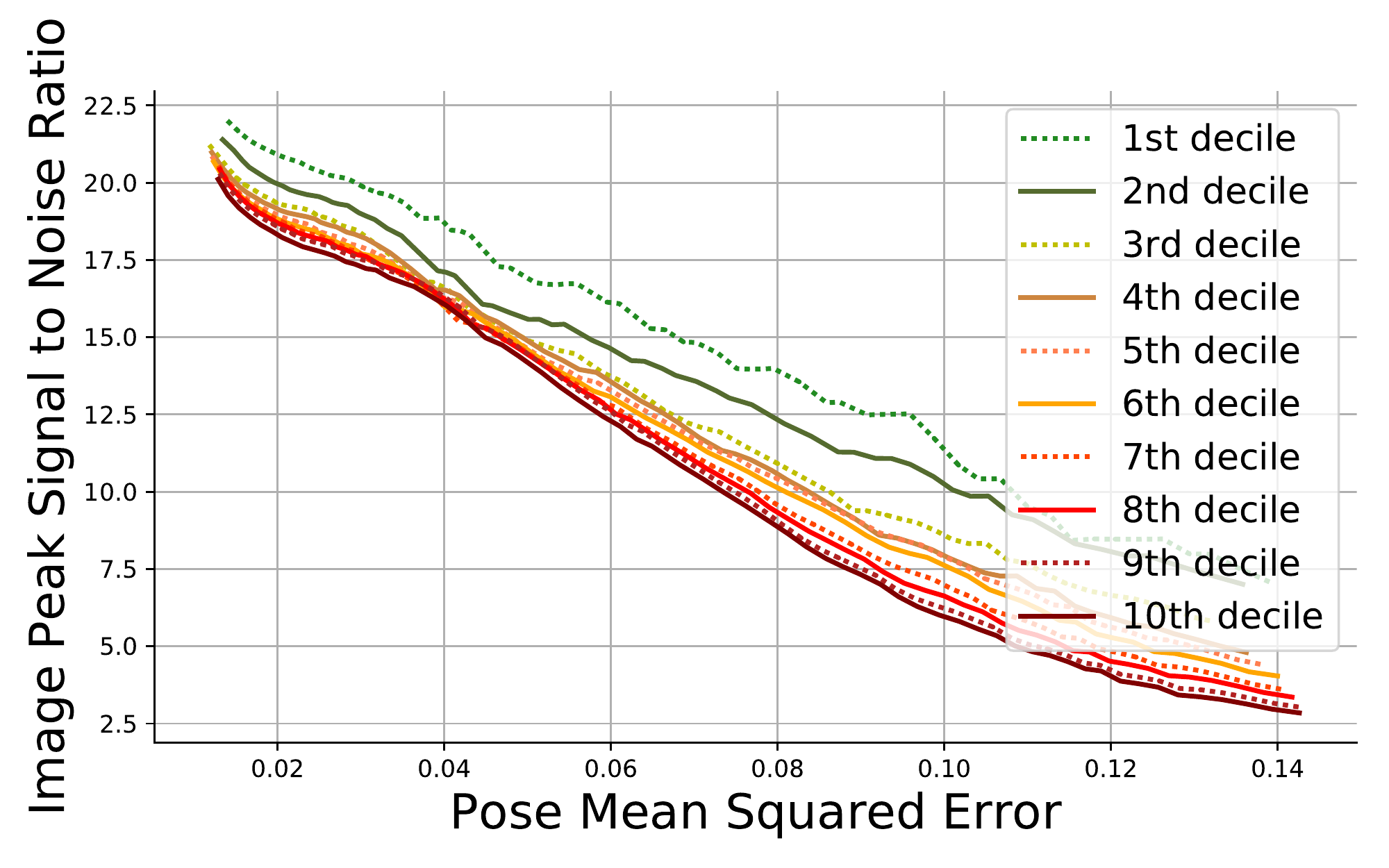}
\vspace{-.3cm}
\caption{Predicted frames PSNR vs. Mean Squared Error on the predicted pose for each motion decile in Penn Action.}
\label{fig:penn_corr}
\end{figure}

\cutsubsectiondown
In Figures~\ref{fig:penn_motion1}, and \ref{fig:penn_motion2}, we show evaluation on each decile of motion.
The plots show that our method outperforms the baselines for long-term frame prediction.
In addition, by using the future pose determined by the oracle as input to our conditional image generator, our method can achieve even higher PSNR scores.
We hypothesize that predicting future frames that reflect similar action semantics as the ground-truth, but with possibly different pose trajectories, causes lower PSNR scores.
Figure~\ref{fig:penn_corr} supports this hypothesis by showing that higher MSE in predicted pose tends to correspond to lower PSNR score.

\begin{figure}[htb!]
\centering
\vspace{30pt}
\includegraphics[width=0.40\linewidth] {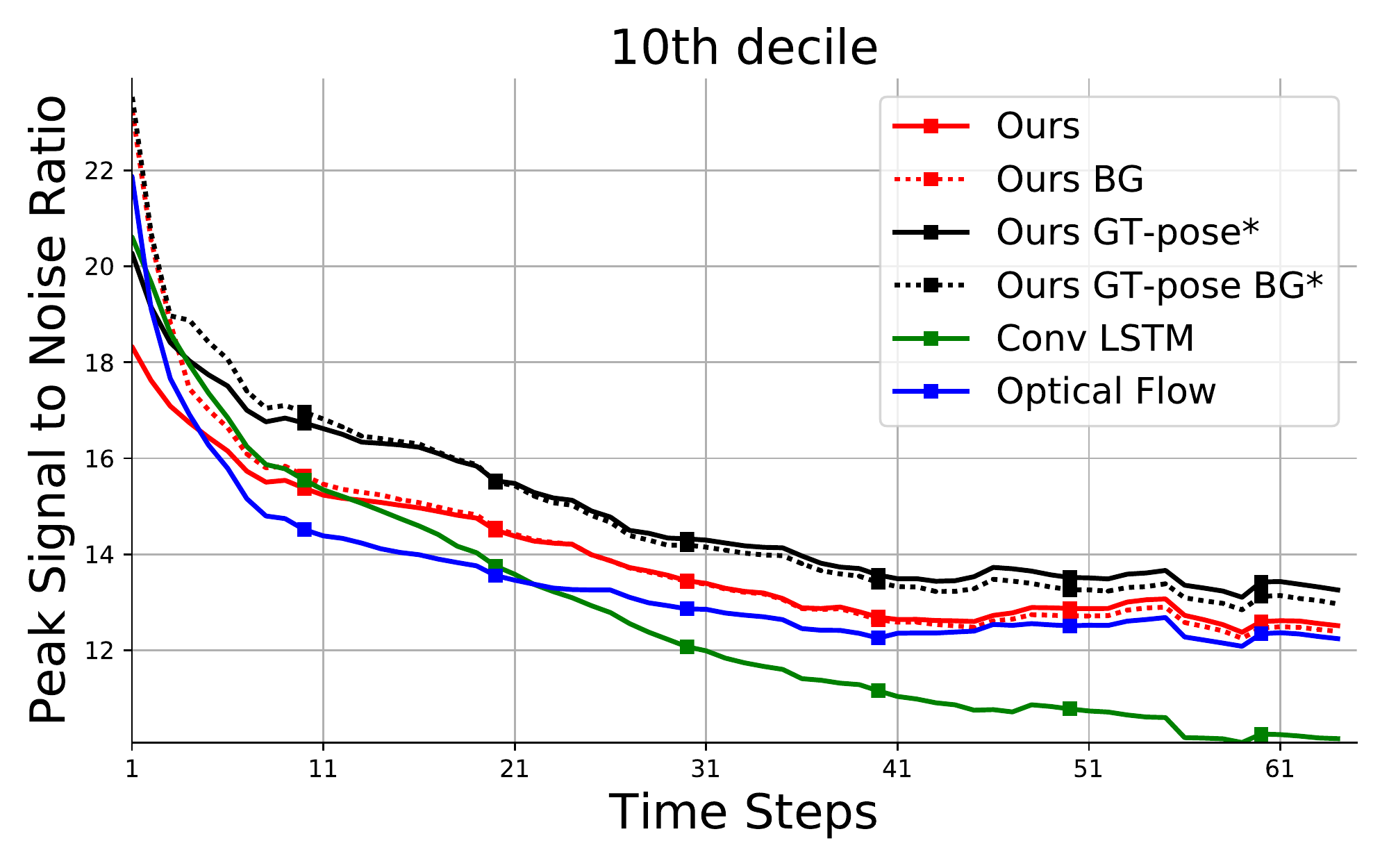}
\hspace{.8cm}
\includegraphics[width=0.40\linewidth] {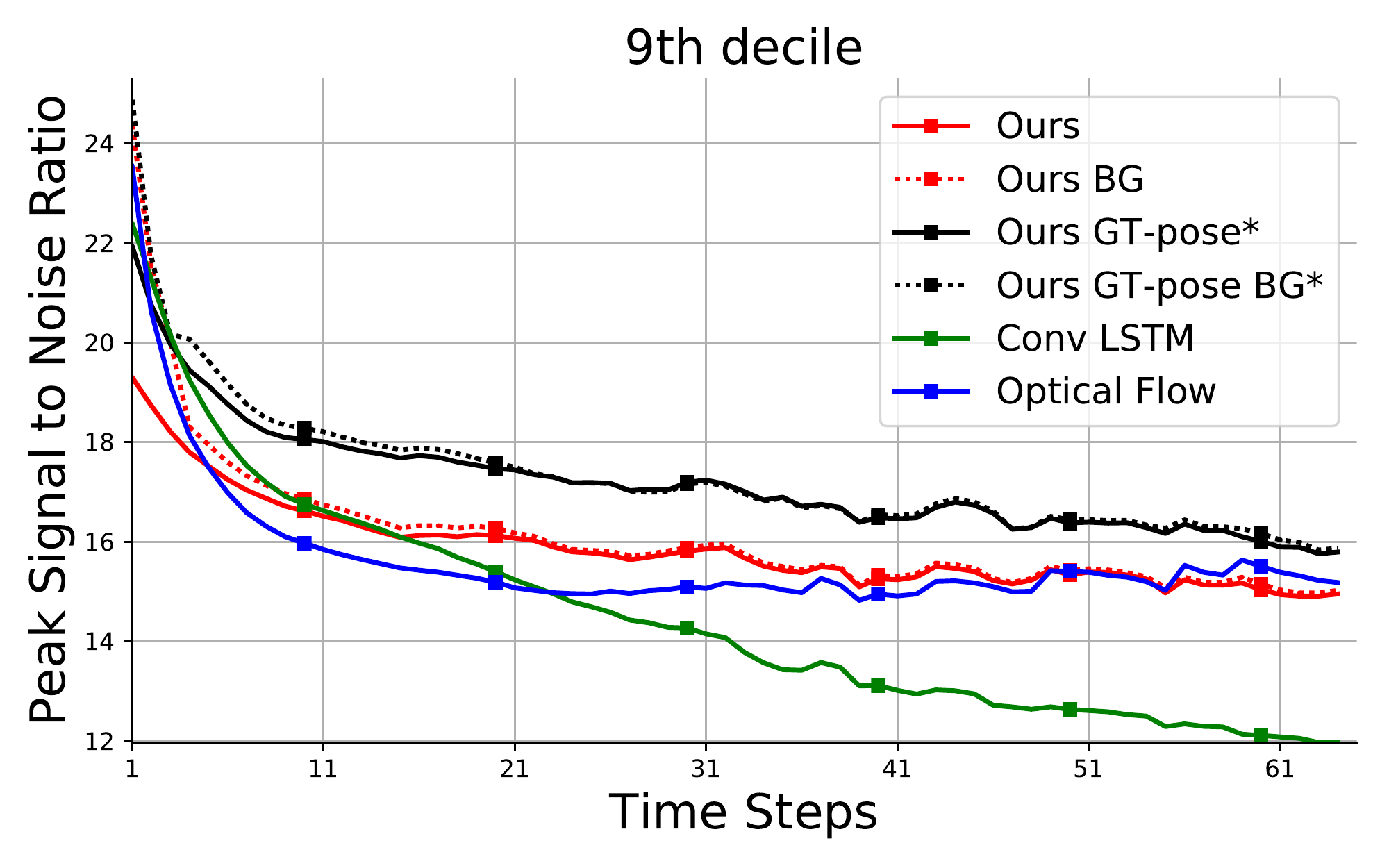} \\
\includegraphics[width=0.40\linewidth] {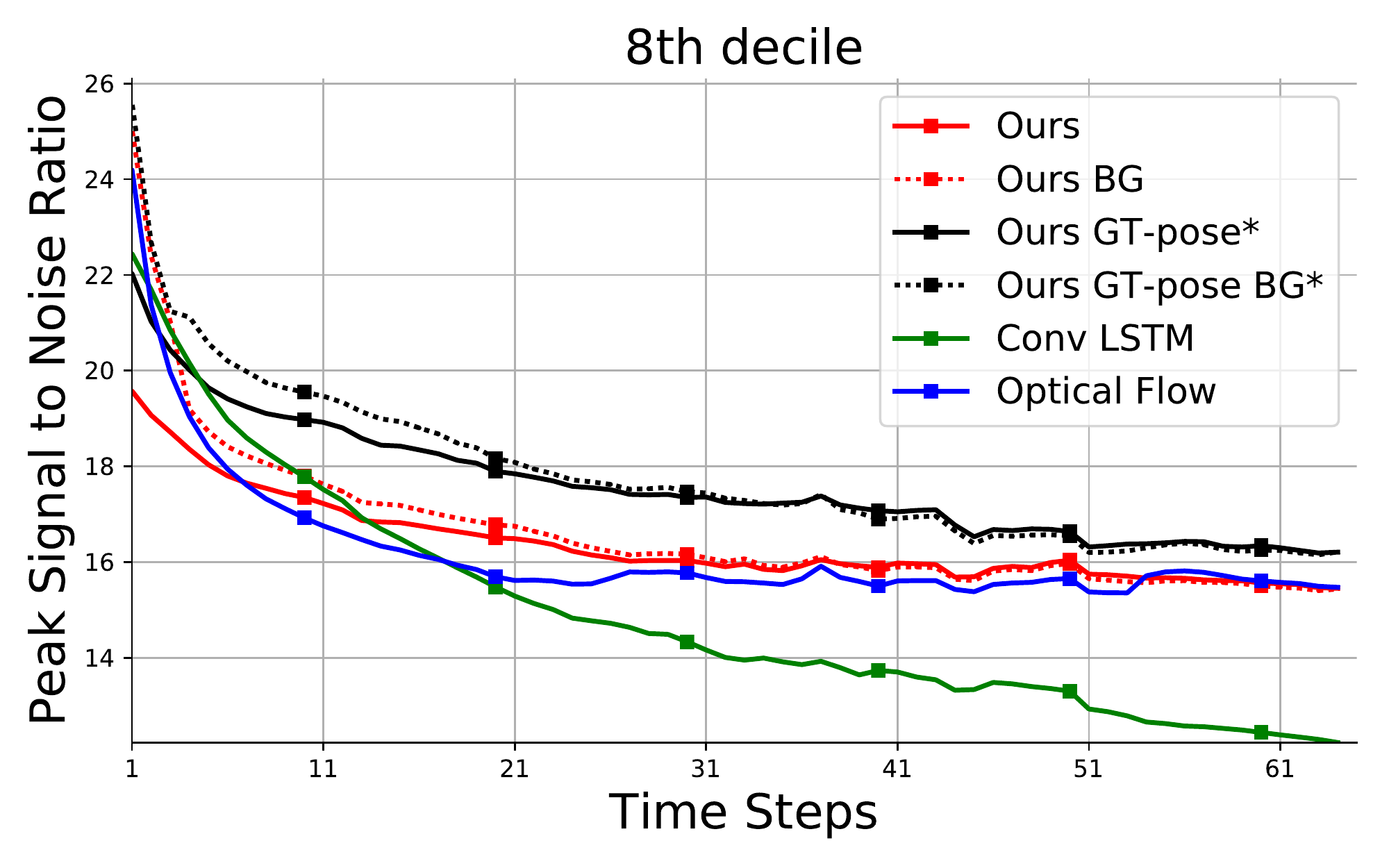}
\hspace{.8cm}
\includegraphics[width=0.40\linewidth] {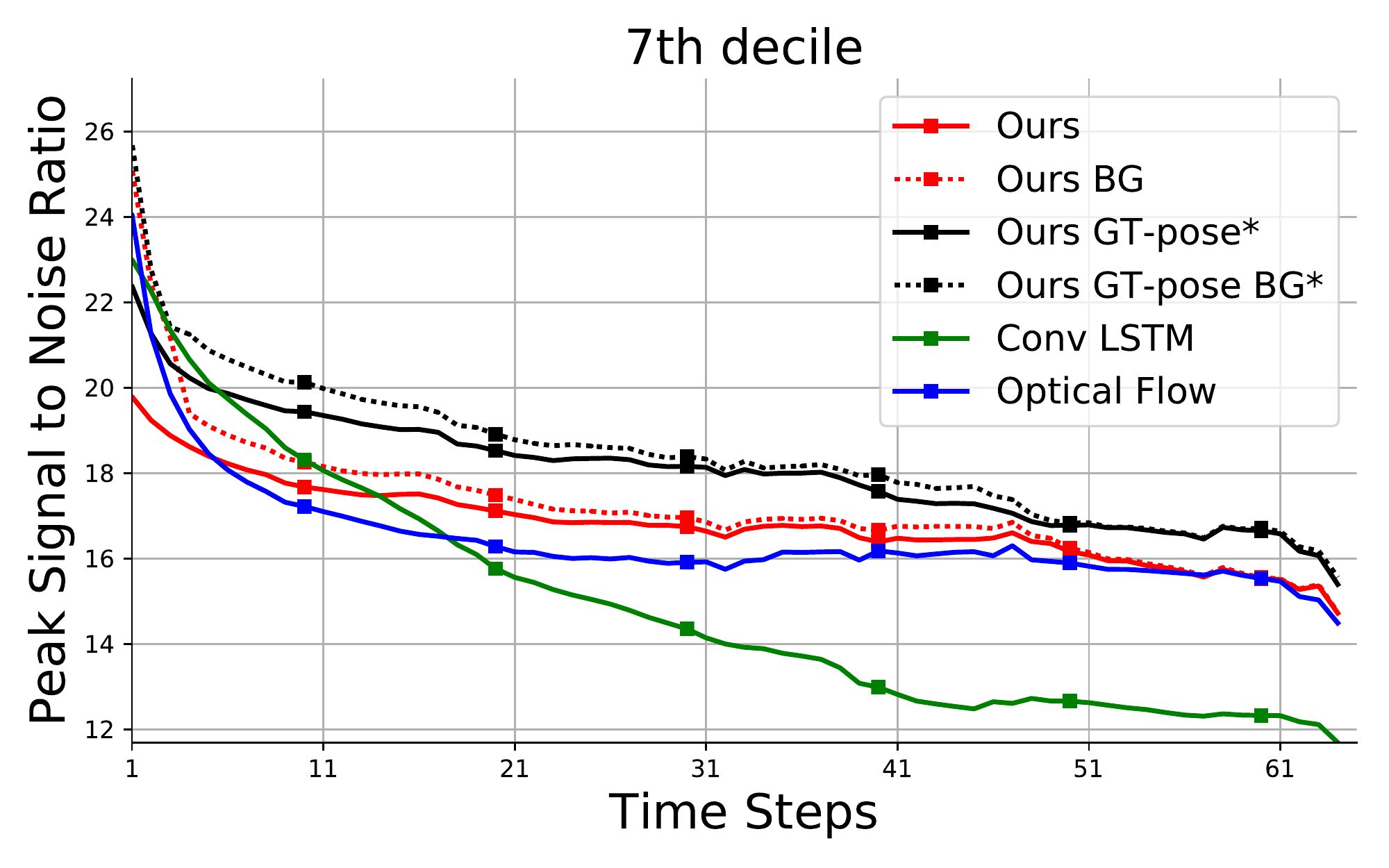} \\
\caption{Quantitative comparison on Penn Action separated by motion decile.}
\label{fig:penn_motion1}
\vspace{-2cm}
\end{figure}

\clearpage
\begin{figure}[htb!]
\vspace{30pt}
\centering
\includegraphics[width=0.40\linewidth] {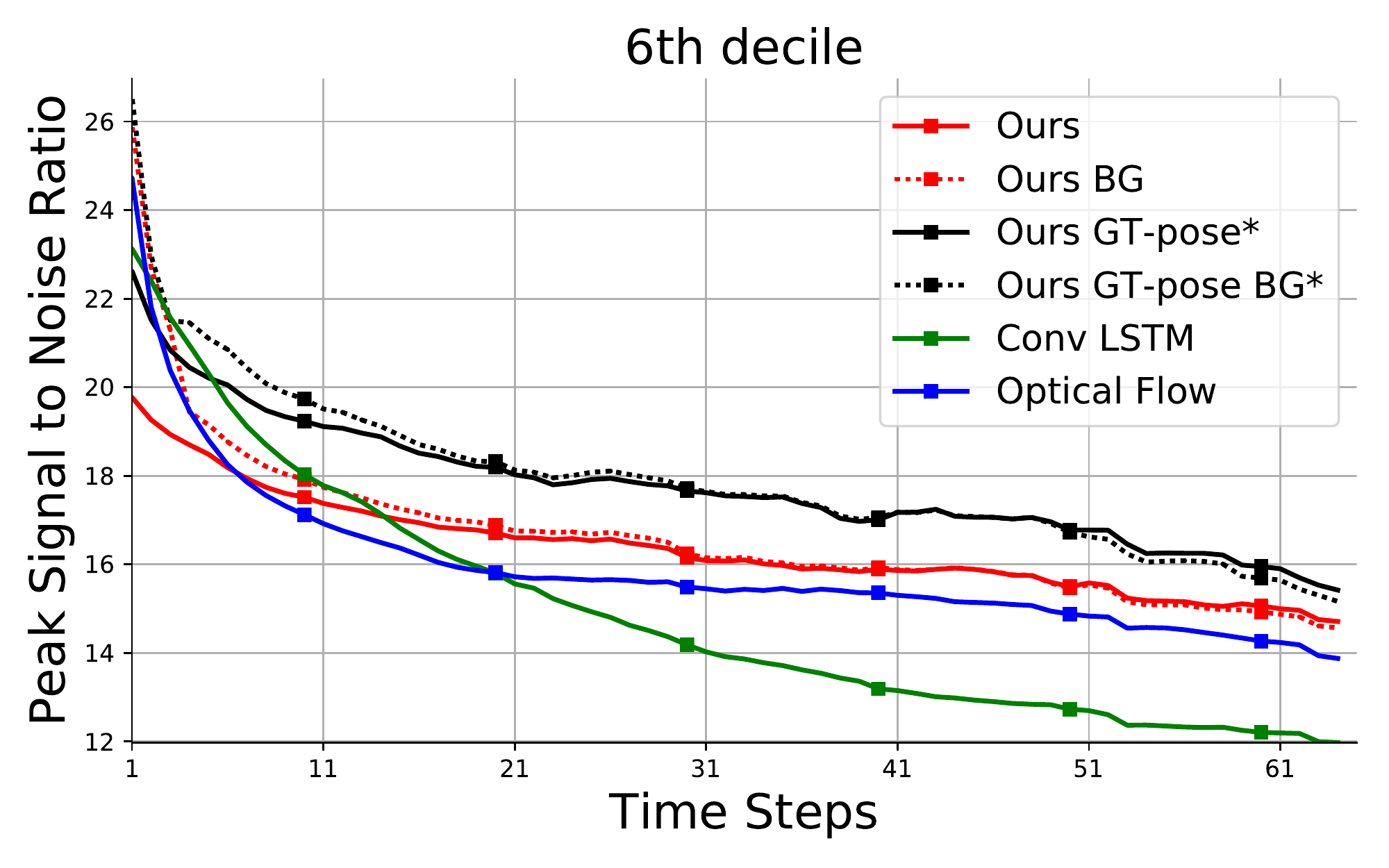}
\hspace{.8cm}
\includegraphics[width=0.40\linewidth] {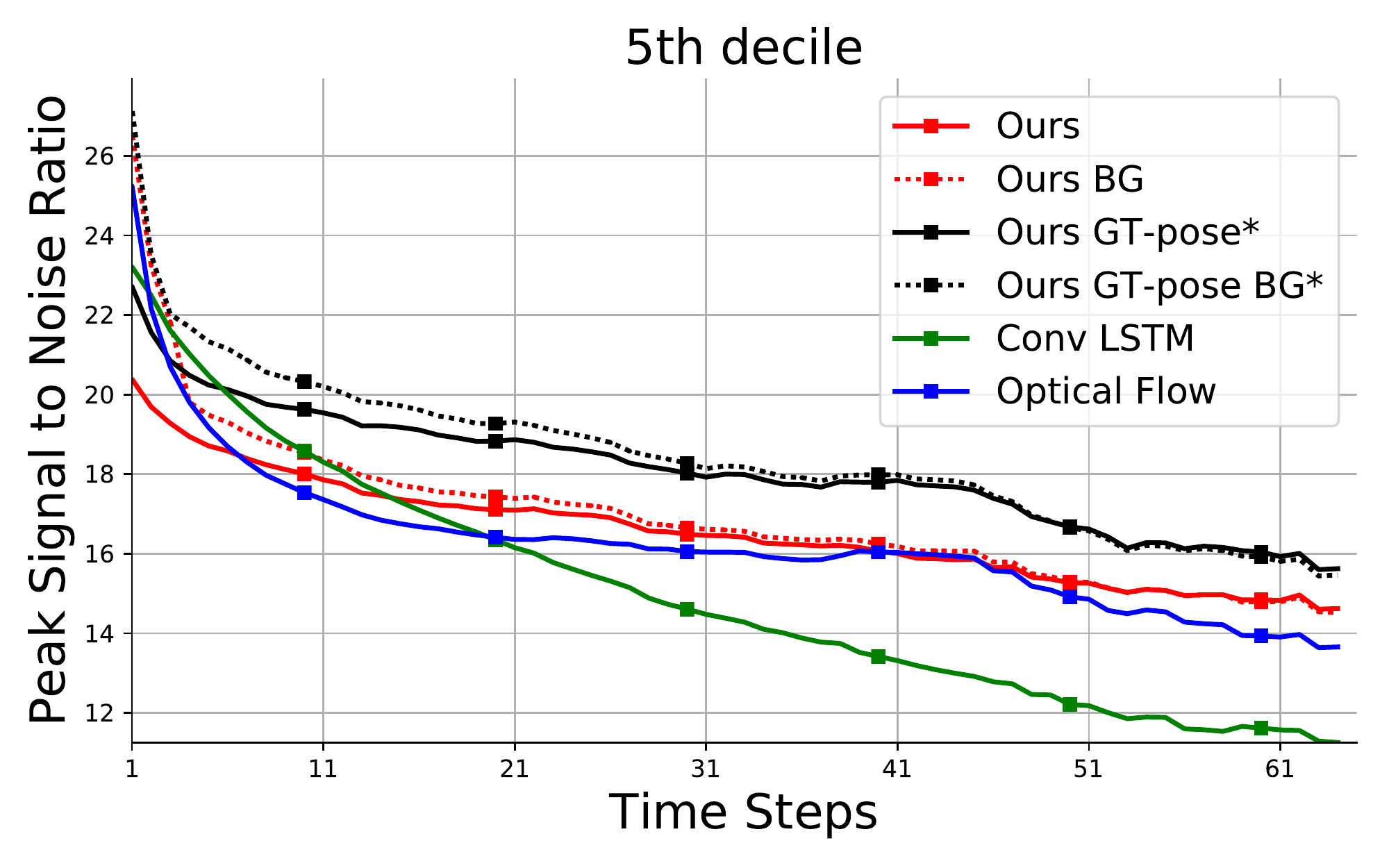}\\
\includegraphics[width=0.40\linewidth] {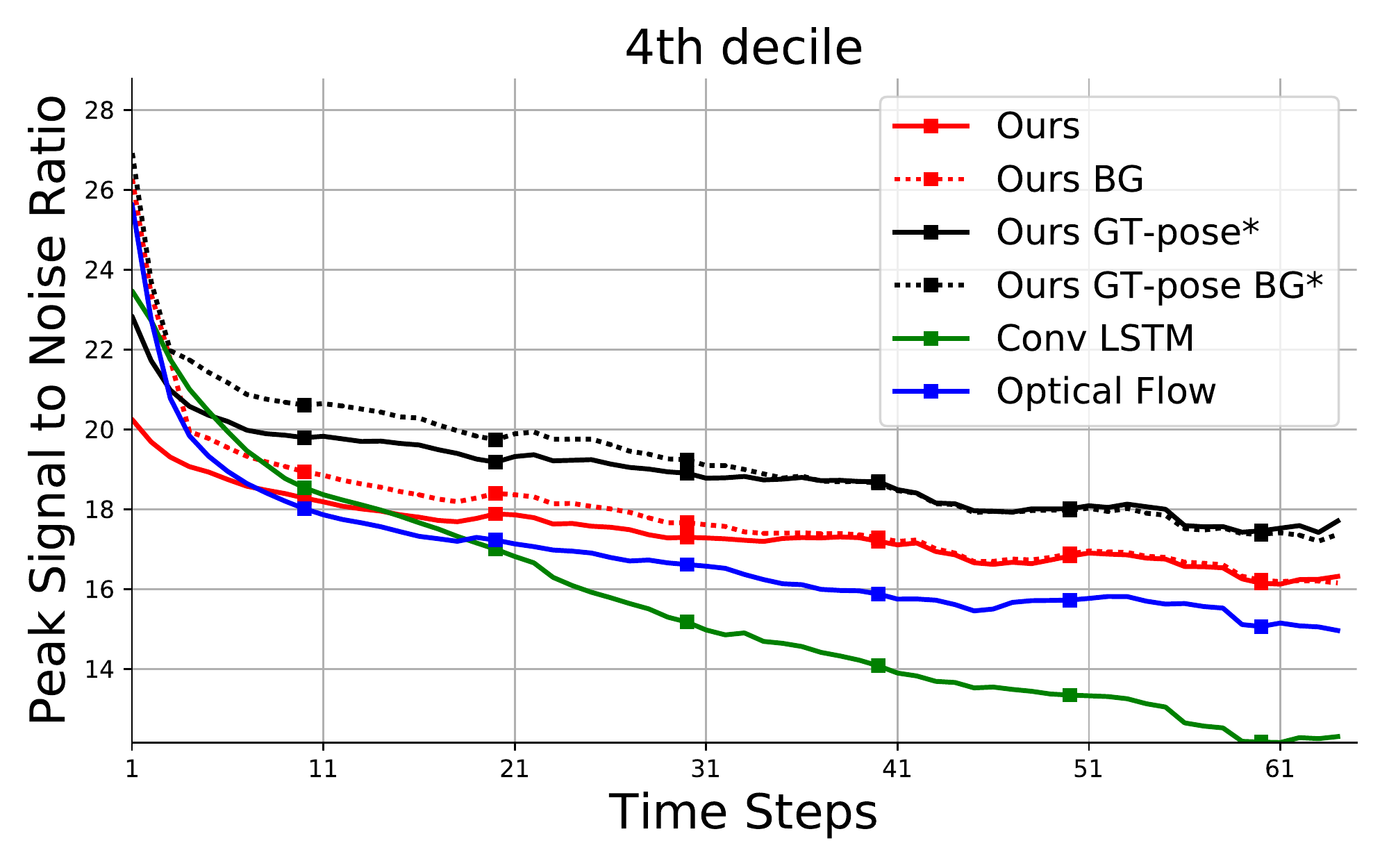}
\hspace{.8cm}
\includegraphics[width=0.40\linewidth] {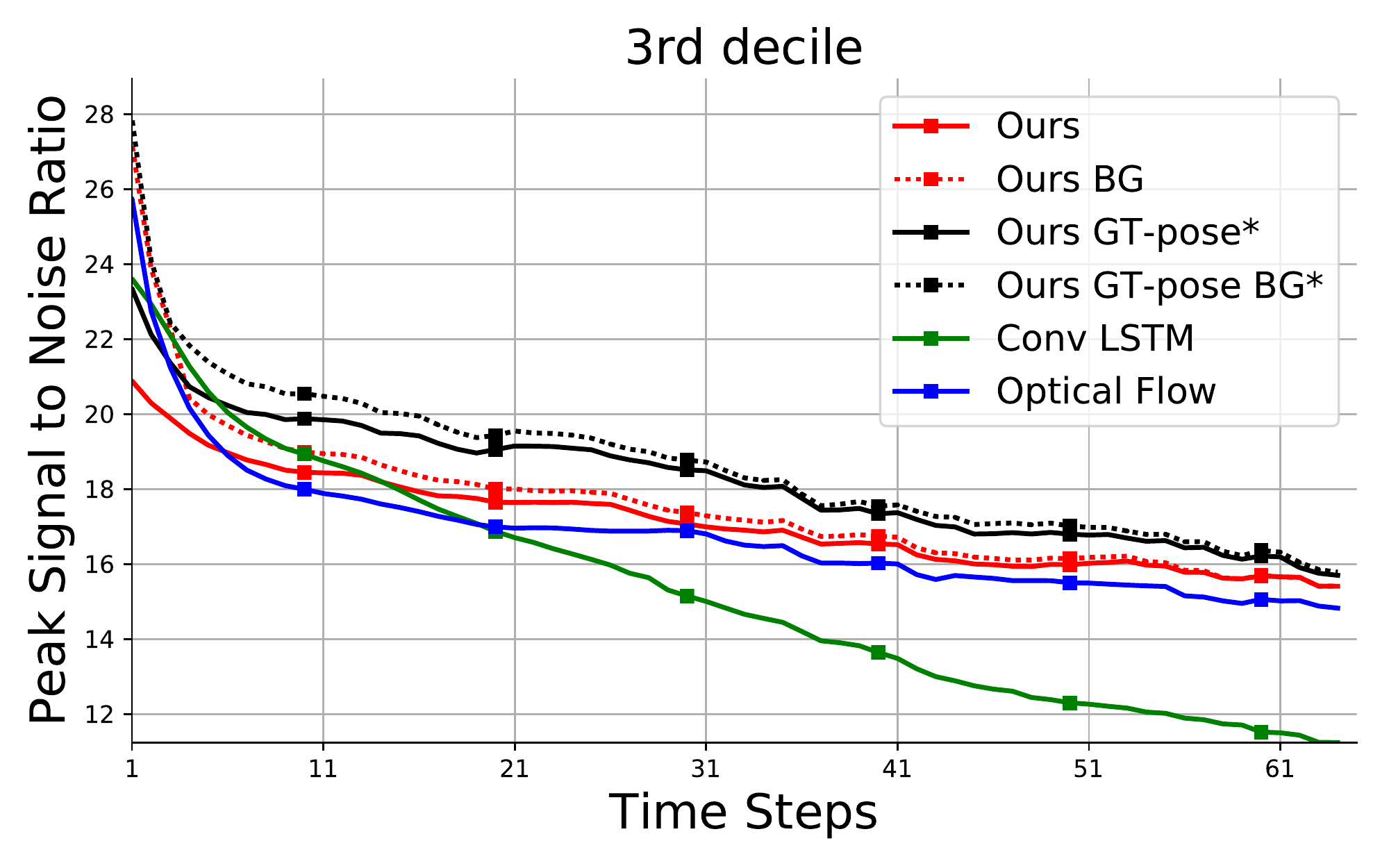}\\
\includegraphics[width=0.40\linewidth] {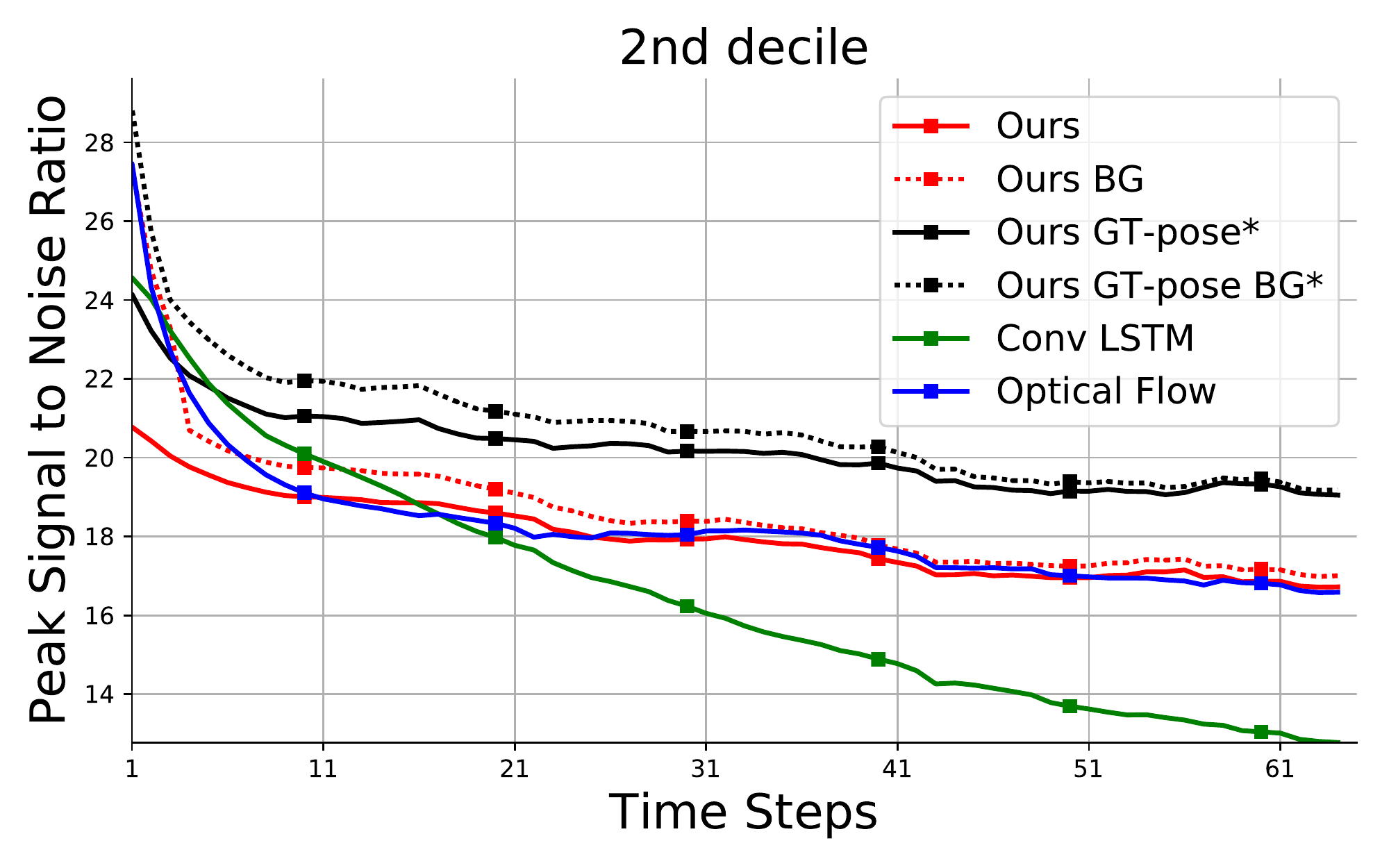}
\hspace{.8cm}
\includegraphics[width=0.40\linewidth] {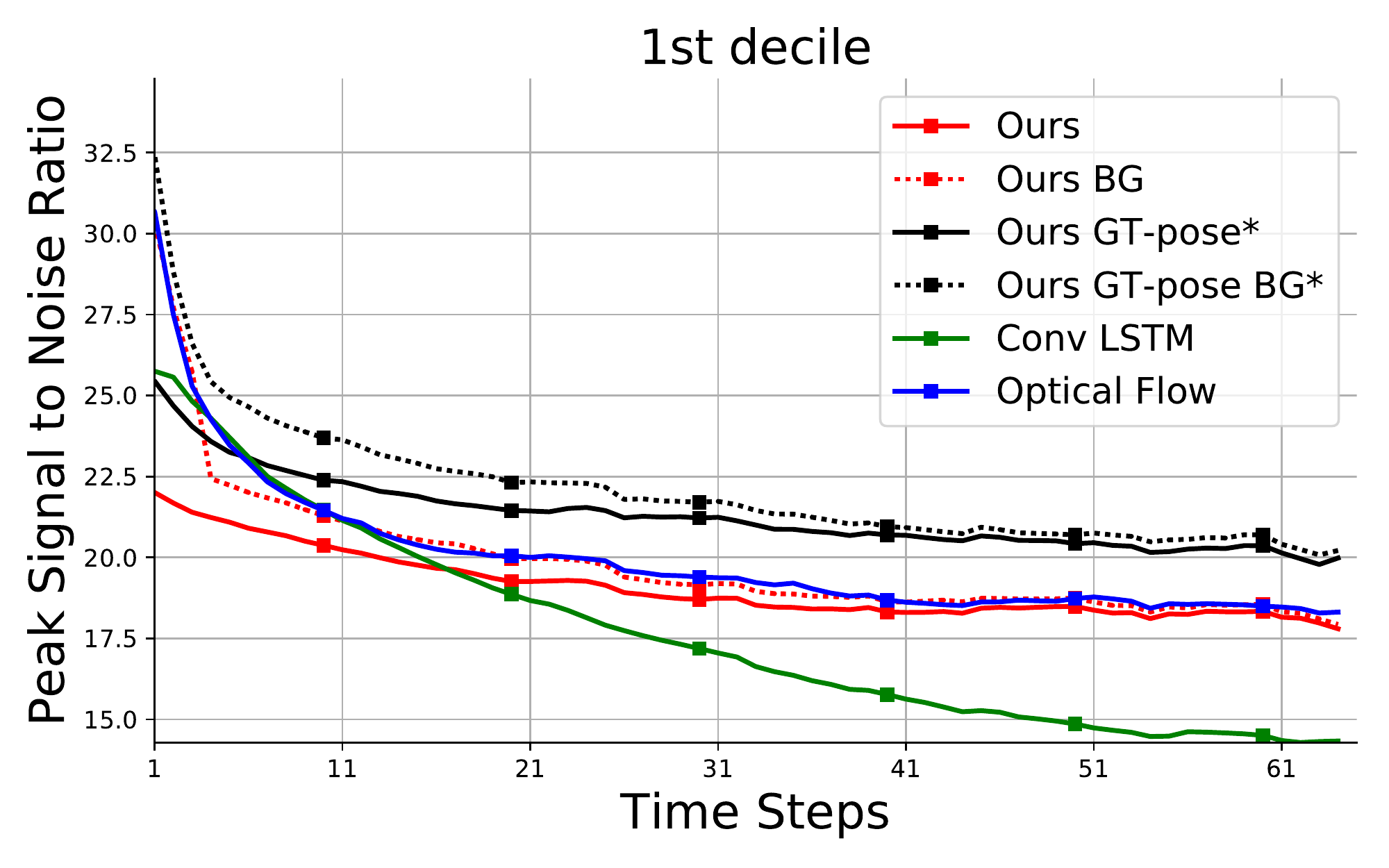}
\vspace{-.3cm}
\caption{(Continued from Figure~\ref{fig:penn_motion1}.) Quantitative comparison on Penn Action separated by motion decile.}
\label{fig:penn_motion2}
\end{figure}

\begin{figure}[htb!]
    \centering
    \vspace{-10pt}
    \begin{subfigure}{0.40\linewidth}
        \caption*{t=17}
        \vspace{-9pt}
	    \includegraphics[width=1\linewidth]{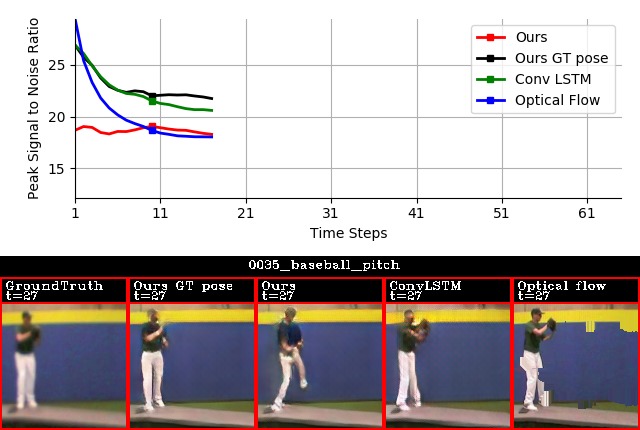}
  		\caption*{t=54}
        \vspace{-9pt}
  		\includegraphics[width=1\linewidth]{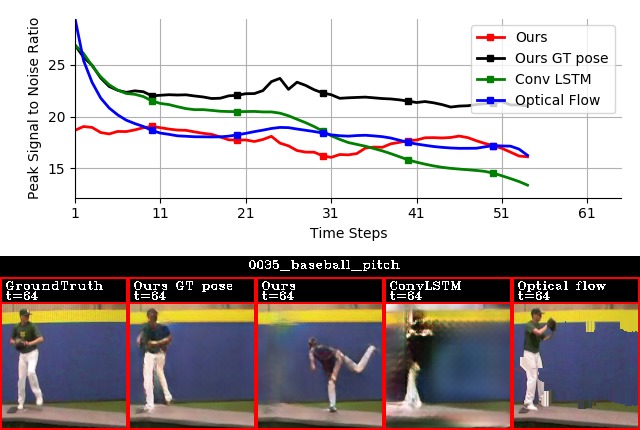}
  		\caption*{\textbf{Low PSNR}}
  		\caption*{-----------------------------------------------------------------}
        \caption*{t=12}
        \vspace{-9pt}
	    \includegraphics[width=1\linewidth]{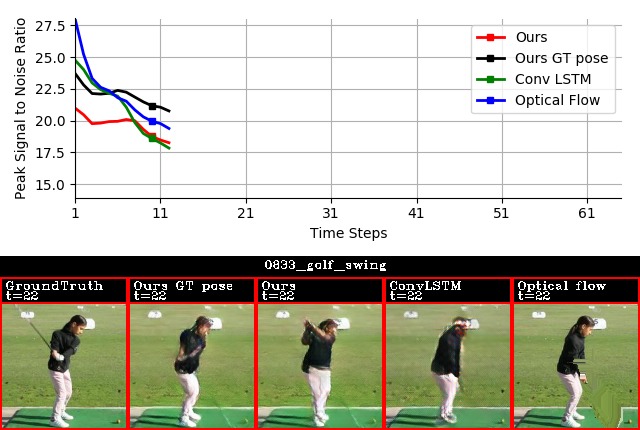}
  		\caption*{t=43}
        \vspace{-9pt}
  		\includegraphics[width=1\linewidth]{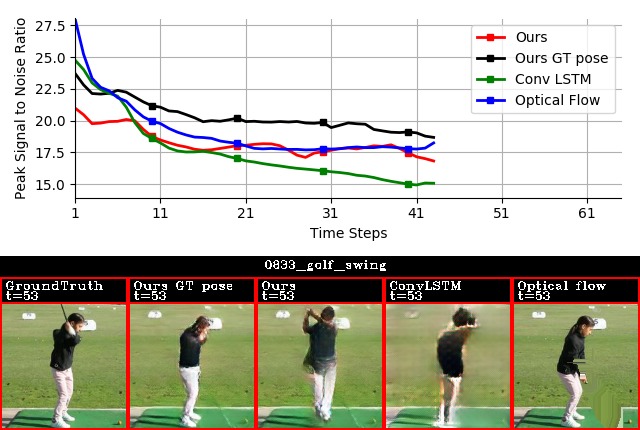}
  		\caption*{\textbf{Low PSNR}}
	\end{subfigure}
	\hspace{10pt}
    \begin{subfigure}{0.40\linewidth}
        \caption*{t=40}
        \vspace{-9pt}
	    \includegraphics[width=1\linewidth]{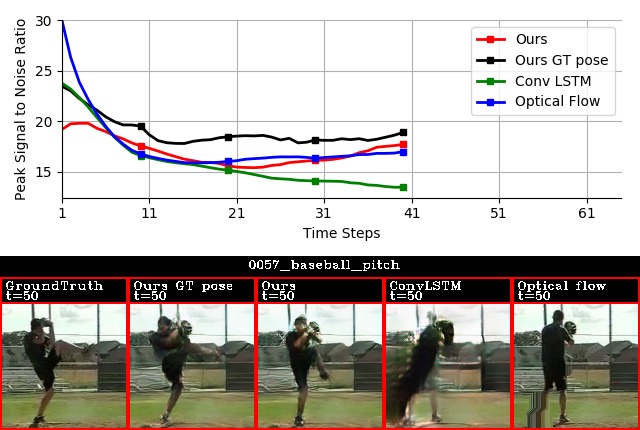}
  		\caption*{t=60}
        \vspace{-9pt}
  		\includegraphics[width=1\linewidth]{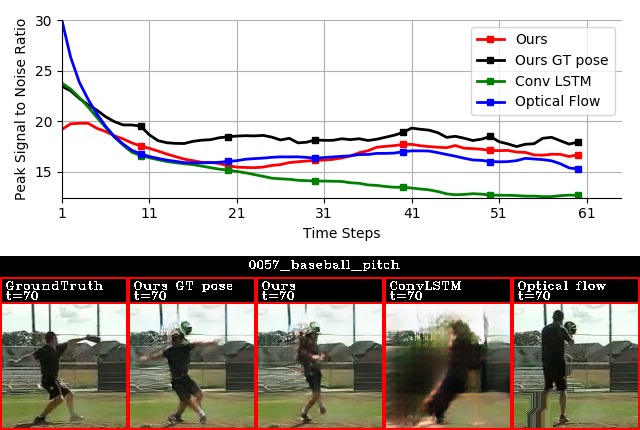}
  		\caption*{\textbf{High PSNR}}
  		\caption*{-----------------------------------------------------------------}
        \caption*{t=30}
        \vspace{-9pt}
	    \includegraphics[width=1\linewidth]{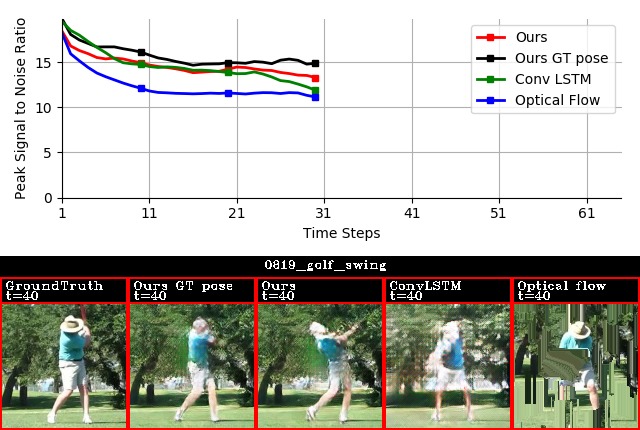}
  		\caption*{t=40}
        \vspace{-9pt}
  		\includegraphics[width=1\linewidth]{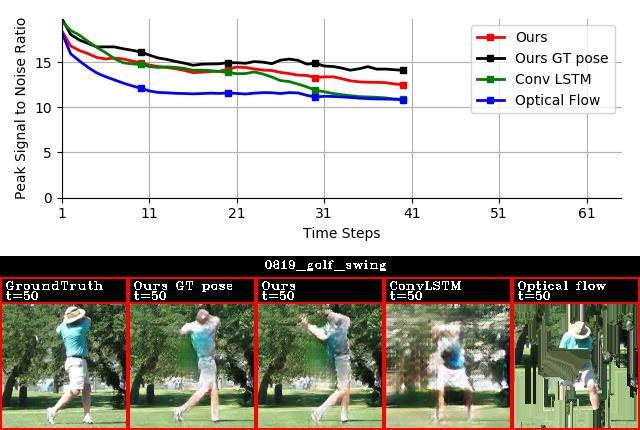}
  		\caption*{\textbf{High PSNR}}
	\end{subfigure}
	\vspace{-.3cm}
    \caption{Quantitative and visual comparison on Penn Action for selected time-steps for the action of \texttt{baseball pitch} (top) and \texttt{golf swing} (bottom). Side by side video comparison can be found in our \href{https://goo.gl/U7UOfy}{project website}}
\label{fig:pennbad1}
\vspace{-40pt}
\end{figure}

\begin{figure}[htb!]
    \centering
    \vspace{20pt}
    \begin{subfigure}{0.40\linewidth}
        \caption*{t=10}
        \vspace{-9pt}
	    \includegraphics[width=1\linewidth]{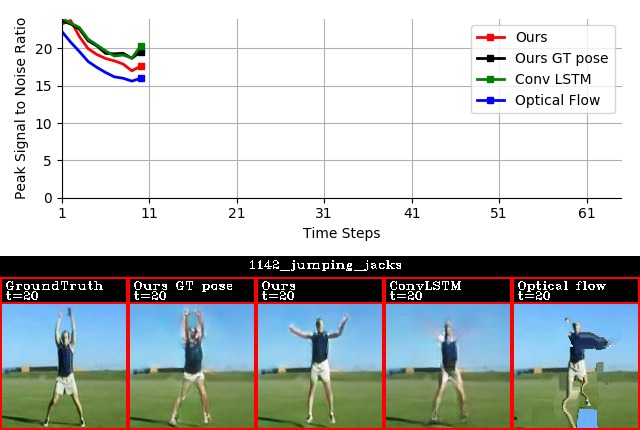}
  		\caption*{t=20}
        \vspace{-9pt}
  		\includegraphics[width=1\linewidth]{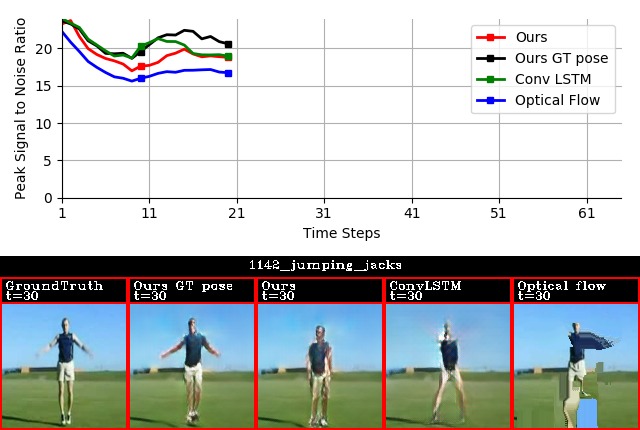}
  		\caption*{\textbf{Low PSNR}}
  		\caption*{-----------------------------------------------------------------}
  		\caption*{t=5}
        \vspace{-9pt}
	    \includegraphics[width=1\linewidth]{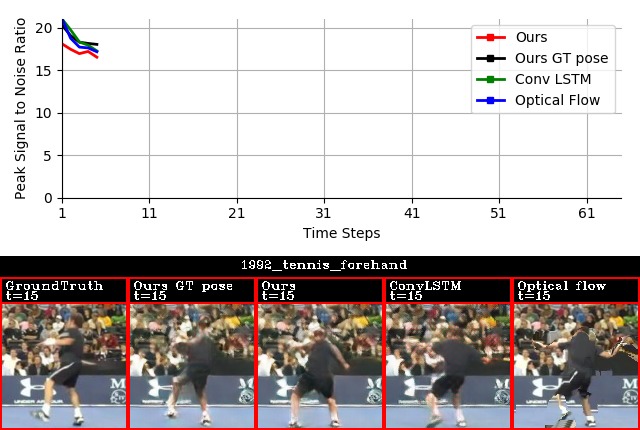}
  		\caption*{t=11}
        \vspace{-9pt}
  		\includegraphics[width=1\linewidth]{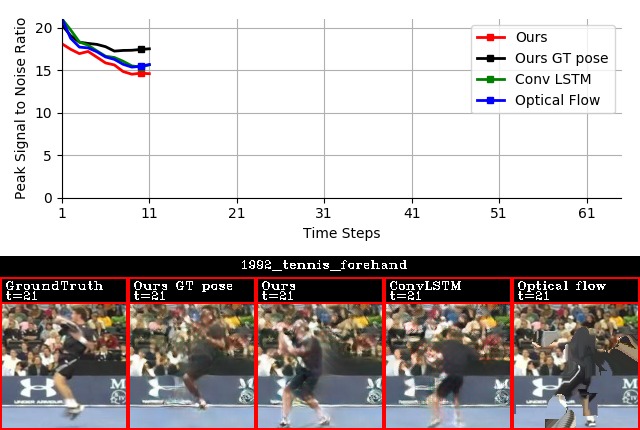}
  		\caption*{\textbf{Low PSNR}}
	\end{subfigure}
	\hspace{10pt}
    \begin{subfigure}{0.40\linewidth}
        \caption*{t=12}
        \vspace{-9pt}
	    \includegraphics[width=1\linewidth]{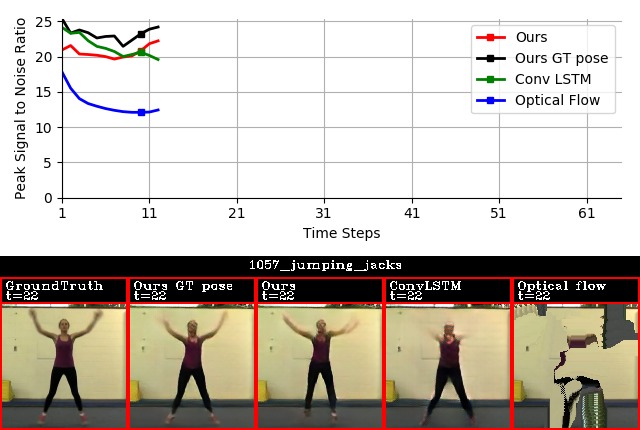}
  		\caption*{t=20}
        \vspace{-9pt}
  		\includegraphics[width=1\linewidth]{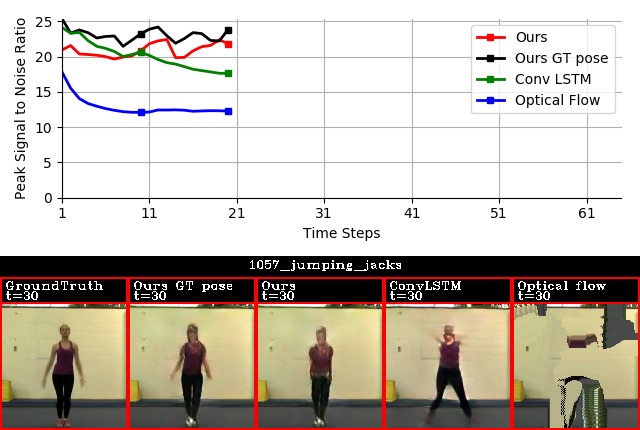}
  		\caption*{\textbf{High PSNR}}
  		\caption*{-----------------------------------------------------------------}
  		\caption*{t=25}
        \vspace{-9pt}
	    \includegraphics[width=1\linewidth]{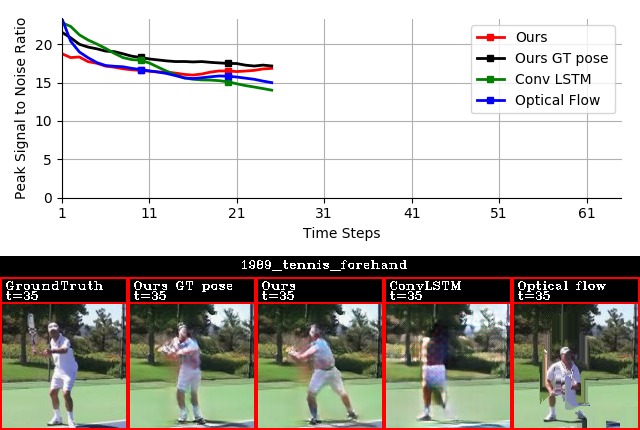}
  		\caption*{t=40}
        \vspace{-9pt}
  		\includegraphics[width=1\linewidth]{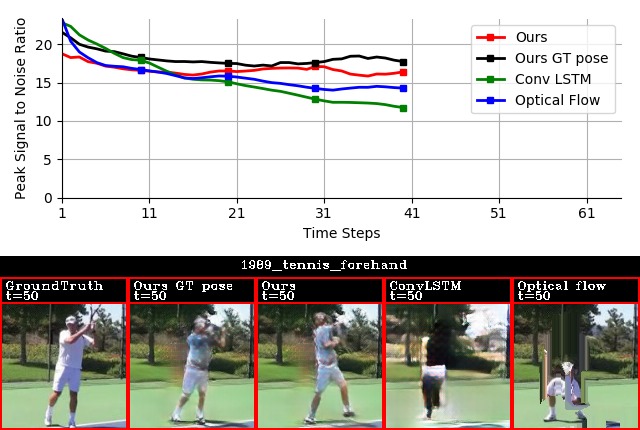}
  		\caption*{\textbf{High PSNR}}
	\end{subfigure}
	\vspace{-.1cm}
    \caption{Quantitative and visual comparison on Penn Action for selected time-steps for the actions of \texttt{jumping jacks} (top) and \texttt{tennis forehand} (bottom). Side by side video comparison can be found in our \href{https://goo.gl/U7UOfy}{project website}}
\label{fig:pennbad2}
\vspace{-40pt}
\end{figure}

The fact that PSNR can be low even if the predicted future is one of the many plausible futures suggest that PSNR may not be the best way to evaluate long-term video prediction when only a single future trajectory is predicted.
This issue might be alleviated when a model can predict multiple possible future trajectories, but this investigation using our hierarchical decomposition is left as future work. 
In Figures~\ref{fig:pennbad1} and \ref{fig:pennbad2}, we show videos where PSNR is low when a different future (from the ground-truth) is predicted (left), and video where PSNR is high because the predicted future is close to the ground-true future (right).

\clearpage

\begin{figure*}[!thbp]
    \centering
    \vspace{20pt}
	\begin{subfigure}{0.04\linewidth}
        \raggedleft
        \rotatebox{90}{
        \hspace{-.3cm}
        \parbox{2cm}{\centering Groundtruth} \parbox{2cm}{\centering Ours GT-pose} \parbox{2cm}{\centering Ours}
        \hspace{.1cm}
        \parbox{2cm}{\centering Groundtruth} \parbox{2cm}{\centering Ours GT-pose} \parbox{2cm}{\centering Ours}
        \hspace{.1cm}
        \parbox{2cm}{\centering Groundtruth} \parbox{2cm}{\centering Ours GT-pose} \parbox{2cm}{\centering Ours}
        }
    \end{subfigure}
    \begin{subfigure}{0.12\linewidth}
        \caption*{t=11}
        \vspace{-7pt}
	    \includegraphics[width=1\linewidth]{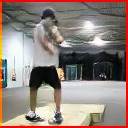}
	    \includegraphics[width=1\linewidth]{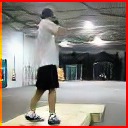}
	    \vspace{.2cm}
  		\includegraphics[width=1\linewidth]{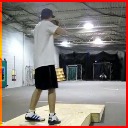}
  		\includegraphics[width=1\linewidth]{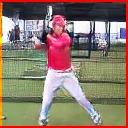}
  		\includegraphics[width=1\linewidth]{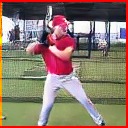}
  		\vspace{.2cm}
  		\includegraphics[width=1\linewidth]{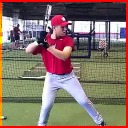}
  		\includegraphics[width=1\linewidth]{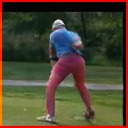}
  		\includegraphics[width=1\linewidth]{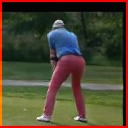}
  		\vspace{.2cm}
  		\includegraphics[width=1\linewidth]{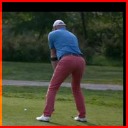}
	\end{subfigure} 
    \begin{subfigure}{0.12\linewidth}
        \caption*{t=20}
        \vspace{-7pt}
	    \includegraphics[width=1\linewidth]{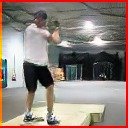}
	    \includegraphics[width=1\linewidth]{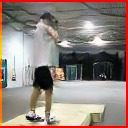}
	    \vspace{.2cm}
  		\includegraphics[width=1\linewidth]{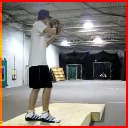}
  		\includegraphics[width=1\linewidth]{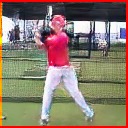}
  		\includegraphics[width=1\linewidth]{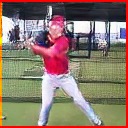}
  		\vspace{.2cm}
  		\includegraphics[width=1\linewidth]{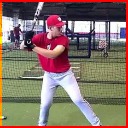}
  		\includegraphics[width=1\linewidth]{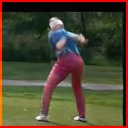}
  		\includegraphics[width=1\linewidth]{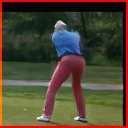}
  		\vspace{.2cm}
  		\includegraphics[width=1\linewidth]{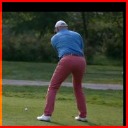}
	\end{subfigure} 
    \begin{subfigure}{0.12\linewidth}
        \caption*{t=29}
        \vspace{-7pt}
	    \includegraphics[width=1\linewidth]{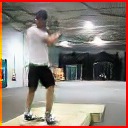}
	    \includegraphics[width=1\linewidth]{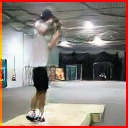}
	    \vspace{.2cm}
  		\includegraphics[width=1\linewidth]{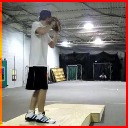}
  		\includegraphics[width=1\linewidth]{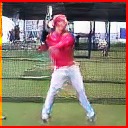}
  		\includegraphics[width=1\linewidth]{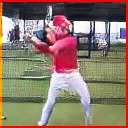}
  		\vspace{.2cm}
  		\includegraphics[width=1\linewidth]{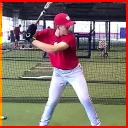}
  		\includegraphics[width=1\linewidth]{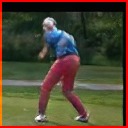}
  		\includegraphics[width=1\linewidth]{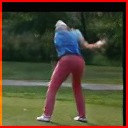}
  		\vspace{.2cm}
  		\includegraphics[width=1\linewidth]{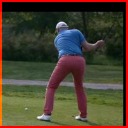}
	\end{subfigure} 
    \begin{subfigure}{0.12\linewidth}
        \caption*{t=38}
        \vspace{-7pt}
	    \includegraphics[width=1\linewidth]{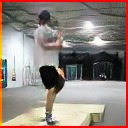}
	    \includegraphics[width=1\linewidth]{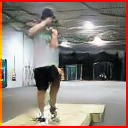}
	    \vspace{.2cm}
  		\includegraphics[width=1\linewidth]{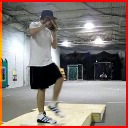}
  		\includegraphics[width=1\linewidth]{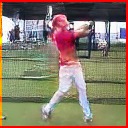}
  		\includegraphics[width=1\linewidth]{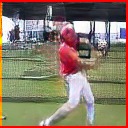}
  		\vspace{.2cm}
  		\includegraphics[width=1\linewidth]{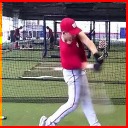}
  		\includegraphics[width=1\linewidth]{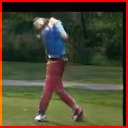}
  		\includegraphics[width=1\linewidth]{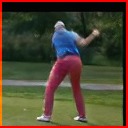}
  		\vspace{.2cm}
  		\includegraphics[width=1\linewidth]{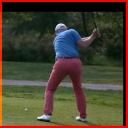}
	\end{subfigure}
	\begin{subfigure}{0.12\linewidth}
        \caption*{t=47}
        \vspace{-7pt}
	    \includegraphics[width=1\linewidth]{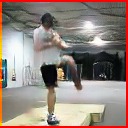}
	    \includegraphics[width=1\linewidth]{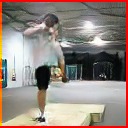}
	    \vspace{.2cm}
  		\includegraphics[width=1\linewidth]{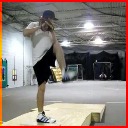}
  		\includegraphics[width=1\linewidth]{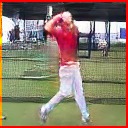}
  		\includegraphics[width=1\linewidth]{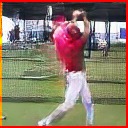}
  		\vspace{.2cm}
  		\includegraphics[width=1\linewidth]{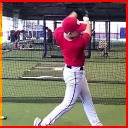}
  		\includegraphics[width=1\linewidth]{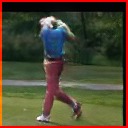}
  		\includegraphics[width=1\linewidth]{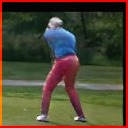}
  		\vspace{.2cm}
  		\includegraphics[width=1\linewidth]{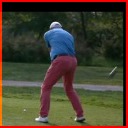}
	\end{subfigure}
	\begin{subfigure}{0.12\linewidth}
        \caption*{t=56}
        \vspace{-7pt}
	    \includegraphics[width=1\linewidth]{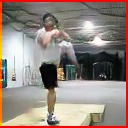}
	    \includegraphics[width=1\linewidth]{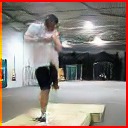}
	    \vspace{.2cm}
  		\includegraphics[width=1\linewidth]{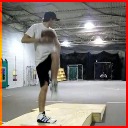}
  		\includegraphics[width=1\linewidth]{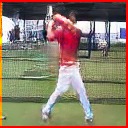}
  		\includegraphics[width=1\linewidth]{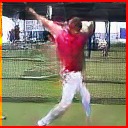}
  		\vspace{.2cm}
  		\includegraphics[width=1\linewidth]{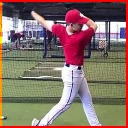}
  		\includegraphics[width=1\linewidth]{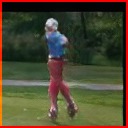}
  		\includegraphics[width=1\linewidth]{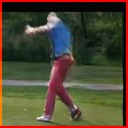}
  		\vspace{.2cm}
  		\includegraphics[width=1\linewidth]{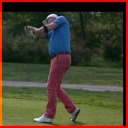}
	\end{subfigure}
	\begin{subfigure}{0.12\linewidth}
        \caption*{t=65}
        \vspace{-7pt}
	    \includegraphics[width=1\linewidth]{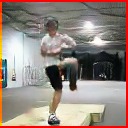}
	    \includegraphics[width=1\linewidth]{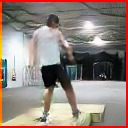}
	    \vspace{.2cm}
  		\includegraphics[width=1\linewidth]{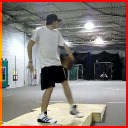}
  		\includegraphics[width=1\linewidth]{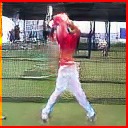}
  		\includegraphics[width=1\linewidth]{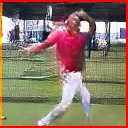}
  		\vspace{.2cm}
  		\includegraphics[width=1\linewidth]{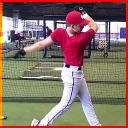}
  		\includegraphics[width=1\linewidth]{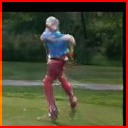}
  		\includegraphics[width=1\linewidth]{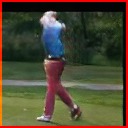}
  		\vspace{.2cm}
  		\includegraphics[width=1\linewidth]{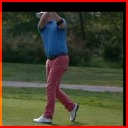}
	\end{subfigure}
    \caption{Qualitative evaluation of our network for long-term pixel-level generation. We show the actions of \texttt{baseball pitch} (top row), \texttt{baseball swing} (middle row), and \texttt{gold swing} (bottom row). Side by side video comparison can be found in our \href{https://goo.gl/U7UOfy}{project website}.}
\label{fig:penngtft}
\vspace{-40pt}
\end{figure*}

\begin{figure*}[!thbp]
    \centering
    \vspace{60pt}
	\begin{subfigure}{0.04\linewidth}
        \raggedleft
        \rotatebox{90}{
        \hspace{-.2cm}
        \parbox{2cm}{\centering Groundtruth} \parbox{2cm}{\centering Ours GT-pose} \parbox{2cm}{\centering Ours}
        \hspace{.6cm}
        \parbox{2cm}{\centering Groundtruth} \parbox{2cm}{\centering Ours GT-pose} \parbox{2cm}{\centering Ours}
        \hspace{.1cm}
        \parbox{2cm}{\centering Groundtruth} \parbox{2cm}{\centering Ours GT-pose} \parbox{2cm}{\centering Ours}
        }
    \end{subfigure}
    \begin{subfigure}{0.12\linewidth}
        \caption*{t=11}
        \vspace{-7pt}
	    \includegraphics[width=1\linewidth]{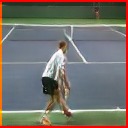}
	    \includegraphics[width=1\linewidth]{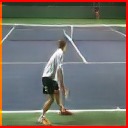}
	    \vspace{.2cm}
  		\includegraphics[width=1\linewidth]{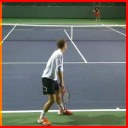}
  		\includegraphics[width=1\linewidth]{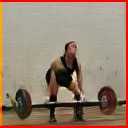}
  		\includegraphics[width=1\linewidth]{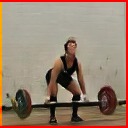}
  		\vspace{.2cm}
  		\includegraphics[width=1\linewidth]{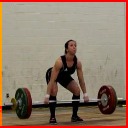}
  		\caption*{t=11}
  		\vspace{-7pt}
  		\includegraphics[width=1\linewidth]{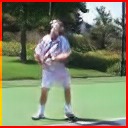}
  		\includegraphics[width=1\linewidth]{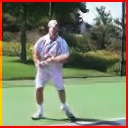}
  		\vspace{.2cm}
  		\includegraphics[width=1\linewidth]{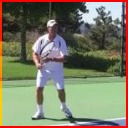}
	\end{subfigure} 
    \begin{subfigure}{0.12\linewidth}
        \caption*{t=20}
        \vspace{-7pt}
	    \includegraphics[width=1\linewidth]{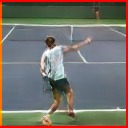}
	    \includegraphics[width=1\linewidth]{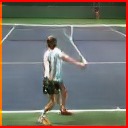}
	    \vspace{.2cm}
  		\includegraphics[width=1\linewidth]{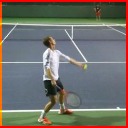}
  		\includegraphics[width=1\linewidth]{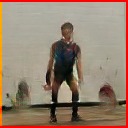}
  		\includegraphics[width=1\linewidth]{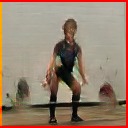}
  		\vspace{.2cm}
  		\includegraphics[width=1\linewidth]{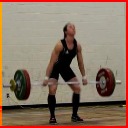}
  		\caption*{t=17}
  		\vspace{-7pt}
  		\includegraphics[width=1\linewidth]{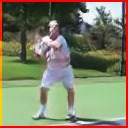}
  		\includegraphics[width=1\linewidth]{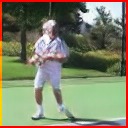}
  		\vspace{.2cm}
  		\includegraphics[width=1\linewidth]{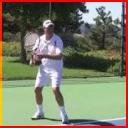}
	\end{subfigure} 
    \begin{subfigure}{0.12\linewidth}
        \caption*{t=29}
        \vspace{-7pt}
	    \includegraphics[width=1\linewidth]{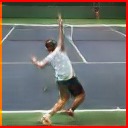}
	    \includegraphics[width=1\linewidth]{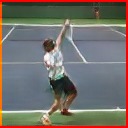}
	    \vspace{.2cm}
  		\includegraphics[width=1\linewidth]{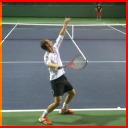}
  		\includegraphics[width=1\linewidth]{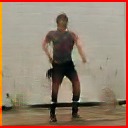}
  		\includegraphics[width=1\linewidth]{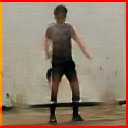}
  		\vspace{.2cm}
  		\includegraphics[width=1\linewidth]{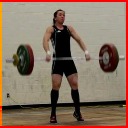}
  		\caption*{t=23}
  		\vspace{-7pt}
  		\includegraphics[width=1\linewidth]{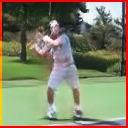}
  		\includegraphics[width=1\linewidth]{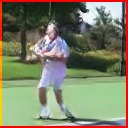}
  		\vspace{.2cm}
  		\includegraphics[width=1\linewidth]{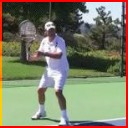}
	\end{subfigure} 
    \begin{subfigure}{0.12\linewidth}
        \caption*{t=38}
        \vspace{-7pt}
	    \includegraphics[width=1\linewidth]{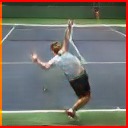}
	    \includegraphics[width=1\linewidth]{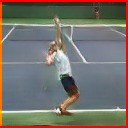}
	    \vspace{.2cm}
  		\includegraphics[width=1\linewidth]{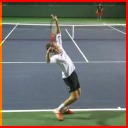}
  		\includegraphics[width=1\linewidth]{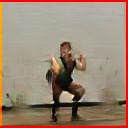}
  		\includegraphics[width=1\linewidth]{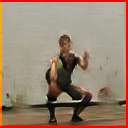}
  		\vspace{.2cm}
  		\includegraphics[width=1\linewidth]{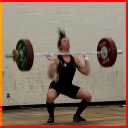}
  		\caption*{t=29}
  		\vspace{-7pt}
  		\includegraphics[width=1\linewidth]{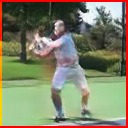}
  		\includegraphics[width=1\linewidth]{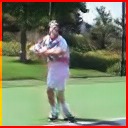}
  		\vspace{.2cm}
  		\includegraphics[width=1\linewidth]{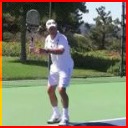}
	\end{subfigure}
	\begin{subfigure}{0.12\linewidth}
        \caption*{t=47}
        \vspace{-7pt}
	    \includegraphics[width=1\linewidth]{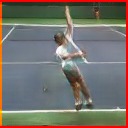}
	    \includegraphics[width=1\linewidth]{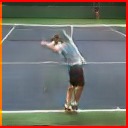}
	    \vspace{.2cm}
  		\includegraphics[width=1\linewidth]{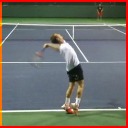}
  		\includegraphics[width=1\linewidth]{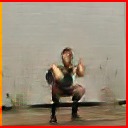}
  		\includegraphics[width=1\linewidth]{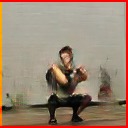}
  		\vspace{.2cm}
  		\includegraphics[width=1\linewidth]{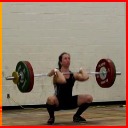}
  		\caption*{t=35}
  		\vspace{-7pt}
  		\includegraphics[width=1\linewidth]{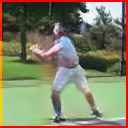}
  		\includegraphics[width=1\linewidth]{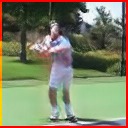}
  		\vspace{.2cm}
  		\includegraphics[width=1\linewidth]{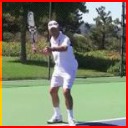}
	\end{subfigure}
	\begin{subfigure}{0.12\linewidth}
        \caption*{t=56}
        \vspace{-7pt}
	    \includegraphics[width=1\linewidth]{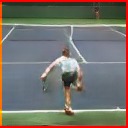}
	    \includegraphics[width=1\linewidth]{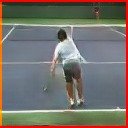}
	    \vspace{.2cm}
  		\includegraphics[width=1\linewidth]{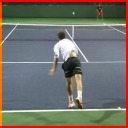}
  		\includegraphics[width=1\linewidth]{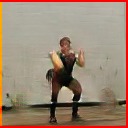}
  		\includegraphics[width=1\linewidth]{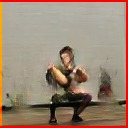}
  		\vspace{.2cm}
  		\includegraphics[width=1\linewidth]{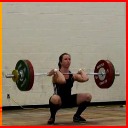}
  		\caption*{t=41}
  		\vspace{-7pt}
  		\includegraphics[width=1\linewidth]{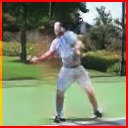}
  		\includegraphics[width=1\linewidth]{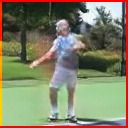}
  		\vspace{.2cm}
  		\includegraphics[width=1\linewidth]{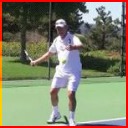}
	\end{subfigure}
	\begin{subfigure}{0.12\linewidth}
        \caption*{t=65}
        \vspace{-7pt}
	    \includegraphics[width=1\linewidth]{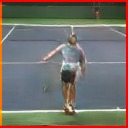}
	    \includegraphics[width=1\linewidth]{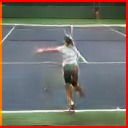}
	    \vspace{.2cm}
  		\includegraphics[width=1\linewidth]{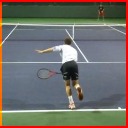}
  		\includegraphics[width=1\linewidth]{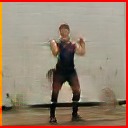}
  		\includegraphics[width=1\linewidth]{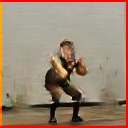}
  		\vspace{.2cm}
  		\includegraphics[width=1\linewidth]{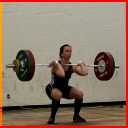}
  		\caption*{t=47}
  		\vspace{-7pt}
  		\includegraphics[width=1\linewidth]{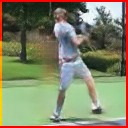}
  		\includegraphics[width=1\linewidth]{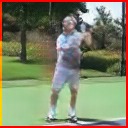}
  		\vspace{.2cm}
  		\includegraphics[width=1\linewidth]{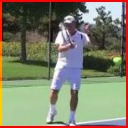}
	\end{subfigure}
    \caption{Qualitative evaluation of our network for long-term pixel-level generation. We show the actions of \texttt{tennis serve} (top row), \texttt{clean and jerk} (middle row), and \texttt{tennis forehand} (bottom row). We show a different timescale for \texttt{tennis forehand} because the ground-truth action sequence does not reach time step 65. Side by side video comparison can be found in our \href{https://goo.gl/U7UOfy}{project website}.}
\label{fig:penngtft2}
\vspace{-40pt}
\end{figure*}

\clearpage
To directly compare our image generator using the predicted future pose (\texttt{Ours}) and the ground-truth future pose given by the oracle (\texttt{Ours GT-pose$^*$}), we present qualitative experiments in Figure~\ref{fig:penngtft} and Figure~\ref{fig:penngtft2}.
We can see that the both predicted videos contain the action in the video. The oracle based video prediction reflects the exact future very well.

\clearpage

\clearpage

\subsection{Human3.6M}\label{supp:h36m}
\begin{figure}[htb!]
\centering
\includegraphics[width=0.40\linewidth] {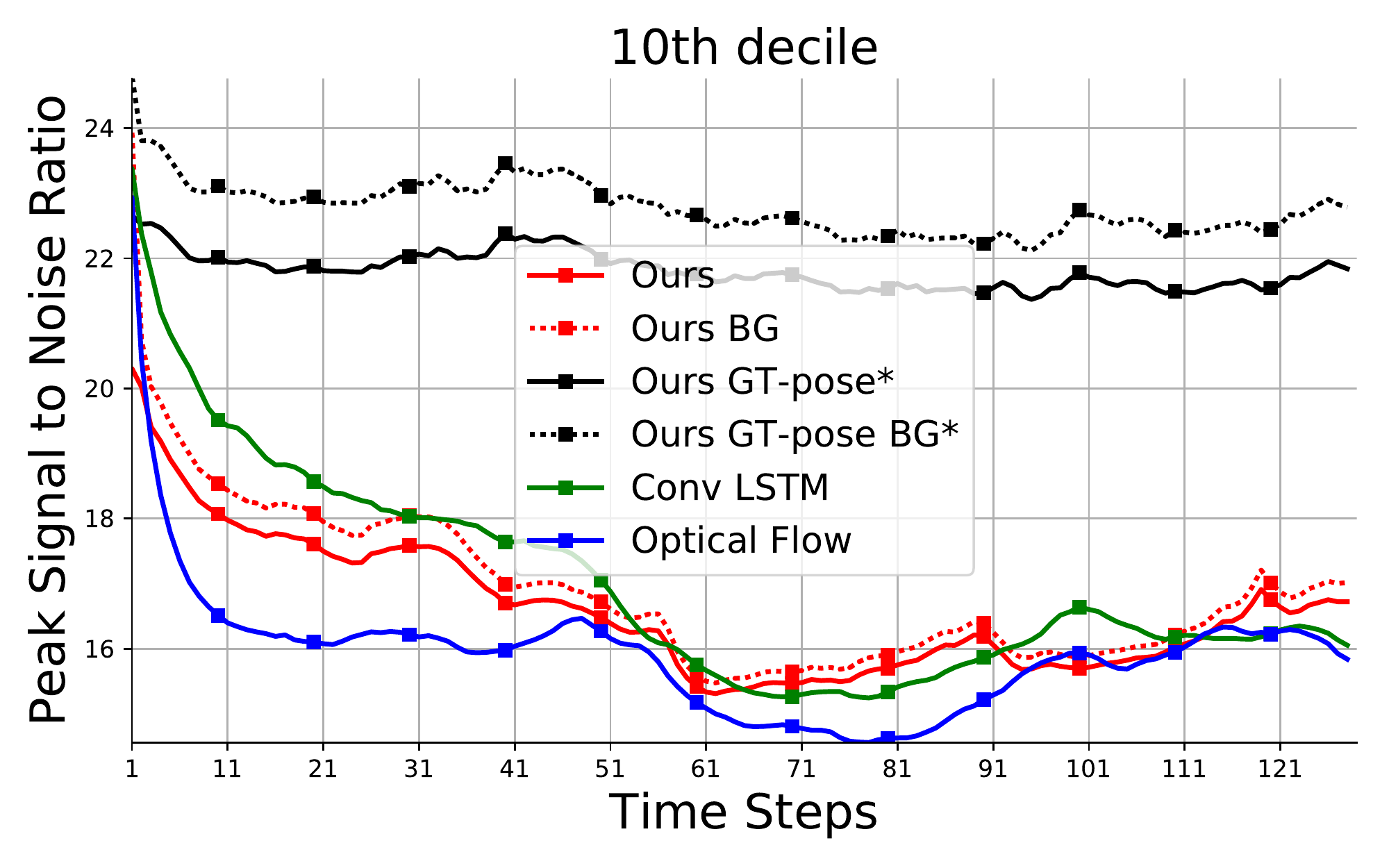}
\hspace{.8cm}
\includegraphics[width=0.40\linewidth] {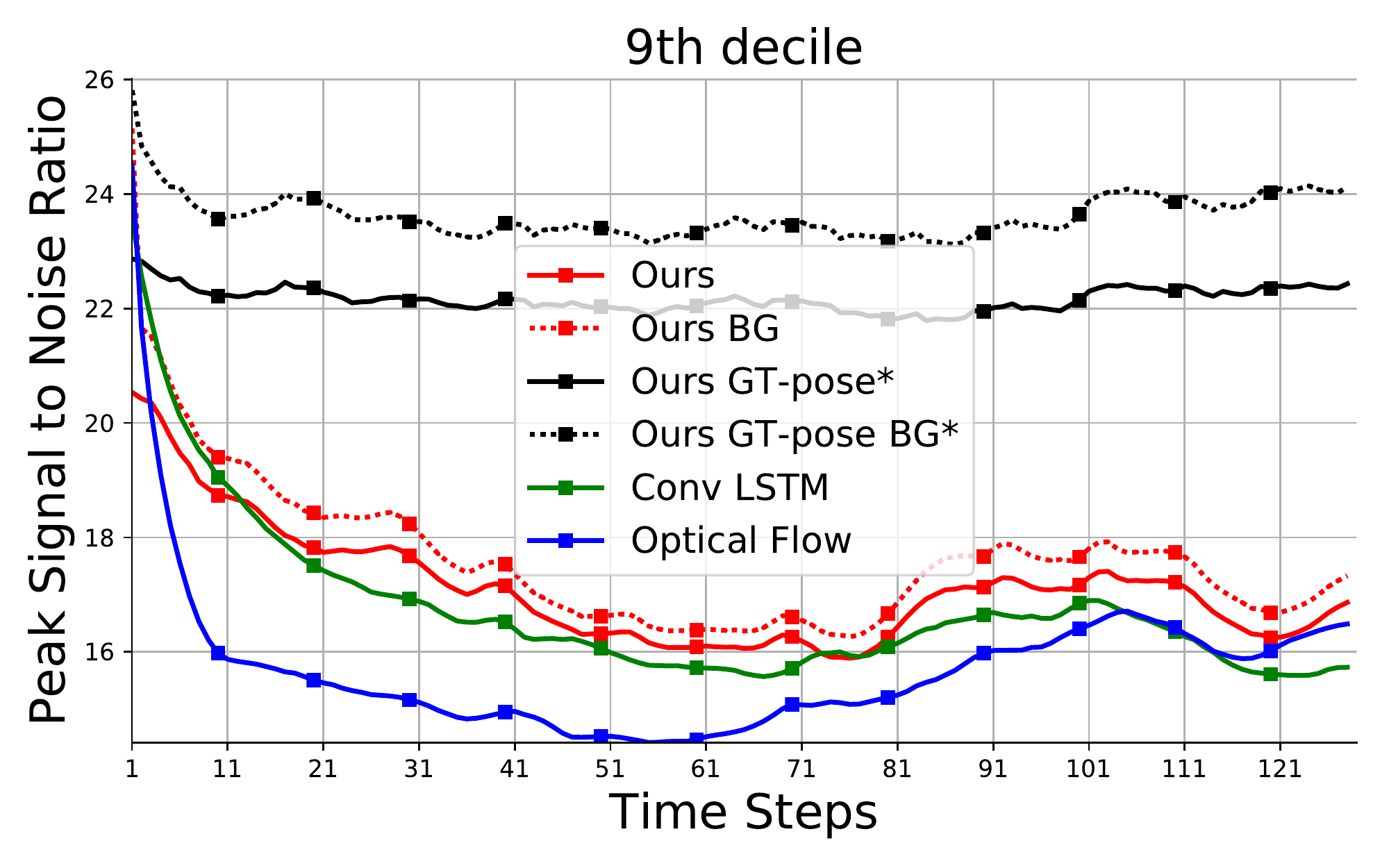} \\
\includegraphics[width=0.40\linewidth] {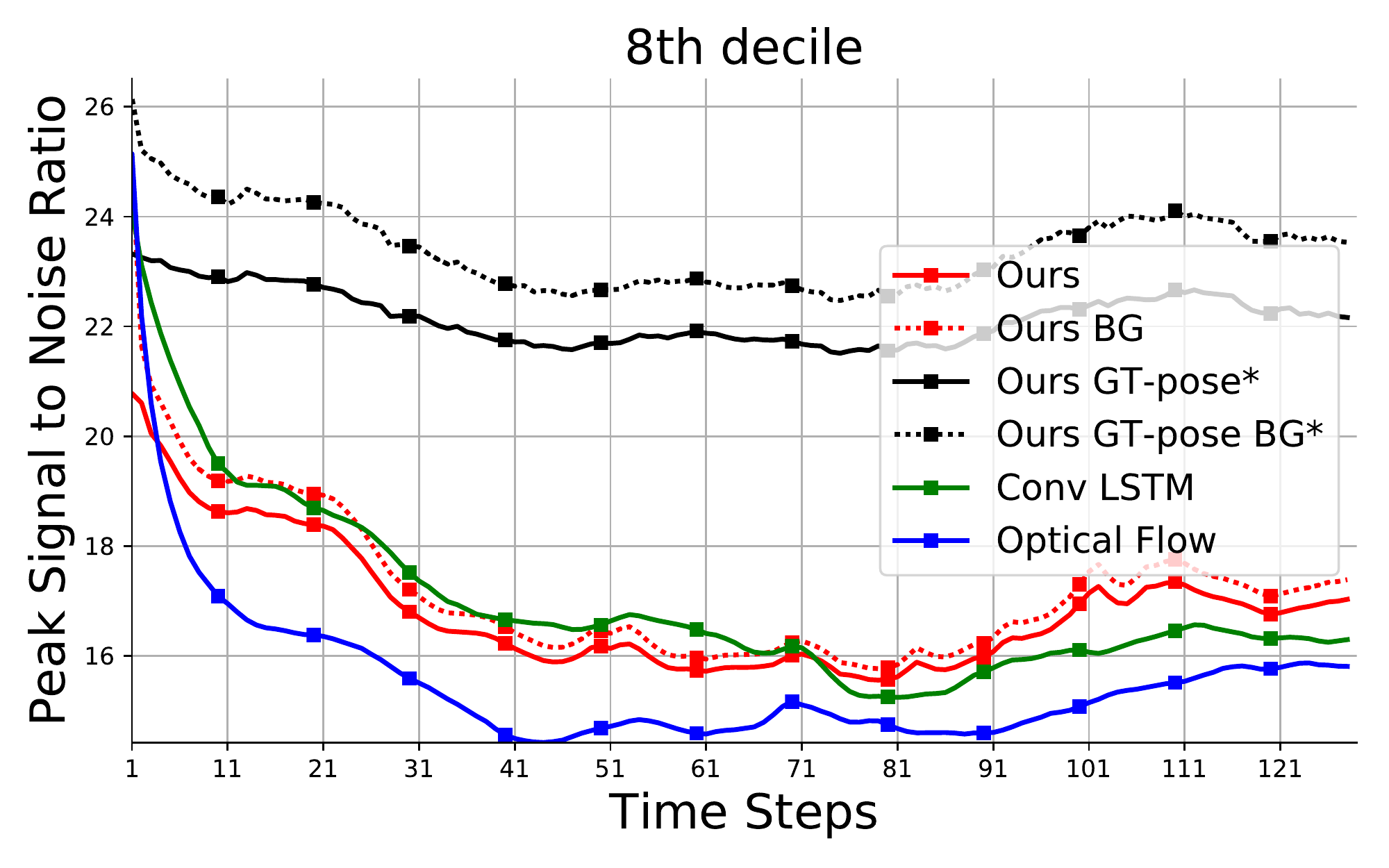}
\hspace{.8cm}
\includegraphics[width=0.40\linewidth] {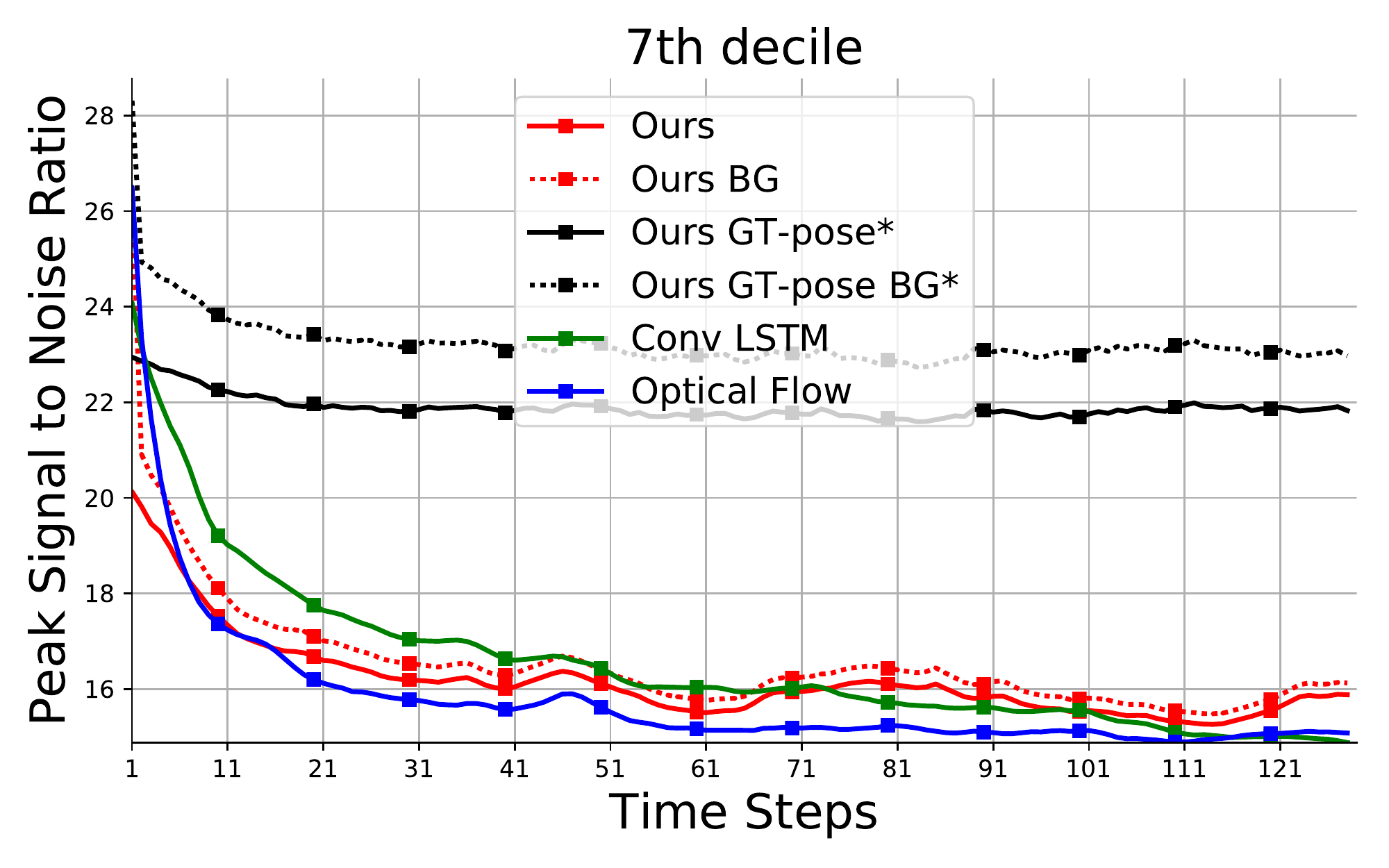} \\
\includegraphics[width=0.40\linewidth] {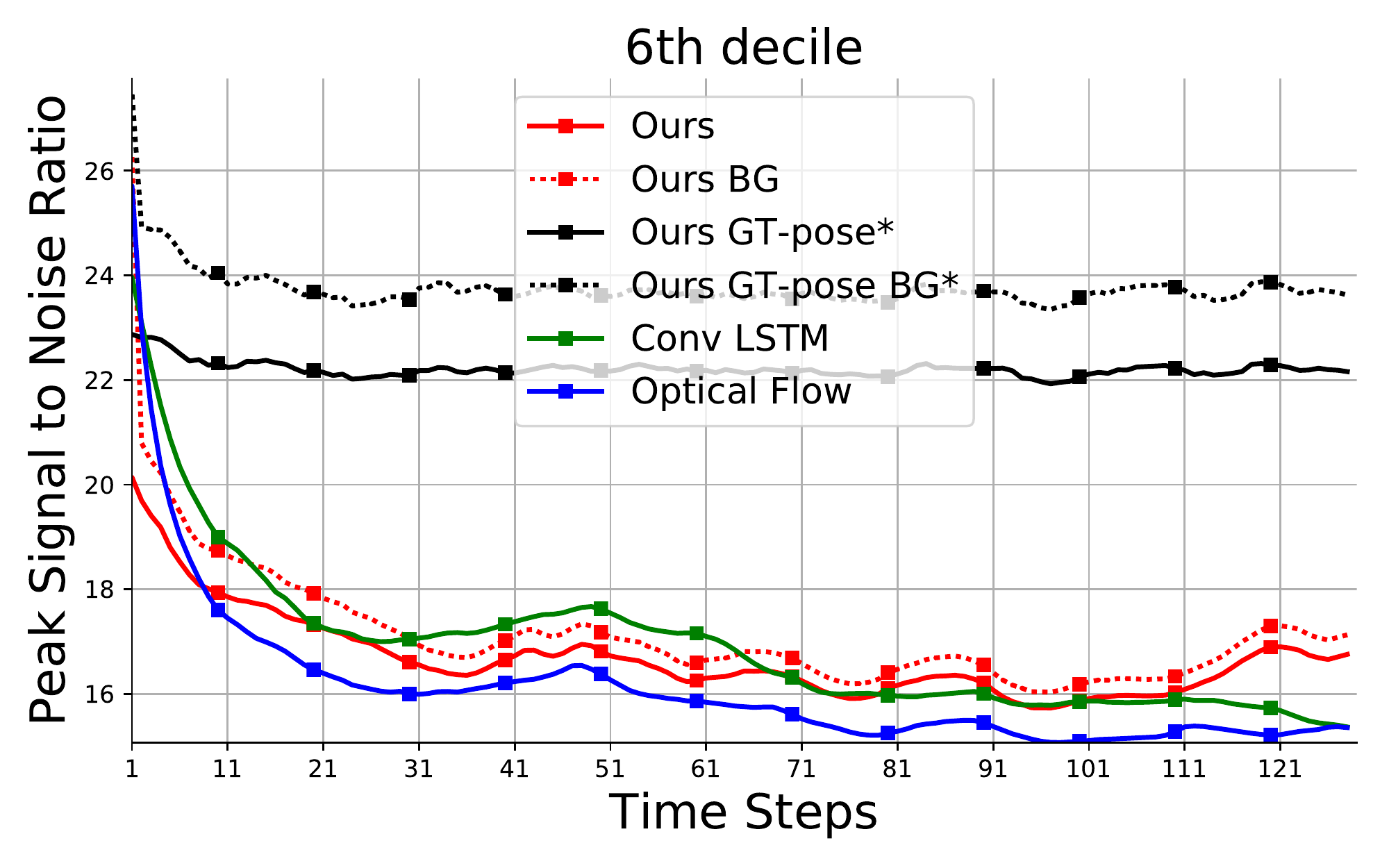}
\hspace{.8cm}
\includegraphics[width=0.40\linewidth] {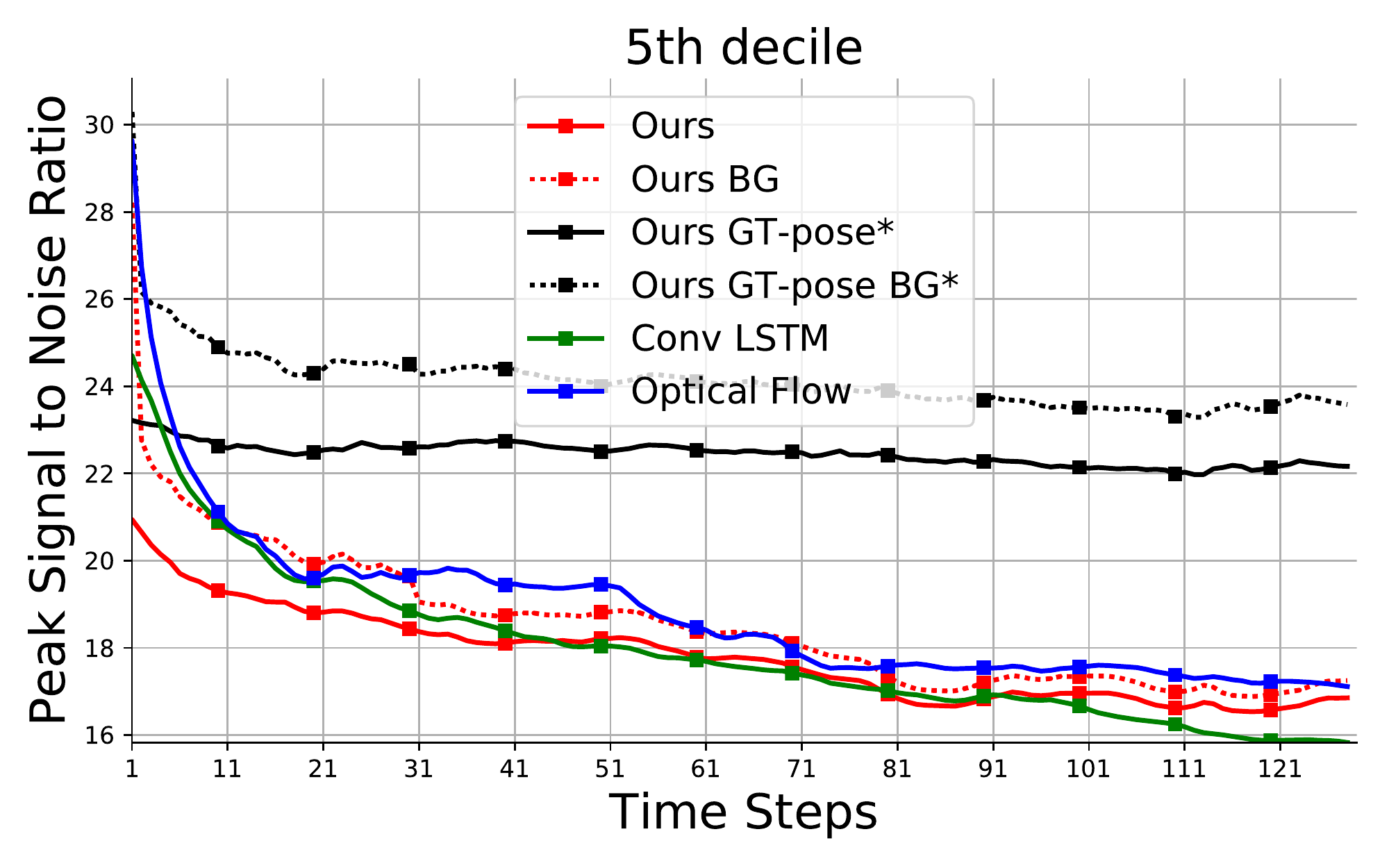} \\
\includegraphics[width=0.40\linewidth] {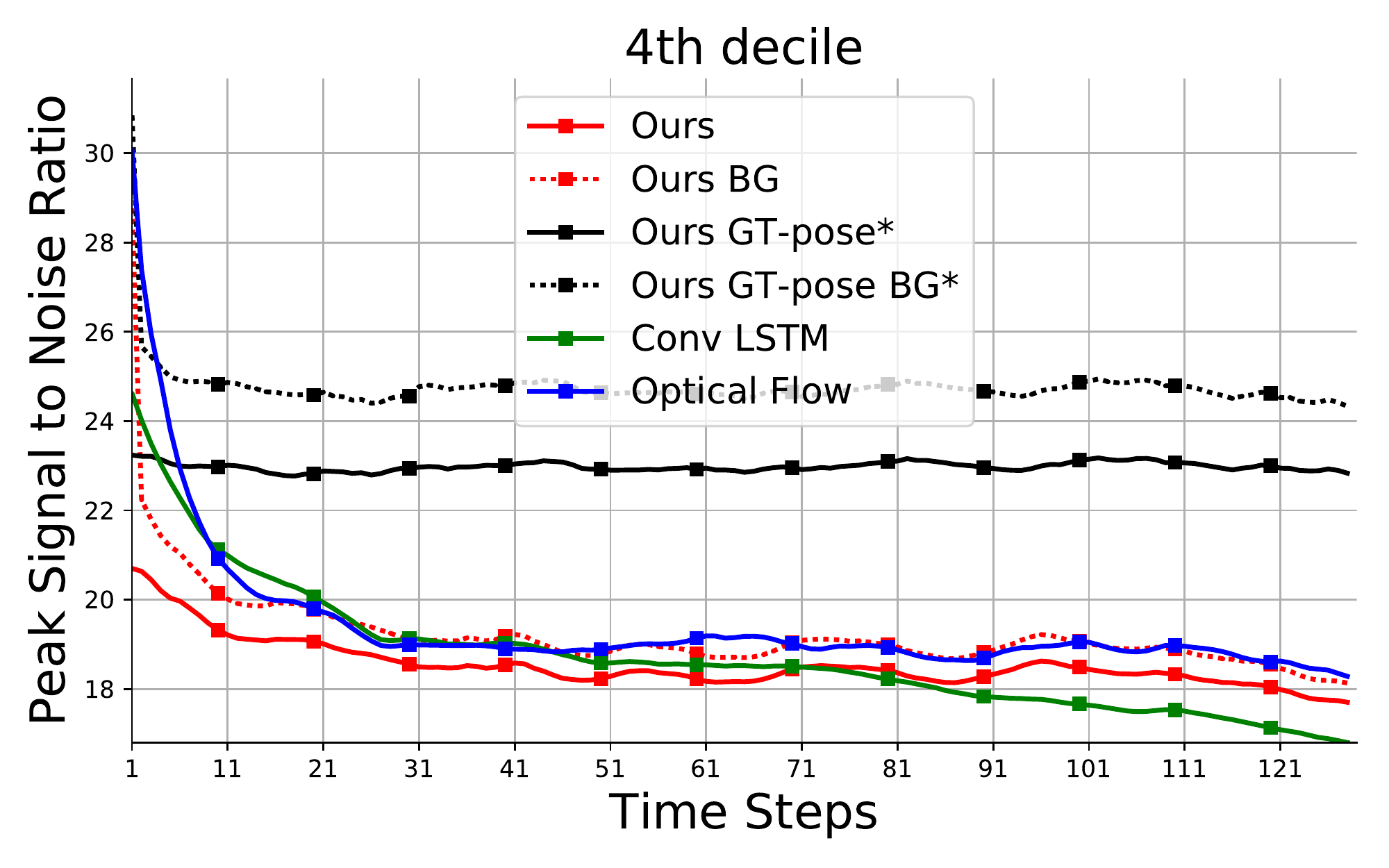}
\hspace{.8cm}
\includegraphics[width=0.40\linewidth] {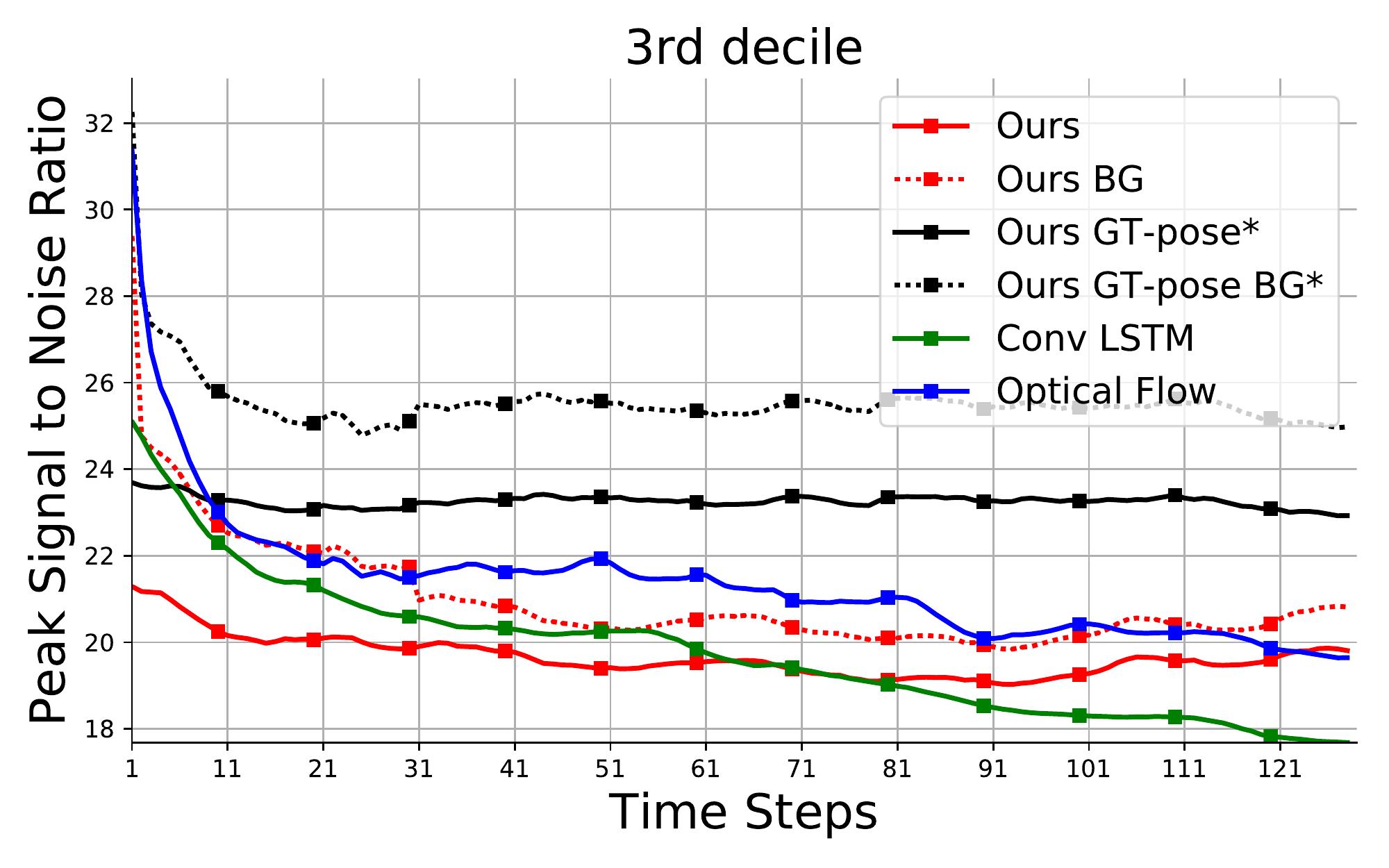} \\
\includegraphics[width=0.40\linewidth] {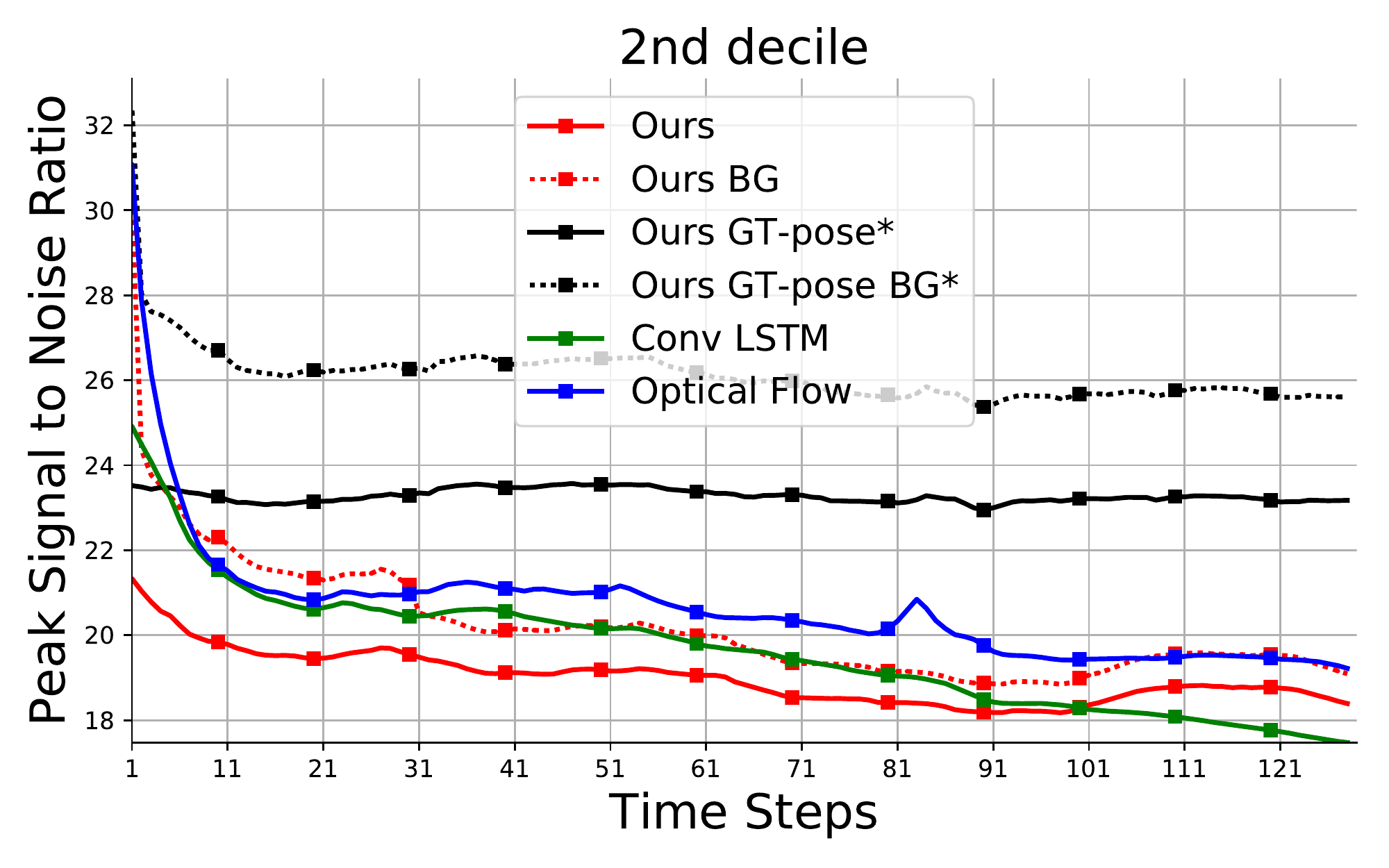}
\hspace{.8cm}
\includegraphics[width=0.40\linewidth] {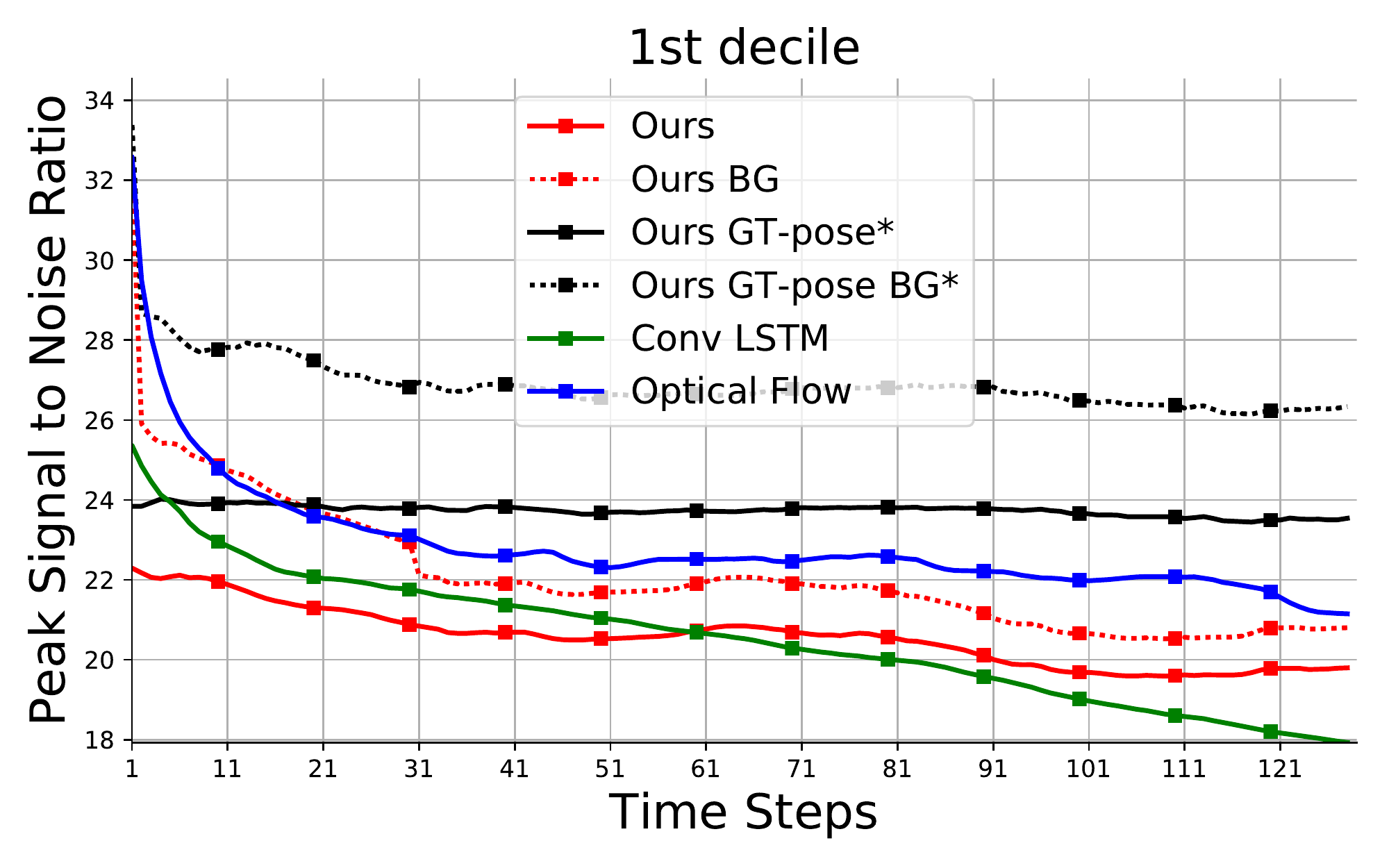}
\vspace{-.3cm}
\caption{Quantitative comparison on Human3.6M separated by motion decile.}
\label{fig:h36m_motion1}
\vspace{-2cm}
\end{figure}

In Figure~\ref{fig:h36m_motion1}, we show evaluation (PSNRs over time) of different methods on each decile of motion.

\clearpage

As shown in Figure~\ref{fig:h36m_motion1}, our hierarchical approach (e.g., \texttt{Ours BG}) tends to achieve PSNR performance that is better than optical flow based method and comparable to convolutional LSTM. In addition, when using the oracle future pose predictor as input to our image generator, the PSNR scores get a larger boost compared to Section~\ref{supp:penn}. 
This is because there is higher uncertainty of the actions being performed in the Human 3.6M dataset compared to Penn Action dataset. 
Therefore, even plausible future predictions can still deviate significantly from the ground-truth future trajectory, which can penalize PSNRs.

\begin{figure}[htb!]
\centering
\vspace{-10pt}
\includegraphics[width=0.40\linewidth] {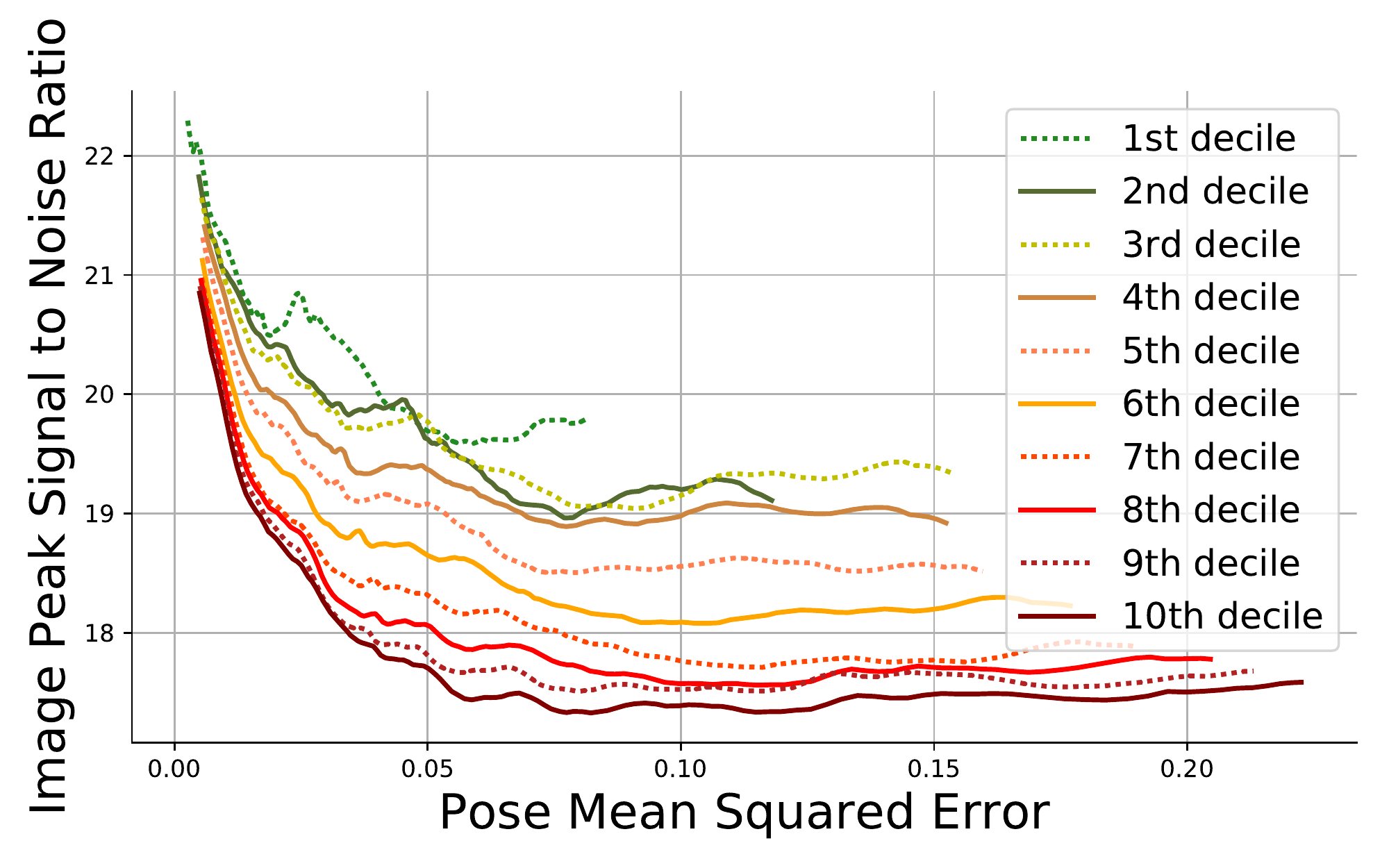}
\vspace{-.3cm}
\caption{Predicted frames PSNR vs. Mean Squared Error on the predicted pose for each motion decile in Human3.6M.}
\label{fig:h36m_corr}
\end{figure}
\vspace{-.3cm}

To gain further insight on this problem,  we provide two additional analysis. 
First, we compute how the average PSNR changes as the future pose MSE increases in Figure~\ref{fig:h36m_corr}.
The figure clearly shows the negative correlation between the predicted pose MSE and frame PSNR, meaning that larger deviation of the predicted future pose from the ground future pose tend to cause lower PSNRs.

\begin{figure}[htb!]
    \centering
    \begin{subfigure}{0.40\linewidth}
        \caption*{t=31}
        \vspace{-9pt}
	    \includegraphics[width=1\linewidth]{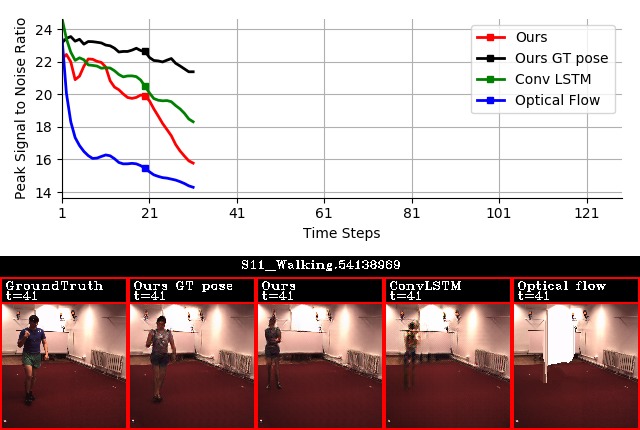}
  		\caption*{t=80}
        \vspace{-9pt}
  		\includegraphics[width=1\linewidth]{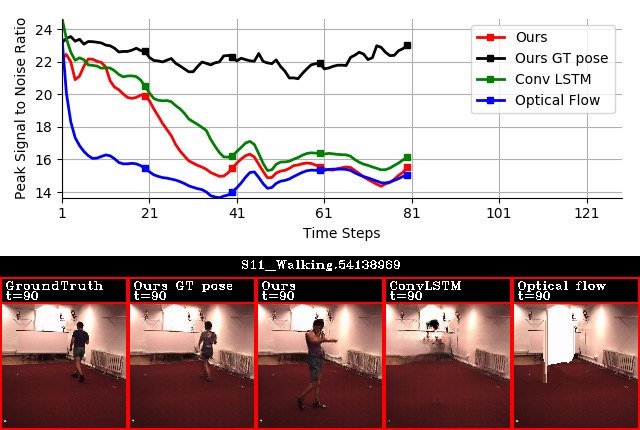}
  		\caption*{\textbf{Low PSNR}}
	\end{subfigure}
	\hspace{10pt}
    \begin{subfigure}{0.40\linewidth}
        \caption*{t=61}
        \vspace{-9pt}
	    \includegraphics[width=1\linewidth]{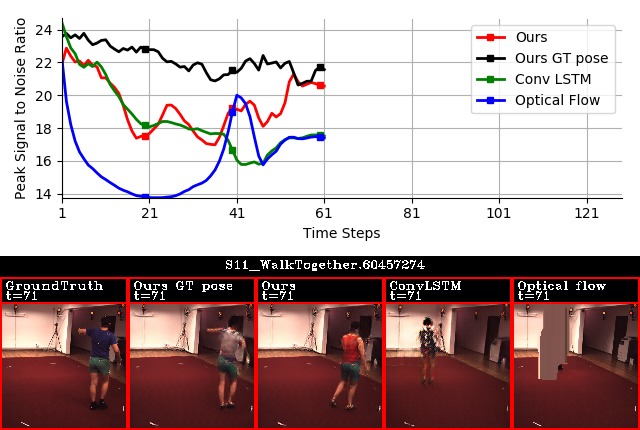}
  		\caption*{t=90}
        \vspace{-9pt}
  		\includegraphics[width=1\linewidth]{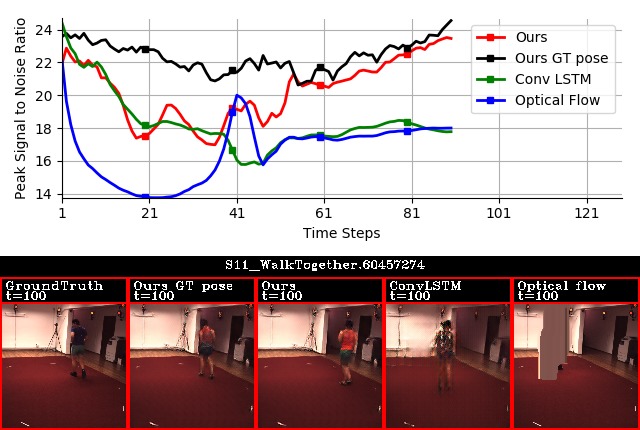}
  		\caption*{\textbf{High PSNR}}
	\end{subfigure}
	\vspace{-.3cm}
    \caption{Quantitative and visual comparison on Human 3.6M for selected time-steps for the action of \texttt{walking} (left) and \texttt{walk together} (right). Side by side video comparison can be found in our \href{https://goo.gl/U7UOfy}{project website}.}
\label{fig:h36mbad1}
\vspace{-40pt}
\end{figure}

\begin{figure}[htb!]
    \centering
    \vspace{30pt}
    \begin{subfigure}{0.40\linewidth}
        \caption*{t=36}
        \vspace{-9pt}
	    \includegraphics[width=1\linewidth]{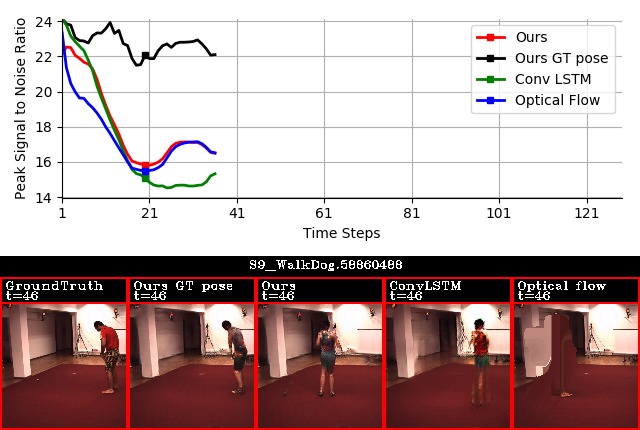}
  		\caption*{t=117}
        \vspace{-9pt}
  		\includegraphics[width=1\linewidth]{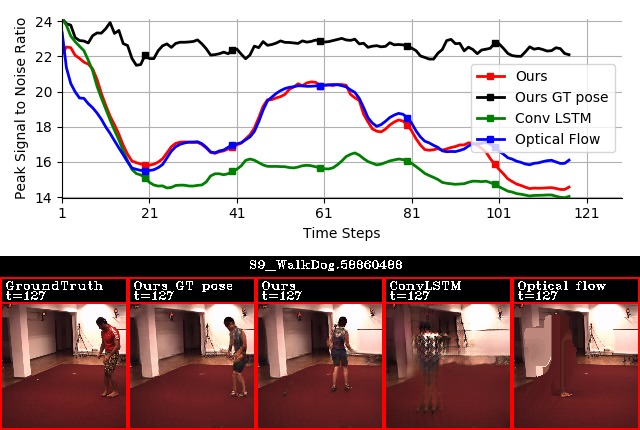}
  		\caption*{\textbf{Low PSNR}}
  		\caption*{-----------------------------------------------------------------}
  		\caption*{t=48}
        \vspace{-9pt}
	    \includegraphics[width=1\linewidth]{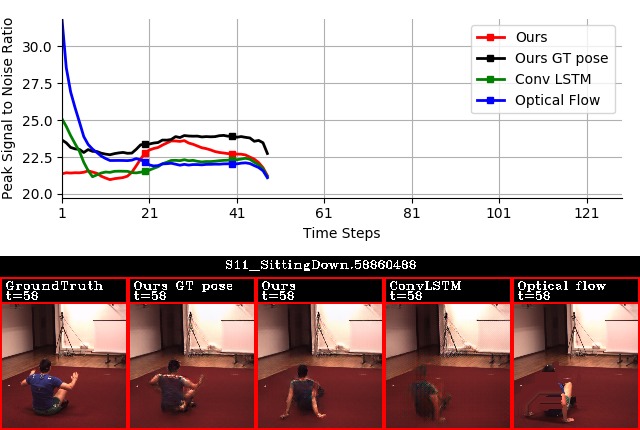}
  		\caption*{t=93}
        \vspace{-9pt}
  		\includegraphics[width=1\linewidth]{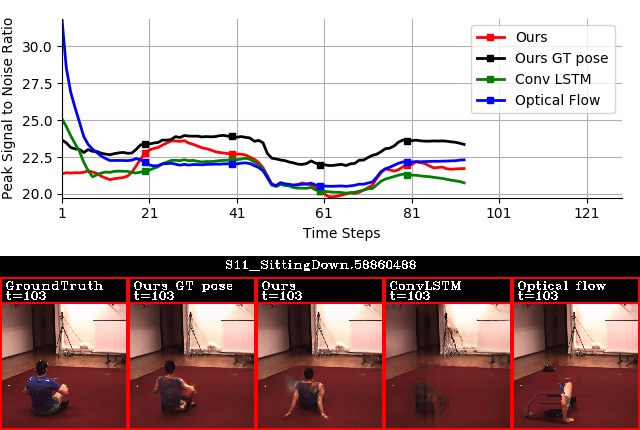}
  		\caption*{\textbf{Low PSNR}}
	\end{subfigure}
	\hspace{10pt}
    \begin{subfigure}{0.40\linewidth}
        \caption*{t=35}
        \vspace{-9pt}
	    \includegraphics[width=1\linewidth]{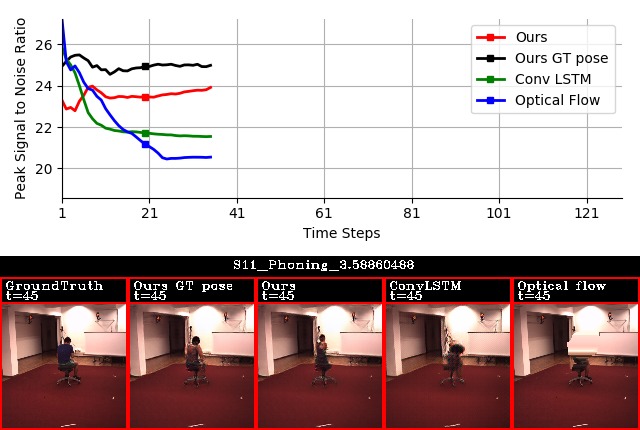}
  		\caption*{t=91}
        \vspace{-9pt}
  		\includegraphics[width=1\linewidth]{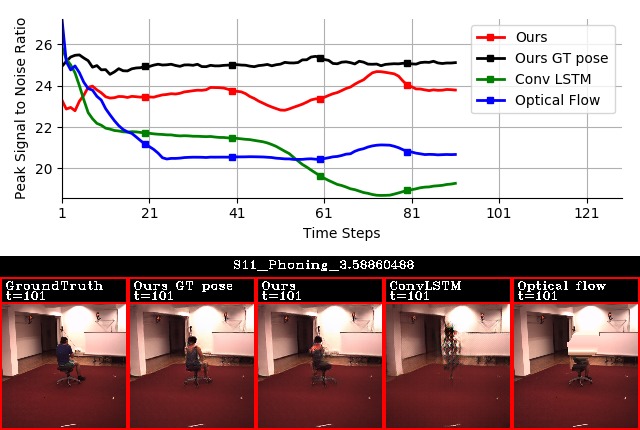}
  		\caption*{\textbf{High PSNR}}
  		\caption*{-----------------------------------------------------------------}
  		\caption*{t=61}
        \vspace{-9pt}
	    \includegraphics[width=1\linewidth]{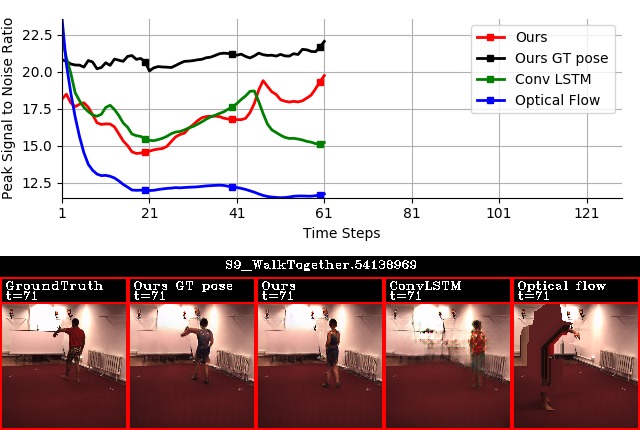}
  		\caption*{t=109}
        \vspace{-9pt}
  		\includegraphics[width=1\linewidth]{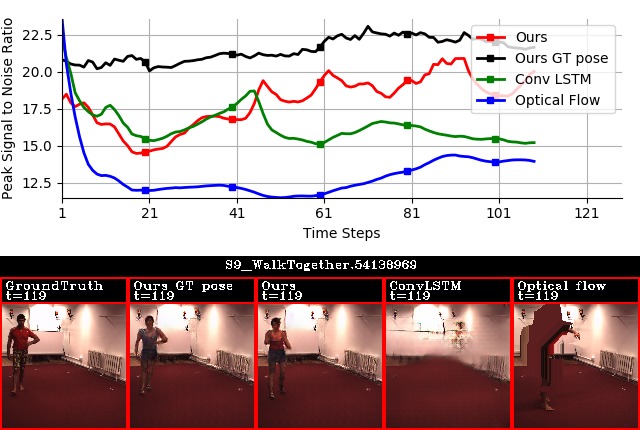}
  		\caption*{\textbf{High PSNR}}
	\end{subfigure}
	\vspace{-.1cm}
    \caption{Quantitative and visual comparison on  Human 3.6M for selected time-steps for the actions of \texttt{walk dog} (top left), \texttt{phoning} (top right), \texttt{sitting down} (bottom left), and \texttt{walk together} (bottom right). Side by side video comparison can be found in our \href{https://goo.gl/U7UOfy}{project website}.}
\label{fig:h36mbad2}
\vspace{-40pt}
\end{figure}

Second, we show snapshots of video prediction from different methods along with the PNSRs that change over time (Figures~\ref{fig:h36mbad1} and \ref{fig:h36mbad2}).
Our method tend to make plausible future pose trajectory but it can deviate from the ground-truth future pose trajectory; in such case, our method tend to achieve low PSNRs. 
However, when the future pose prediction from our method matches well with the ground-truth, the PSNR is much higher and the generated image frame is perceptually very similar to the ground-truth frame.
In contrast, optical flow and convolutional LSTM make prediction that often loses the structure of the foreground (e.g., human) over time, and eventually their predicted videos tend to become \emph{static}.
It is interesting to note that our method is comparable to convolutional LSTM in terms of PSNR, but that our method still strongly outperforms convolutional LSTM in terms of human evaluation, as described in Section~\ref{sec:experiments_h36m}. 

\clearpage

\begin{figure*}[!thbp]
    \centering
    \vspace{20pt}
	\begin{subfigure}{0.04\linewidth}
        \raggedleft
        \rotatebox{90}{
        \hspace{-.3cm}
        \parbox{2cm}{\centering Groundtruth} \parbox{2cm}{\centering Ours GT-pose} \parbox{2cm}{\centering Ours}
        \hspace{.1cm}
        \parbox{2cm}{\centering Groundtruth} \parbox{2cm}{\centering Ours GT-pose} \parbox{2cm}{\centering Ours}
        \hspace{.1cm}
        \parbox{2cm}{\centering Groundtruth} \parbox{2cm}{\centering Ours GT-pose} \parbox{2cm}{\centering Ours}
        }
    \end{subfigure}
    \begin{subfigure}{0.12\linewidth}
        \caption*{t=11}
        \vspace{-7pt}
	    \includegraphics[width=1\linewidth]{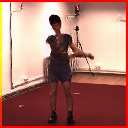}
	    \includegraphics[width=1\linewidth]{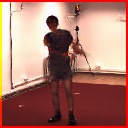}
	    \vspace{.2cm}
  		\includegraphics[width=1\linewidth]{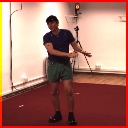}
  		\includegraphics[width=1\linewidth]{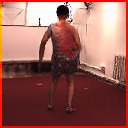}
  		\includegraphics[width=1\linewidth]{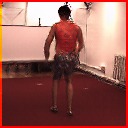}
  		\vspace{.2cm}
  		\includegraphics[width=1\linewidth]{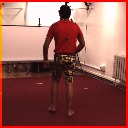}
  		\includegraphics[width=1\linewidth]{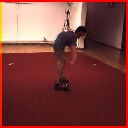}
  		\includegraphics[width=1\linewidth]{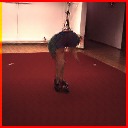}
  		\vspace{.2cm}
  		\includegraphics[width=1\linewidth]{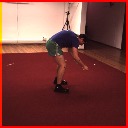}
	\end{subfigure} 
    \begin{subfigure}{0.12\linewidth}
        \caption*{t=29}
        \vspace{-7pt}
	    \includegraphics[width=1\linewidth]{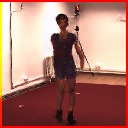}
	    \includegraphics[width=1\linewidth]{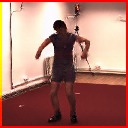}
	    \vspace{.2cm}
  		\includegraphics[width=1\linewidth]{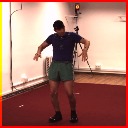}
  		\includegraphics[width=1\linewidth]{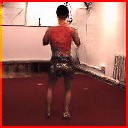}
  		\includegraphics[width=1\linewidth]{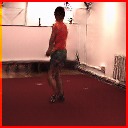}
  		\vspace{.2cm}
  		\includegraphics[width=1\linewidth]{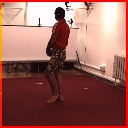}
  		\includegraphics[width=1\linewidth]{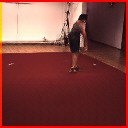}
  		\includegraphics[width=1\linewidth]{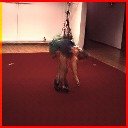}
  		\vspace{.2cm}
  		\includegraphics[width=1\linewidth]{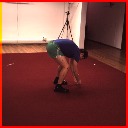}
	\end{subfigure} 
    \begin{subfigure}{0.12\linewidth}
        \caption*{t=47}
        \vspace{-7pt}
	    \includegraphics[width=1\linewidth]{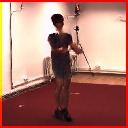}
	    \includegraphics[width=1\linewidth]{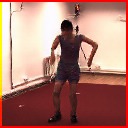}
	    \vspace{.2cm}
  		\includegraphics[width=1\linewidth]{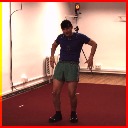}
  		\includegraphics[width=1\linewidth]{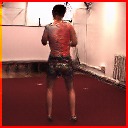}
  		\includegraphics[width=1\linewidth]{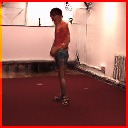}
  		\vspace{.2cm}
  		\includegraphics[width=1\linewidth]{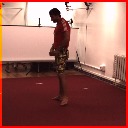}
  		\includegraphics[width=1\linewidth]{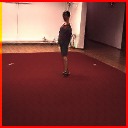}
  		\includegraphics[width=1\linewidth]{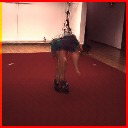}
  		\vspace{.2cm}
  		\includegraphics[width=1\linewidth]{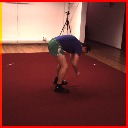}
	\end{subfigure} 
    \begin{subfigure}{0.12\linewidth}
        \caption*{t=65}
        \vspace{-7pt}
	    \includegraphics[width=1\linewidth]{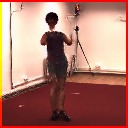}
	    \includegraphics[width=1\linewidth]{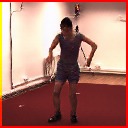}
	    \vspace{.2cm}
  		\includegraphics[width=1\linewidth]{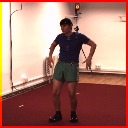}
  		\includegraphics[width=1\linewidth]{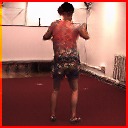}
  		\includegraphics[width=1\linewidth]{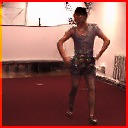}
  		\vspace{.2cm}
  		\includegraphics[width=1\linewidth]{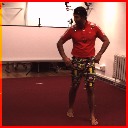}
  		\includegraphics[width=1\linewidth]{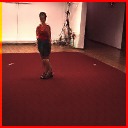}
  		\includegraphics[width=1\linewidth]{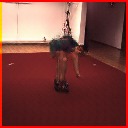}
  		\vspace{.2cm}
  		\includegraphics[width=1\linewidth]{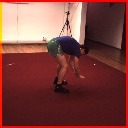}
	\end{subfigure}
	\begin{subfigure}{0.12\linewidth}
        \caption*{t=83}
        \vspace{-7pt}
	    \includegraphics[width=1\linewidth]{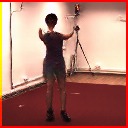}
	    \includegraphics[width=1\linewidth]{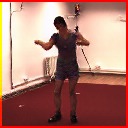}
	    \vspace{.2cm}
  		\includegraphics[width=1\linewidth]{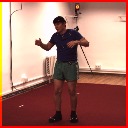}
  		\includegraphics[width=1\linewidth]{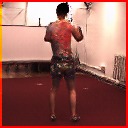}
  		\includegraphics[width=1\linewidth]{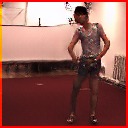}
  		\vspace{.2cm}
  		\includegraphics[width=1\linewidth]{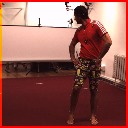}
  		\includegraphics[width=1\linewidth]{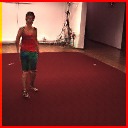}
  		\includegraphics[width=1\linewidth]{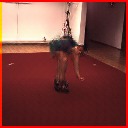}
  		\vspace{.2cm}
  		\includegraphics[width=1\linewidth]{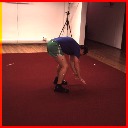}
	\end{subfigure}
	\begin{subfigure}{0.12\linewidth}
        \caption*{t=101}
        \vspace{-7pt}
	    \includegraphics[width=1\linewidth]{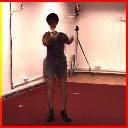}
	    \includegraphics[width=1\linewidth]{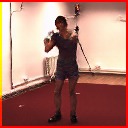}
	    \vspace{.2cm}
  		\includegraphics[width=1\linewidth]{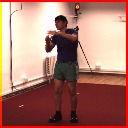}
  		\includegraphics[width=1\linewidth]{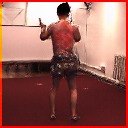}
  		\includegraphics[width=1\linewidth]{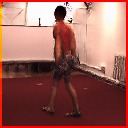}
  		\vspace{.2cm}
  		\includegraphics[width=1\linewidth]{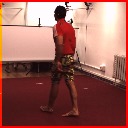}
  		\includegraphics[width=1\linewidth]{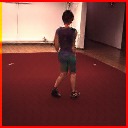}
  		\includegraphics[width=1\linewidth]{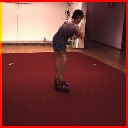}
  		\vspace{.2cm}
  		\includegraphics[width=1\linewidth]{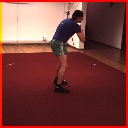}
	\end{subfigure}
	\begin{subfigure}{0.12\linewidth}
        \caption*{t=119}
        \vspace{-7pt}
	    \includegraphics[width=1\linewidth]{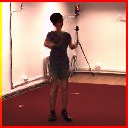}
	    \includegraphics[width=1\linewidth]{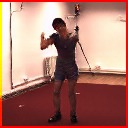}
	    \vspace{.2cm}
  		\includegraphics[width=1\linewidth]{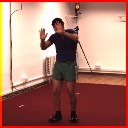}
  		\includegraphics[width=1\linewidth]{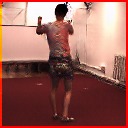}
  		\includegraphics[width=1\linewidth]{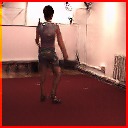}
  		\vspace{.2cm}
  		\includegraphics[width=1\linewidth]{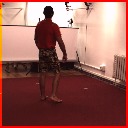}
  		\includegraphics[width=1\linewidth]{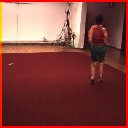}
  		\includegraphics[width=1\linewidth]{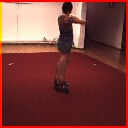}
  		\vspace{.2cm}
  		\includegraphics[width=1\linewidth]{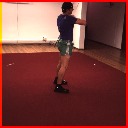}
	\end{subfigure}
    \vspace{-10pt}
    \caption{Qualitative evaluation of our network for long-term pixel-level generation. We show the actions of \texttt{giving directions} (top three rows), \texttt{posing} (middle three rows), and \texttt{walk dog} (bottom three rows). Side by side video comparison can be found in our \href{https://goo.gl/U7UOfy}{project website}.}
\label{fig:h36mgtft}
\vspace{-40pt}
\end{figure*}

\begin{figure*}[!thbp]
    \centering
    \vspace{60pt}
	\begin{subfigure}{0.04\linewidth}
        \raggedleft
        \rotatebox{90}{
        \hspace{-.3cm}
        \parbox{2cm}{\centering Groundtruth} \parbox{2cm}{\centering Ours GT-pose} \parbox{2cm}{\centering Ours}
        \hspace{.1cm}
        \parbox{2cm}{\centering Groundtruth} \parbox{2cm}{\centering Ours GT-pose} \parbox{2cm}{\centering Ours}
        \hspace{.1cm}
        \parbox{2cm}{\centering Groundtruth} \parbox{2cm}{\centering Ours GT-pose} \parbox{2cm}{\centering Ours}
        }
    \end{subfigure}
    \begin{subfigure}{0.12\linewidth}
        \caption*{t=11}
        \vspace{-7pt}
	    \includegraphics[width=1\linewidth]{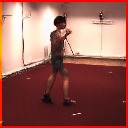}
	    \includegraphics[width=1\linewidth]{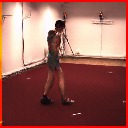}
	    \vspace{.2cm}
  		\includegraphics[width=1\linewidth]{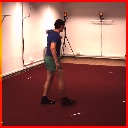}
  		\includegraphics[width=1\linewidth]{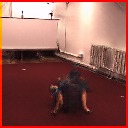}
  		\includegraphics[width=1\linewidth]{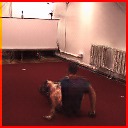}
  		\vspace{.2cm}
  		\includegraphics[width=1\linewidth]{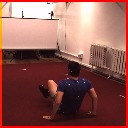}
  		\includegraphics[width=1\linewidth]{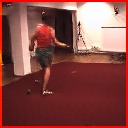}
  		\includegraphics[width=1\linewidth]{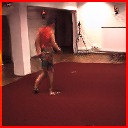}
  		\vspace{.2cm}
  		\includegraphics[width=1\linewidth]{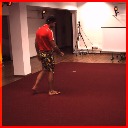}
	\end{subfigure} 
    \begin{subfigure}{0.12\linewidth}
        \caption*{t=29}
        \vspace{-7pt}
	    \includegraphics[width=1\linewidth]{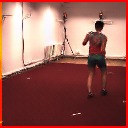}
	    \includegraphics[width=1\linewidth]{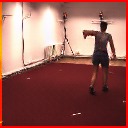}
	    \vspace{.2cm}
  		\includegraphics[width=1\linewidth]{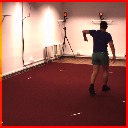}
  		\includegraphics[width=1\linewidth]{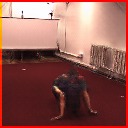}
  		\includegraphics[width=1\linewidth]{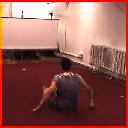}
  		\vspace{.2cm}
  		\includegraphics[width=1\linewidth]{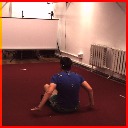}
  		\includegraphics[width=1\linewidth]{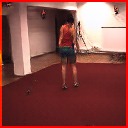}
  		\includegraphics[width=1\linewidth]{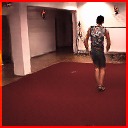}
  		\vspace{.2cm}
  		\includegraphics[width=1\linewidth]{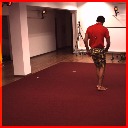}
	\end{subfigure} 
    \begin{subfigure}{0.12\linewidth}
        \caption*{t=47}
        \vspace{-7pt}
	    \includegraphics[width=1\linewidth]{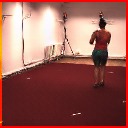}
	    \includegraphics[width=1\linewidth]{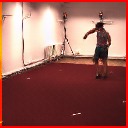}
	    \vspace{.2cm}
  		\includegraphics[width=1\linewidth]{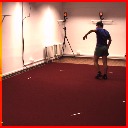}
  		\includegraphics[width=1\linewidth]{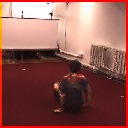}
  		\includegraphics[width=1\linewidth]{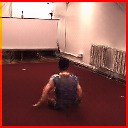}
  		\vspace{.2cm}
  		\includegraphics[width=1\linewidth]{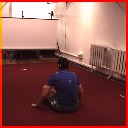}
  		\includegraphics[width=1\linewidth]{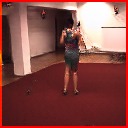}
  		\includegraphics[width=1\linewidth]{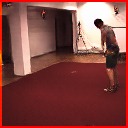}
  		\vspace{.2cm}
  		\includegraphics[width=1\linewidth]{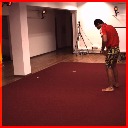}
	\end{subfigure} 
    \begin{subfigure}{0.12\linewidth}
        \caption*{t=65}
        \vspace{-7pt}
	    \includegraphics[width=1\linewidth]{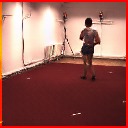}
	    \includegraphics[width=1\linewidth]{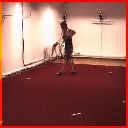}
	    \vspace{.2cm}
  		\includegraphics[width=1\linewidth]{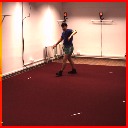}
  		\includegraphics[width=1\linewidth]{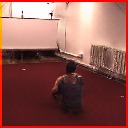}
  		\includegraphics[width=1\linewidth]{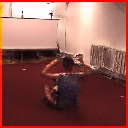}
  		\vspace{.2cm}
  		\includegraphics[width=1\linewidth]{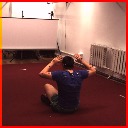}
  		\includegraphics[width=1\linewidth]{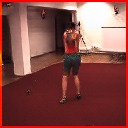}
  		\includegraphics[width=1\linewidth]{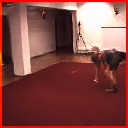}
  		\vspace{.2cm}
  		\includegraphics[width=1\linewidth]{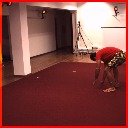}
	\end{subfigure}
	\begin{subfigure}{0.12\linewidth}
        \caption*{t=83}
        \vspace{-7pt}
	    \includegraphics[width=1\linewidth]{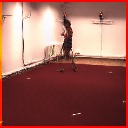}
	    \includegraphics[width=1\linewidth]{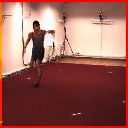}
	    \vspace{.2cm}
  		\includegraphics[width=1\linewidth]{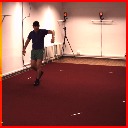}
  		\includegraphics[width=1\linewidth]{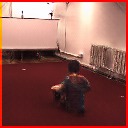}
  		\includegraphics[width=1\linewidth]{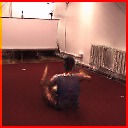}
  		\vspace{.2cm}
  		\includegraphics[width=1\linewidth]{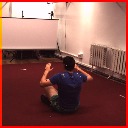}
  		\includegraphics[width=1\linewidth]{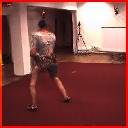}
  		\includegraphics[width=1\linewidth]{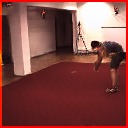}
  		\vspace{.2cm}
  		\includegraphics[width=1\linewidth]{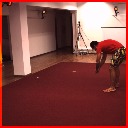}
	\end{subfigure}
	\begin{subfigure}{0.12\linewidth}
        \caption*{t=101}
        \vspace{-7pt}
	    \includegraphics[width=1\linewidth]{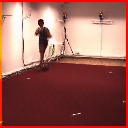}
	    \includegraphics[width=1\linewidth]{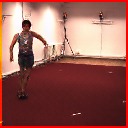}
	    \vspace{.2cm}
  		\includegraphics[width=1\linewidth]{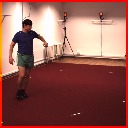}
  		\includegraphics[width=1\linewidth]{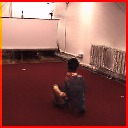}
  		\includegraphics[width=1\linewidth]{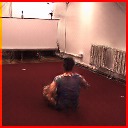}
  		\vspace{.2cm}
  		\includegraphics[width=1\linewidth]{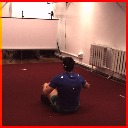}
  		\includegraphics[width=1\linewidth]{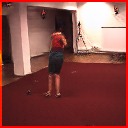}
  		\includegraphics[width=1\linewidth]{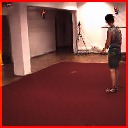}
  		\vspace{.2cm}
  		\includegraphics[width=1\linewidth]{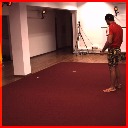}
	\end{subfigure}
	\begin{subfigure}{0.12\linewidth}
        \caption*{t=119}
        \vspace{-7pt}
	    \includegraphics[width=1\linewidth]{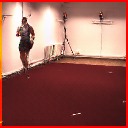}
	    \includegraphics[width=1\linewidth]{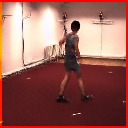}
	    \vspace{.2cm}
  		\includegraphics[width=1\linewidth]{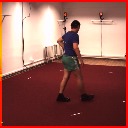}
  		\includegraphics[width=1\linewidth]{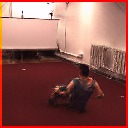}
  		\includegraphics[width=1\linewidth]{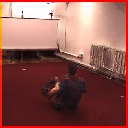}
  		\vspace{.2cm}
  		\includegraphics[width=1\linewidth]{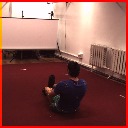}
  		\includegraphics[width=1\linewidth]{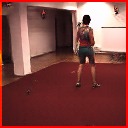}
  		\includegraphics[width=1\linewidth]{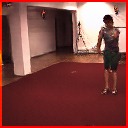}
  		\vspace{.2cm}
  		\includegraphics[width=1\linewidth]{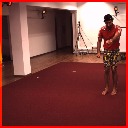}
	\end{subfigure}
    \vspace{-10pt}
    \caption{Qualitative evaluation of our network for long-term pixel-level generation.  We show the actions of \texttt{walk together} (top three rows), \texttt{sitting down} (middle three rows), and \texttt{walk dog} (bottom three rows). Side by side video comparison can be found in our \href{https://goo.gl/U7UOfy}{project website}.}
\label{fig:h36mgtft2}
\vspace{-40pt}
\end{figure*}

\clearpage
To directly compare our image generator using the predicted future pose (\texttt{Ours}) and the ground-truth future pose given by the oracle (\texttt{Ours GT-pose$^*$}), we present qualitative experiments in Figure~\ref{fig:h36mgtft} and Figure~\ref{fig:h36mgtft2}.
We can see that the both predicted videos contain the action in the video. However, the oracle based video reflects the exact future very well.

\clearpage

\end{appendix}
\end{document}